%% file: active.tex
\documentclass[12pt%
]{article}

\usepackage[margin=1in]{geometry}
\usepackage{setspace}
\usepackage{latexsym}
\usepackage{amssymb,amsmath, bm}
\usepackage{graphicx}
\usepackage{marvosym}
\usepackage{multirow}
\usepackage{makecell} % for more vertical space in cells
\usepackage{adjustbox} % to rotate table

\setcellgapes{5pt}
\usepackage{tikz}
\usetikzlibrary{shapes,arrows}
\usetikzlibrary{positioning}

\usepackage{natbib}
\usepackage{color}
\usepackage[hidelinks, bookmarksopen=true, bookmarksnumbered=true,
pdfstartview=FitH, breaklinks=true, urlbordercolor={0 1 0}, citebordercolor={0 0 1}]{hyperref}
\definecolor{darkgreen}{rgb}{0,0.545,0}
\definecolor{darkyellow}{rgb}{0.933,0.604,0}

\usepackage[ruled,vlined]{algorithm2e}

\usepackage{longtable}
\usepackage{booktabs}
\usepackage{multirow}
\usepackage{dcolumn}
\newcolumntype{.}{D{.}{.}{-1}}
\newcolumntype{d}[1]{D{.}{.}{#1}}

\usepackage{rotating}
\usepackage{tcolorbox}
\definecolor{cobalt}{rgb}{0.5, 0.5, 0.5}

\usepackage{bbm}

\newcommand{\blind}{0}

\usepackage{theorem}
\theoremstyle{break}
\theoremheaderfont{\scshape}
\theorembodyfont{\rm}

\newcommand\bbeta{{\boldsymbol \beta}}

\newcommand\aT{\textit{activeText}}

\usepackage{xr}
\usepackage{arydshln}

\usepackage[compact]{titlesec}

\usepackage{subfigure}

\graphicspath{{figs/}}

\allowdisplaybreaks[4]
\begin{document}

\newcommand\spacingset[1]{\renewcommand{\baselinestretch}%
{#1}\small\normalsize}

\spacingset{1.1}

%\makeatletter
%\def\thefootnote{\ifnum\c@footnote>\z@\leavevmode\lower.5ex\hbox{$^{*}$}\fi}
%\makeatother

\input{inputs/title}
\newcommand{\tit}{\titbase}

\input{inputs/author}

\pdfbookmark[1]{Title Page}{Title Page}

\thispagestyle{empty}
\setcounter{page}{0}

\begin{abstract}
\input{inputs/abstract}
\end{abstract}

\clearpage

%=== footnote number in the style of #)
% \makeatletter
% \def\thefootnote{\ifnum\c@footnote>\z@\leavevmode\lower.5ex\hbox{$^{\@arabic\c@footnote)}$}\fi}
% \makeatother

\onehalfspacing

\input{active_includes}

\pdfbookmark[1]{References}{References}
\bibliography{active}

% \newpage
% \appendix
% \input{inputs/appendix}
%\clearpage
%\input{inputs/lit_review}
\end{document}

\newcommand\spacingset[1]{\renewcommand{\baselinestretch}%
{#1}\small\normalsize}

\spacingset{1.1}

%\makeatletter
%\def\thefootnote{\ifnum\c@footnote>\z@\leavevmode\lower.5ex\hbox{$^{*}$}\fi}
%\makeatother

\input{inputs/title}
\newcommand{\tit}{{\bf Supplementary Information for} \titbase}

\input{inputs/author}

\pdfbookmark[1]{Title Page}{Title Page}
\thispagestyle{empty}

\tableofcontents
\thispagestyle{empty}

\setcounter{page}{0}
\clearpage

%=== footnote number in the style of #)
% \makeatletter
% \def\thefootnote{\ifnum\c@footnote>\z@\leavevmode\lower.5ex\hbox{$^{\@arabic\c@footnote)}$}\fi}
% \makeatother

\onehalfspacing

\appendix
\counterwithin{figure}{section}
\counterwithin{table}{section}

\input{inputs/appendix}

\newpage
\pdfbookmark[1]{References}{References}
\bibliography{active}

%% file: inputs/title.tex
\newcommand{\titbase}{\bf Improving Probabilistic Models in Text Classification via Active Learning}

%% file: inputs/author.tex
\if0\blind
{
\title{\tit\thanks{
\protect\input{inputs/acknowledgment}
}
}

\author{
 Mitchell Bosley\thanks{These authors have contributed equally to this work.} \thanks{
    Ph.D.~Candidate, Department of Political Science, University of Michigan.
    Email: \href{mailto:mcbosley@umich.edu}{\tt mcbosley@umich.edu}.
    }
    \and
 Saki Kuzushima\footnotemark[2] \thanks{
    Ph.D.~Candidate, Department of Political Science, University of Michigan.
    Email: \href{mailto:skuzushi@umich.edu}{\tt skuzushi@umich.edu}
    }
    \and
 Ted Enamorado\thanks{
    Assistant Professor, Department of Political Science, Washington University in St.~Louis.
    Siegle Hall, 244. One Brookings Dr. St Louis, MO 63130-4899. 
    Phone: 314-935-5810, Email: \href{mailto:ted@wustl.edu}{\tt ted@wustl.edu}, URL: \href{https://tedenamorado.com}{\tt www.tedenamorado.com}.
    }
    \and   
 Yuki Shiraito\thanks{
    Assistant Professor, Department of Political Science, University of Michigan.
    Center for Political Studies, 4259 Institute for Social Research, 426 Thompson Street, Ann Arbor, MI 48104-2321.
    Phone: 734-615-5165, Email: \href{mailto:shiraito@umich.edu}{\tt shiraito@umich.edu}, URL: \href{https://shiraito.github.io}{\tt shiraito.github.io}.
    }
    % \thanks{
    % Correspondence: Center for Political Studies, 4259 Institute for Social Research, 426 Thompson Street, Ann Arbor, MI 48104-2321.  Phone: 734-615-5165, Email: \href{mailto:shiraito@umich.edu}{\tt shiraito@umich.edu}.
    % }
 }

\date{
First draft: September 10, 2020 \\
This draft: September 23, 2022
}

\maketitle
}\fi

\if1\blind
{
\title{\bf \tit}
\date{}
\maketitle
}
\fi

%% file: inputs/acknowledgment.tex
We thank Ken Benoit, Yaoyao Dai, Chris Fariss, Yusaku Horiuchi, Kosuke Imai, Walter Mebane, Daichi Mochihashi, Kevin Quinn, and audiences at the 2020 Annual Meeting of the American Political Science Association, the 2021 Annual Meeting of the Midwest Political Science Association, the 11th Annual Conference on New Directions in Analyzing Text as Data, and the 2022 Summer Meeting of the Japanese Society for Quantitative Political Science, and seminar participants at the University of Michigan and members of the Junior Faculty Workshop at Washington University in St. Louis for useful comments and suggestions.

%% file: inputs/abstract.tex
Social scientists often classify text documents to use the resulting labels as an outcome or a predictor in empirical research.
Automated text classification has become a standard tool, since it requires less human coding.
However, scholars still need many human-labeled documents to train automated classifiers.
To reduce labeling costs, we propose a new algorithm for text classification that combines a probabilistic model with active learning.
The probabilistic model uses both labeled and unlabeled data, and active learning concentrates labeling efforts on difficult documents to classify.
Our validation study shows that the classification performance of our algorithm is comparable to state-of-the-art methods at a fraction of the computational cost.
Moreover, we replicate two recently published articles and reach the same substantive conclusions with only a small proportion of the original labeled data used in those studies.
We provide \emph{activeText}, an open-source software to implement our method.

% Social scientists often classify text documents to use the resulting labels as an outcome or a predictor in empirical research.
% Automated text classification has become a standard tool, since it requires less human coding.
% However, scholars still need many human-labeled documents for training.
% To reduce labeling costs, we propose a new algorithm for text classification that combines a probabilistic model with active learning.
% The probabilistic model uses both labeled and unlabeled data, and active learning concentrates labeling efforts on difficult documents to classify.
% Our validation study shows that the classification performance of our algorithm is comparable to state-of-the-art methods.
% Moreover, we replicate two published articles and reach the same substantive conclusions with only a small fraction of the original labeled data used in those studies.
% We provide open-source software to implement our method.

%% file: active_includes.tex
\input{inputs/intro}

\input{inputs/pedagogy}

\input{inputs/method}

\input{inputs/performance}

\input{inputs/reanalysis}

\input{inputs/discussion}

\input{inputs/concluding}

%% file: inputs/intro.tex
%%% (setq reftex-default-bibliography '("../active.bib"))
%%%
\section{Introduction}
As the amount and diversity of available information have rapidly increased, social scientists are increasingly resorting to multiple forms of data to answer substantive questions.
%See also Appendix~\ref{si-subsec:lit_review}.}
In particular, the use of text-as-data in social science research has exploded over the past decade.\footnote{See e.g.,~\cite{grim:stew:2013} for an excellent overview of these methods in political science.}
Document classification has been the primary task in political science, with researchers classifying documents such as legislative speeches \citep{peterson2018classification, moto:2020}, correspondences to administrative agencies \citep{lowande2018polices,lowande2019politicization}, public statements of politicians \citep{airoldi2007whose, stewart2009use}, news articles \citep{boydstun2013making}, election manifestos \citep{catalinac2016electoral}, social media posts \citep{king2017chinese}, treaties \citep{spirling2012us}, religious speeches \citep{nielsen2017deadly}, and human rights text \citep{farissphysical, greene2019machine} into two or more categories.
Researchers use the category labels of documents produced by the classification task as the outcome or predictive variable to test substantive hypotheses.

Statistical methods are used for document classification.
Although text data in political science is typically smaller than data in some other fields (where millions of documents are common), the cost of having human coders categorize all documents is still prohibitively high.
Relying on automated text classification allows researchers to avoid classifying all documents in their data set manually.

Broadly speaking, there are two types of classification methods: supervised and unsupervised algorithms. Supervised approaches use labels from a set of hand-coded documents to categorize unlabeled documents, whereas unsupervised methods cluster documents without needing labeled documents. Both of these methods have downsides, however: in the former, hand-coding documents is labor-intensive and costly; in the latter, the substantive interpretation of the categories discovered by the clustering process can be difficult.

% Broadly speaking, there are two types of classification methods: supervised and unsupervised algorithms.
% Supervised approaches use associations between word frequencies and labels from a set of hand-coded documents 
% to categorize unlabeled documents, whereas unsupervised methods cluster documents without needing labeled documents.
% Both of these methods have downsides, however: in the former, hand-coding documents is labor-intensive and costly, 
% and requires expert knowledge and reconciliation of disagreements between coders to ensure label validity; in the latter, the 
% substantive interpretation of the categories discovered by the clustering process can be difficult, and performance is 
% severely threatened when the data lacks the necessary structure such that signal can be distinguished from noise.

Supervised methods are more popular in political science research because substantive interpretability is important in using category labels to test substantive hypotheses, and justifies the cost associated with labeling many documents manually.
For example, \citet{gohdes2020repression} hand-labeled about $2000$ documents, and \citet{park:etal:2020} used $4000$ human-coded documents. These numbers are much smaller than the size of their entire data sets ($65,274$ and $2,473,874$, respectively), however, having human coders label thousands of (potentially long and complicated) documents still requires a large amount of researchers' time and effort.

We propose \aT, a new algorithm that augments a probabilistic mixture model with active learning.
We use the mixture model of \citet{nigam2000text} to combine the information from both labeled and unlabeled documents, making use of all available information.
%The main disadvantage of the method advanced by \citet{nigam2000text} is that pre-existing labeled data is required to train and test the classifier.
%\citet{nigam2000text} require labeled data to train and test the classifier, but in practice, only in exceptional cases a social science researcher will have access to labeled data in advance.
%Moreover, labeling data is considered an expensive and time-consuming task, thus even if a researcher is willing to obtain labeled data, there is no guidance on how to obtain the most informative data to be labeled such that the process is more efficient.
%\textbf{Better connection here.}
In the model, latent classes are observed as labels for labeled documents and estimated as a latent variable for unlabeled documents.
% , and
% active learning iteratively provides observed labels for the documents that the cluster estimates are most uncertain about.
%
%
Active learning is a technique that reduces the cost of hand-coding.
It uses measures of label uncertainty to iteratively flag highly informative documents to reduce the number of labeled documents needed to train an accurate classifier, particularly when the classification categories are imbalanced.
% However, current implementations of active learning have only been used to augment supervised approaches.
% That is, in each iteration of an active learning algorithm, only labeled documents are used to train the 
% %supervised 
% classifier that indicates which documents should be labeled.
% % in the next active step.

Our validation study shows that our model outperforms Support Vector Machines (SVM), a popular supervised learning model when both models are using active learning.
We also show that our algorithm performs favorably in terms of classification accuracy when compared to an off-the-shelf version of Bidirectional Encoder Representations from Transformers (BERT), a state-of-the-art classification model in natural language processing, using several orders of magnitude less computational resources.
Furthermore, because our model is generative, it is straightforward to use a researcher's domain expertise, such as keywords associated with a category, to improve text classification.
% For example, we show that iteratively upweighting model parameters associated with keywords that the researcher identifies as highly associated with one of the possible document labels can improve classification.

We also use \aT\ to replicate two published political science studies and show that the authors of these papers could have reached the same substantive conclusions with fewer labeled documents.
The first study is \citet{gohdes2020repression}, which focuses on the relationship between internet access and the form of state violence.
The second study is \citet{park:etal:2020}, which analyzes the association (or the lack thereof) between information communication technologies (ICTs) and the U.S.~Department of State's reports on human rights.
For both studies, we replicate their text classification tasks using \aT\ and conduct the same empirical analyses using the document labels.
Our replication analysis recovers their original conclusions---a higher level of internet access is associated with a larger proportion of targeted killings, and ICTs are not associated with the sentiment of the State Department's human rights reports, respectively---using far fewer labeled documents.
These replication exercises demonstrate that \aT\ performs well on complex documents commonly used in political science research, such as human rights reports.

% Specifically, we improve upon existing approaches in three ways.
% First, we show that our model outperforms simpler active learning methods based just on supervised learning in terms of accuracy.
% Second, we show that in some cases, allowing multiple latent clusters to be linked with one category improves the classification performance. Furthermore, we show that accuracy is also improved when a set of keywords is used as auxiliary information to inform classification. Third, 
% we develop the open-source R package \textit{activeText}, which provides fast and scalable implementation of our methods.
%%The \textit{activeText} package will allow researchers in a variety of fields to quickly and intuitively use the proposed methods to accurately classify large collections of text.
We provide an \textbf{R} package called \aT\ with the goal of providing researchers from all backgrounds with easily accessible tools to minimize the amount of hand-coding of documents and improve the performance of classification models for their own work.

Before proceeding to a description of our algorithm and analysis, we first offer an accessible primer on the use of automated text classification.
We introduce readers to several basic concepts in machine learning:
tokenization, preprocessing, and the encoding of a corpus of text data into a matrix;
the difference between supervised and unsupervised learning, between discriminative and generative models, and between active and passive learning;
and a set of tools for the evaluation of classification models.
Readers who are already well acquainted with these concepts may prefer to skip directly to the description of our model in Section~\nameref{sec:method}.

%% file: inputs/pedagogy.tex
\section{Using Machine Learning for Text Classification}
\label{sec:teaching}

\subsection{Encoding Text in Matrix Form}
Suppose that a researcher has a collection of social media text data, 
called a corpus, and wishes to classify whether each text in a corpus is 
political (e.g., refers to political protest, human rights violations, unfavorable views of a given candidate,
targeted political repression, etc.) or not solely based on the words used in a given observation.\footnote{For
simplicity, the exposition here focuses on a binary classification task, however, 
our proposed method can be extended to multiple classes e.g., classifying a
document as either a positive, negative, or neutral position about a candidate.
See Sections \nameref{sec:method} and \nameref{sec:reanalysis}, and Supplementary Information (SI)~\ref{si-sec:multiple_class} for more details.}
Critically, the researcher does not yet know which of the texts are
political or not at this point.

The researcher must first choose how to represent text as a series of \textit{tokens}, and decide which tokens to include in their analysis.
This involves a series of sub-choices, such as whether each token represents an individual word (such as ``political'') or a combination of words (such as ``political party''), whether words should be stemmed or not (e.g., reducing both ``political'' and ``politics'' their common stem ``politic''), and whether to remove stop-words (such as ``in'', ``and'', ``on'', etc.) that are collectively referred to as \textit{pre-processing.}\footnote{For a survey of pre-processing techniques and their implications for political science research, see \citet{denny2018text}.}

The researcher must then choose how to encode information about these tokens in matrix form.
% There are two main approaches: bag-of-words or word embeddings.
The most straightforward way to accomplish this is using a \textit{bag-of-words} approach, where the corpus is transformed into a document-feature matrix (DFM) \(\mathbf{X}\) with \(n\) rows and \(m\) columns,
where \(n\) is the number of documents and \(m\) is the number of tokens, which are more generally referred to as
features.\footnote{Note that in the machine learning literature, the concept
  typically described by the term ``variable'' is communicated using the term ``feature.''}
Each element of the DFM encodes the frequency that a token occurs in a given document.\footnote{An alternative to the bag-of-words approach is to encode tokens as \textit{word embeddings}, where in addition to the matrix summarizing the incidences of words in each document, neural network models are used to create vector representations of each token. In this framework, each token is represented by a vector of some arbitrary length, and tokens that are used in similar contexts in the corpus (such as ``minister'' and ``cabinet'') will have similar vectors. While this approach is more complicated, it yields considerably more information about the use of words in the corpus than the simple count that the bag-of-words approach does. For an accessible introduction to the construction and use of word embeddings in political science research, see \citet{rodriguez2022word}. For a more technical treatment, see \citet{pennington2014glove}.}
Once the researcher chooses how to encode their corpus as a matrix, she is left with a set of features corresponding to each document \(\mathbf{X}\) and an unknown vector of true labels \(Y\), where each element of \(Y\) indicates
whether a given document is political or not.
Then, we can repose the classification question as follows: given \(\mathbf{X}\), how might we best learn \(Y\), that is, whether each document
is political or not?

% \textbf{[MB]: We should use this section to (1) introduce readers who are new to this literature to the basic modeling decisions that have to be made re: discriminative vs. generative, supervised vs. unsupervised, active vs. passive while (2) at the same time justifying our choices and explaining how we innovate in this space. We can take the info from 3.1 and expand on it in this section, leaving section 3 only for the description of our model.}

\subsection{Supervised vs. Unsupervised Learning}
\label{subsec:sup-unsup}

A researcher must then choose whether to use a
supervised or unsupervised approach to machine learning.\footnote{For a comprehensive discussion on supervised and unsupervised algorithms for the analysis of text as data, we refer the interested reader to \citet{grimm:etal:2022}.}
The supervised approach to this problem would be to (1) obtain true labels 
of some of the documents using human coding  e.g., an expert classifies documents such as the following news headline by CNN: 
 ``White House says Covid-19 policy unchanged despite President Biden's comments that the `pandemic is over''' as political or not;
(2) learn the relationship between the text features encoded in the matrix $\mathbf{X}$ 
and the true label encoded in the vector \(Y\) for the documents with 
known labels. In other words, it learns the importance of words such as ``policy'', ``President'', ``Biden'', ``pandemic'' in explaining
whether a document refers to politics or not;\footnote{That is, learn \(P(Y_{\text{labeled}}| \mathbf{X}_{\text{labeled}})\). 
This can be accomplished with a variety of models, including e.g. linear or logistic 
regression, support vector machines (SVM), Naive Bayes, $K$-nearest neighbor, etc.}
and (3) using the learned association between the text data and the known labels, predict 
whether the remaining documents in the corpus (that is, those that were not coded by 
a human) are political or not. 

In contrast, an unsupervised approach would \textit{not} obtain the true 
labels of some of the documents.
Rather, a researcher using an unsupervised approach would choose a 
model that \textit{clusters} documents from the corpus that have common patterns 
of word frequency.\footnote{Examples of clustering algorithms include \(K\)-means
and Latent Dirichlet Allocation (LDA).}
%SK: Removed \(K\)-Nearest Neighbor (KNN) from the list of clustering algorithms because it is included as an example of supervised ones.  
% Using this model, the researcher would choose the number of discrete 
% classes to divide the corpus into, and learn the relationship between the the 
% matrix of text data $\mathbf{X}$ and each of the possible clusters to 
% assign each document to a cluster.
Using the assignment of documents to clusters, the researcher would then use some 
scheme to decide which of the clusters corresponds to the actual outcome of 
interest: whether a document is political or not.

% The main advantage of a supervised approach over an unsupervised approach is the direct interpretability of results,
% since a well-defined measure of the concept of interest exists. In other words,
% it does not include the step of translating the clustering of documents to the 
% classification of documents as political or not.
% Relatedly, as noted by \cite{hast:etal:2009}, in supervised learning, the success of an algorithm 
%  is an objective measure that can be used to assess its suitability for a particular situation. 
%  For example, for each observation, success can be assessed by 
%  the distance between the predictions made by the supervised learning algorithm and the true values of  \(Y\). 
%  Such an objective measure of success does not exist in unsupervised
%  learning (as \(Y\) is unknown). Thus, in unsupervised learning, the researcher needs to 
% rely on heuristics to assess the adequacy of the algorithm.
The main advantage of a supervised approach over an unsupervised approach is the direct interpretability of results, since it requires the translation of clusters to classifications. This also allows for a more straightforward evaluation of model performance in terms of the distance between the predictions made by the supervised learning algorithm and the true values of $Y$. Because such an objective measure does not exist in unsupervised learning, the researcher needs to rely on heuristics to assess the adequacy of the algorithm \citep{hast:etal:2009}.\footnote{In most political science 
applications of unsupervised learning techniques, the author either is 
conducting an exploratory analysis and is therefore uninterested in classification, 
or performs an \textit{ad hoc} interpretation of the clusters by reading top examples 
of a given cluster, and on that basis infers the classification from the 
clustering \citep{knox:etal:2022}.} 

On the other hand, the main disadvantage of a supervised approach is that 
obtaining labels for the documents in the corpus is often time-consuming and costly.
For example, it requires expert knowledge to classify each document to be either political
or non-political.  Researchers using an unsupervised approach instead 
will avoid this cost since they do not require a set of 
labels \textit{a priori}.

Semi-supervised methods combine the strengths of supervised and unsupervised approaches to improve classification \citep{mille:uyar:1997, nigam2000text}.
These methods 
are particularly useful in situations where there is a large amount of 
unlabeled data, and acquiring labels is costly.
A semi-supervised model proceeds similarly to the supervised 
approach, with the difference being that the model learns the 
relationship between the matrix of text data $\mathbf{X}$ and the 
classification outcome \(Y\) using information from both the labeled 
and unlabeled data.\footnote{While $Y$ is not observed for
the unlabeled data, these observations do contain information
about the joint distribution of the features $\mathbf{X}$,
and as such can be used with labeled data to increase the accuracy of a text
classifier \citep{nigam2000text}.} 
% The balance between information from the labeled and unlabeled data varies depending on the model used.
% In general, though, because
Since a supervised approach learns the 
relationship between the labels and the data solely based on 
the labeled documents, a classifier trained with a supervised approach 
maybe less accurate than if it were provided information from 
both the labeled and unlabeled documents \citep{nigam2000text}.

\subsection{Discriminative vs. Generative Models}
\label{subsec:disc-gen}

In addition to choosing a supervised, unsupervised, or semi-supervised approach, 
a researcher must also choose whether to use a discriminative or generative model.
As noted by \citet{ng:jord:2001} and \citet{bish:lass:2007}, 
when using a discriminative model (e.g., logistic regression, SVM, etc.), the 
goal is to directly estimate the probability of the classification outcomes \(Y\) 
given the text data $\mathbf{X}$ i.e., directly estimate \(p(Y| \mathbf{X})\).
In contrast, when using a generative model (e.g., Naive Bayes), learning the 
relationship between the \(Y\) and $\mathbf{X}$ is a two-step process.
In the first step, the likelihood of the matrix of text data $\mathbf{X}$ and 
outcome labels \(Y\) is estimated given the data and a set of parameters \(\theta\) 
that indicate structural assumptions about how the data is generated 
i.e., \(p(\mathbf{X}, Y | \theta)\) is directly estimated.
In the second step, the researcher uses Bayes' rule to calculate the probability of 
the outcome vector given the features and the learned distribution of 
the parameters i.e.,  \(p(Y| \mathbf{X}; \theta)\).

In addition to allowing for the use of unlabeled data (which reduces labeling costs), 
one of the main benefits of a generative rather than a discriminative model is that 
the researcher can include information they know about the data 
generating process by choosing appropriate functional forms.\footnote{This is 
particularly true when e.g., the researcher knows that the data has a complicated 
hierarchical structure since the hierarchy can be incorporated directly into the generative model.}
This can help prevent overfitting when the amount of data in a corpus is small.\footnote{
Overfitting occurs when a model learns to predict classification outcomes based on patterns 
in the training set (i.e., the data used to fit the model) that does not generalize to the broader universe of cases to be classified. A 
model that is overfitted may predict the correct class with an extremely high degree of accuracy 
for items in the training set, but will perform poorly when used to predict the class for items 
that the model has not seen before.
}
Conversely, because it is not necessary to model the data generating process directly, the 
main benefit of a discriminative rather than generative model is simplicity (in general
it involves estimating fewer parameters).
Discriminative models are therefore appropriate in situations where the amount of data 
in a corpus is very large, and/or when the researcher is unsure about the data-generating 
process, which could lead to mis-specification \citep{bish:lass:2007}.\footnote{Another benefit of generative
models is that they can yield better estimates 
of how certain we are about the relationship between the outcome and the features. This is the case 
when a researcher uses an inference algorithm like Markov Chain Monte Carlo (MCMC) that 
learns the entire distribution for each of the parameters, rather than only point estimates.}

\subsection{Model Evaluation}
\label{subsec:model-eval}
% In the previous section, we referred to rules for stopping the labeling of new documents in terms of measures of model performance such as accuracy or F1.
A researcher must also decide when she is satisfied with the predictions generated by the model.
In most circumstances, the best way to evaluate the performance of a classification algorithm is to reserve a subset of the corpus for validation, which is sometimes referred to as validation and/or test set.
At the very beginning of the classification process, a researcher puts aside and label a set of randomly chosen documents that the active learning algorithm does not have access to.\footnote{It is important to use a set-aside validation set for testing model performance, rather than a subset of the documents used to train the model, to avoid \textit{overfitting}.}
Then, after training the model on the remainder of the documents (often called the training set), the researcher should generate predictions for the documents in the validation set using the trained model.
By comparing the predicted labels generated by the model to the actual labels, the researcher can evaluate how well the model does at predicting the correct labels.

A common tool for comparing the predicted labels to the actual labels is a \textit{confusion matrix}.
In a binary classification setting, a confusion matrix will be a 2 by 2 matrix, with rows corresponding to the actual label, and the columns corresponding to the predicted label. Returning to our running example, imagine that the classification is to predict whether documents are political or not, Table~\ref{tab:conf-mat} shows the corresponding confusion matrix. In this scenario, True Positives (TP) are the number of documents that the model predicts to be about politics and that is in fact labeled as such.
Correspondingly, True Negatives (TN), are the number of documents that the model predicts to be non-political and is labeled as such in the validation set.
%%The upper-right and bottom-left quadrants provide counts of False Negative (FN) and False Positives (FP), respectively.
A False Negative (FN) occurs when the model classifies a document as non-political, but according to the validation set, the document is about politics.
Similarly, a False Positive (FP) occurs when the model classifies as political a document that is non-political.

\begin{table}
  \centering
  \begin{tabular}{cc|cc}
    \multicolumn{2}{c}{}
    &   \multicolumn{2}{c}{Predicted Label} \\
    &       &   Political &   Non-political              \\
    \cline{2-4}
    \multirow{2}{*}{Actual Label}
    & Political   & True Positive (TP) & False Negative (FN)                 \\
    & Non-political    & False Positive (FP) & True Negative (TN) \\
%    \cline{2-4}
  \end{tabular}
  \caption{\textbf{Confusion Matrix: Comparison of the Predictions of a Classifier to Documents' True Labels}}
  %A confusion matrix compares the results of a classification model to documents' true labels. The upper-left quadrant is the count of True Positives (TP), the number of documents that the model predicts are the positive classification outcome that are labeled as such. Correspondingly, the bottom-right quadrant is the count of True Negatives (TN), the number of documents that the model predicts to be negative which are labeled negative in the validation set. The upper-right and bottom-left quadrants provide counts of False Negative (FN) and False Positives (FP), respectively.}
  \label{tab:conf-mat}
\end{table}

Using the confusion matrix, the researcher can calculate a variety of evaluation statistics.
Some of the most common of these are accuracy, precision, and recall.
Accuracy is the proportion of documents that have been correctly classified.
Precision is used to evaluate the false positivity rate and is the proportion of the model's positive classifications that are true positives.
As the number of false positives increases (decreases), precision decreases (increases).
Recall is used to evaluate the false negativity rate, and is the proportion of the actual positive documents that are true positives.
As the number of false negatives increases, recall decreases, and \textit{vice-versa}.
Accuracy, precision, and recall can be formally calculated as:
\[
  \text{Accuracy} = \frac{\text{TP} + \text{TN}}{\text{TP} + \text{TN} + \text{FP} + \text{FN}} \qquad
  \text{Precision} = \frac{\text{TP}}{\text{TP} + \text{FP}} \qquad
  \text{Recall} = \frac{\text{TP}}{\text{TP} + \text{FN}}
\]

When the proportion of political and non-political documents in a corpus is balanced,
accuracy is an adequate measure of model performance.
However, it is often the case in text classification that the corpus is unbalanced, and the proportion of
documents associated with one class is low.
When this is the case, accuracy does a poor job at model evaluation.
%%, and precision and recall should be considered.
Consider the case when 99 percent of documents are non-political, and 1 percent are about politics.
A model which simply predicts that all documents belong to the non-politics class would have an accuracy score of 0.99, but would be poorly suited to the actual classification task. In contrast, the precision and recall rates would be 0, which would signal to the researcher that the model does a poor job
at classifying documents as political. Precision and recall are not perfect measures of model performance, however.
There is a fundamental trade-off involved in controlling the false positivity and false negativity rates: you can have few false positives if you are content with an extremely high number of false negatives, and you can have few false negatives if you are content with an extremely high number of false positives.

%%A model that classified all of the actual positives correctly (i.e., when there are no false positives) but classified all of the actual negatives incorrectly would get a perfect precision score.
%%Similarly, a model that classified all documents as positive would have a perfect recall.
%%Note, however, that in the case with the perfect precision score, recall would be extremely low.
%%And in the case with the perfect recall score, precision would be extremely low.
%These examples illustrate the fact that there is a fundamental trade-off involved in controlling the false positivity and false negativity rates: you can have few false positives if you are content with an extremely high number of false negatives, and you can have few false negatives if you are content with an extremely high number of false positives.

Recognizing this trade-off, researchers often combine precision and recall scores to find a model that has the optimal balance of the two.
One common way of combining the two is an F1 score, which is the harmonized mean of precision and recall.
Formally, the F1 score is calculated as:
\[\text{F1} = 2 \cdot \frac{\text{Precision} \cdot \text{Recall}}{\text{Precision} + \text{Recall}}\]
The F1 score evenly weights precision and recall, and so a high F1 score would indicate that both the false negativity and false positivity rate are low.
It is worth noting these evaluation measures (accuracy, precision, recall, and the F1 score) are computed using labeled data (``ground truth''), which in practice,
are available only for a limited subset of the records.

\subsection{Active vs. Passive Learning}
\label{subsec:active-learning}

Finally, if the researcher in our running example decides to use a supervised or
semi-supervised approach for predicting whether documents in their corpus are political
or not, the next step is to decide how many documents to label, and how to choose them.
% In supervised and semi-supervised settings,
% increasing the size of the pool of labeled cases has proven to be a promising
% avenue to improve multiclass text classifiers \citep{nigam2000text}.
% However, as discussed above, 
Since labeling is the bottleneck of any classification task of this kind, it is critical that she also selects an approach to label observations that minimizes the number of documents to be labeled in order to produce an accurate classifier.

% Not only will an expert (or group of experts) need to carefully provide a label
% that matches the concept of interest, but the researcher needs to define
% which observations to label such that if she is presented with a new unlabeled data point
%  from the same source, the classifier would agree with the label the expert would produce
% if she were asked to do so \citep{hann:2014}.
%%As described in Section \ref{subsec:sup-unsup}, a researcher using a supervised
%%(or semi-supervised) approach must choose to label some documents, and on 
%%the basis of the learned relationship between those documents and the classification 
%%outcome, predict whether the rest of the documents in the corpus are political or not.

There are two popular strategies on how to retrieve cases to be labeled: 1) passively 
and 2) actively. The difference between a passive and an active approach amounts to 
whether the researcher randomly chooses which documents to label (i.e., choose documents 
\textit{passively}), or whether to use some selection scheme (i.e., choose documents 
\textit{actively}). Ideally, an active approach must require fewer labels than the number 
of randomly labeled data sufficient for a passive approach to achieve the same level of 
accuracy.

% Active and passive learning are algorithms where a model interacts
% with a labeler (a human) iteratively. At each step, the classifier
% is trained and then, the labeler provides newly annotated data points. The
% difference between approaches is whether an active or a passive approach
% to labeling is used.
\citet{cohn:etal:1994} and \citet{lewi:gale:1994}
established that a good active learning algorithm should be fast, and should reliably 
choose documents for labeling that provide more information to 
the model than a randomly chosen document, particularly in situations when the amount
of labeled data is scarce.\footnote{See also \cite{dasg:2011, settles2011active, hann:2014, hino:2021} and the references therein.}
One of the most studied active learning approaches 
is called \textit{uncertainty sampling} \citep{lewi:gale:1994, yang:2015}, a process 
where documents are chosen for labeling based on how uncertain the model 
is about the correct classification category 
for each document in the corpus.\footnote{This is just one of many possible approaches. 
Other uncertainty-based approaches to active learning include query-by-committee, variance reduction, 
expected model change, etc. We refer the interested reader to~\citet{settles2011active} for an
accessible review on active learning and \citet{hann:2014} for a more technical exposition.
}

\begin{figure}
  \begin{center}
    \includegraphics[width = 16cm, height = 6.5cm]{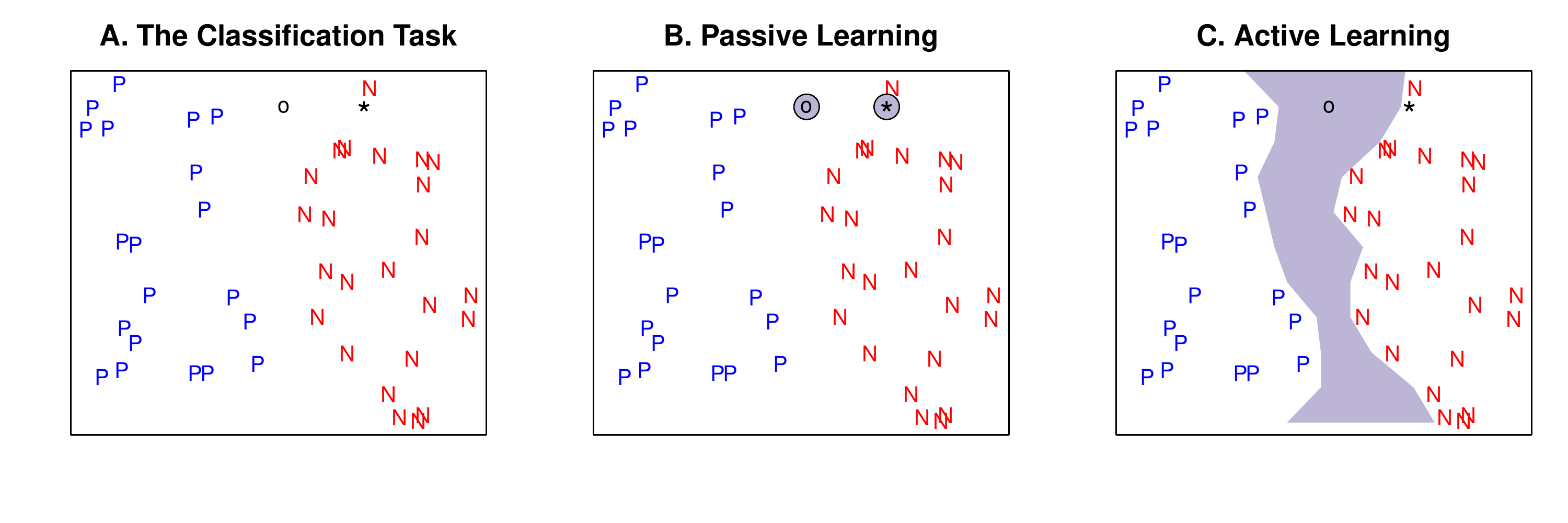}
  \end{center} \vspace{-16mm}
  \caption{\textbf{Passive vs Active Learning.} For a classifier defined in two dimensions, Panel A illustrates
  the task: classify unlabeled documents (denoted by $\circ$ and $\ast$) as Political (P) or Non-political (N). A
a passive learning algorithm will request the labels of   $\circ$ and $\ast$ with equal probability (Panel C). In contrast, in
active learning approach, $\circ$ will be prioritized for labeling as it is located in the region where the classifier is most
  uncertain (shaded region).} \label{fig:active:passive}
\end{figure}

As noted above, an active learning process using uncertainty sampling alternates between estimating 
the probability that each document belongs to a particular classification outcome, sampling 
a subset of the documents that the model is most uncertain about for labeling,\footnote{
While in our presentation, we have focused on instances of labeling one observation per iteration,
exactly how many observations to select and label at each 
active iterations is also an important practical consideration for any researcher. 
As noted by \citet{hoi:etal:2006}, to reduce the cost of retraining the model per instance of labeling, labeling many
documents per iteration (as a batch) is the best approach. This is especially 
important when working with a large amount of data.}
%%When both the document selection scheme and 
%%the model fit are fast, labeling a single document (that is, the document the model is most 
%%uncertain about) is optimal. On the other hand, if the model takes some time to fit, and there 
%%are a large number of documents that potentially need labeling, a batch approach is justifiable.}
 then estimating the probabilities again using the information from the 
 newly labeled documents. In our running example, a researcher is interested in
classifying documents as political (P) or non-political (N), and needs to decide how to prioritize
her labeling efforts. 
As shown in Figure~\ref{fig:active:passive} (Panel A), imagine
there are two new data points to be labeled (denoted by ``$\circ$''  and ``$\ast$''). 
A passive learning algorithm would give equal labeling priority to both (Panel B). However, an active approach 
would give priority to ``$\circ$'' as the classifier is most uncertain about the 
label of ``$\circ$'' if compared to ``$\ast$'' (which is surrounded by many non-political documents).
% The question then becomes, which approach to use?

% An active learning approach is superior to a passive one when (1) the information
% that some documents contribute to the model results in more accurate predictions
% than the information contributed by other documents would and (2) there is a direct
% way of predicting which unlabeled documents will provide the best information.
% When both of these conditions hold, an active learning approach will be more efficient than a passive one.
% Conversely, when either of these conditions does not hold (as when randomly
% selected documents provide as good or better information to the model as one
% chosen by a particular scheme), a passive approach is as good as an active one \citep{hann:2014}.
%%Alternatively, if the active approach performs slightly better than the passive approach, but 
%%is computationally intensive and/or very time-consuming to run, then one 
%%may be better off using the passive approach.

A critical question for a researcher using an iterative algorithm is when to stop labelling.
Many active learning algorithms resort to
heuristics such as a fixed-budget approach, which stops when
the number of newly labeled data points reaches a predetermined size. The problem with 
such an approach is that it may lead to under- or over-sampling.\footnote{This is due to the fact the fixed budget
has not been set using an optimality criterion other than to stop human coding at some point. See \citet{ishi:hide:2020} for further discussion of this point.}
One popular strategy is to randomly label a subset of documents at the beginning of the process, which is then used for assessing the performance of the classifier on data that the model has not seen.\footnote{For a discussion of this approach in our own application, see Section~\nameref{subsec:model-eval}.}
% Many accuracy-based approaches for stopping rules use out-of-sample predictions obtained
% after each iteration of the active learning
% algorithm.
With this approach, the process stops when the difference in measures of out-of-sample accuracy
between two consecutive iterations does not surpass a certain threshold pre-established by the researcher (e.g.,
the F1 score does not improve in more than 0.01 units from iteration to iteration) \citep{alts:bood:2019}.
If labeled data does not exist or cannot be set aside for testing due to its scarcity, a stopping rule where
the algorithm stops once in-sample predictions generated by the model (i.e., using the documents that have been labeled by the researcher during the active learning process) do not change from
one iteration to the next.
This is often referred to as a stability-based method \citep{ishi:hide:2020}.

With all these concepts in mind, in the next section we describe our proposed approach with a special
focus on its flexibility that it affords a researcher to both balance the tradeoffs of working with labeled and unlabeled data, and use existing domain expertise to improve classification with the use of keyword upweighting.

%%% Local Variables:
%%% mode: yatex
%%% TeX-master: "../active.tex"
%%% End:

%% file: inputs/method.tex
\section{The Method}
\label{sec:method}

In this section, we present our modeling strategy and describe our active learning algorithm. 
For the probabilistic model (a mixture model for discrete data) at the heart of the algorithm, we build on the work of~\cite{nigam2000text}, who show
that probabilistic classifiers can be augmented by combining the information coming from labeled and unlabeled data.
In other words, our model makes the latent classes for the unlabeled data interpretable 
by connecting them to the hand-coded classes from the labeled data. It also takes advantage of the fact that 
the unlabeled data provides more information about the features used to predict 
the classes for each document. As we will discuss below, we insert our model into 
an active learning algorithm and use the Expectation-Maximization (EM) algorithm \citep{demp:etal:1977}
to maximize the observed-data log-likelihood function and estimate the model
parameters.

\subsection{Model}
%\label{subsec:model}
%We present our model and the observed document likelihood in Section~\ref{subsubsec:prob_model}, and show how we use the EM algorithm to estimate the class and word probability parameters in~\ref{subsubsec:em}.

%\subsubsection{Probabilistic Model}
%\label{subsubsec:prob_model}
%%Our model takes a semi-supervised approach introduced in Section~\ref{subsec:sup-unsup}, which takes a middle ground between supervised and unsupervised approaches.

Consider the task of classifying $N$ documents as one of two classes (e.g., political vs non-political).
Let $\mathbf{D}$ be a $N \times V$ document feature matrix, where $V$ is the size of features.
%We use a binary $N \times 2$ matrix, $C$, to indicate the class of each document.
%If a document $i$ is assigned to the negative class, $C_{i1} = 1$ and $C_{i2} = 0$.
%If it is assigned to the positive class, $C_{i1} = 0$ and $C_{i2} = 1$.
We use $\mathbf{Z}$, a vector of length $N$, where each entry represents the latent classes assigned to each document.
If a document $i$ is assigned to the $k$th class, we have that $Z_i = k$, where $k \in \{0, 1 \}$ e.g., in our running example,
$k = 1$ represents the class of documents about politics, and $k = 0$ those that are non-political. 
Because we use a semi-supervised approach, it can be the case that some documents 
are already hand-labeled. This means that the value of $Z_i$ is known for the labeled 
documents and is unknown for unlabeled documents. 
To facilitate exposition, we assume that the classification goal is binary, however,  
our approach can be extended to accommodate for 
1) multiclass classification setting, where $k > 2$ and each document needs to be classified into
one of the $k$ classes e.g., classifying 
news articles into 3 classes: politics, business, and sports; and 2) 
modeling more than two classes but keeping the final classification to be binary. 
In other words, a hierarchy that maps multiple sub-classes into one class e.g., 
collapsing the classification of documents that are about business and sports 
into a larger class (non-politics), and letting the remaining documents to 
be about politics (the main category of interest). 
(For more details, see SI~\ref{si-subsec:em}, \ref{si-sec:multiple_cluster_model}, and \ref{si-sec:multiple_class}). 

The following sets of equations summarize the model:\\

\begin{tcolorbox}[colback=blue!0,colframe = cobalt,title= {\bf Labeled Data}]
{\small 
\vspace{-3mm}  
\begin{center}
\begin{eqnarray*}
    Z_i = k &\sim& \textrm{hand-coded}, \quad k \in \{0,1\}  \\
    \mathbf{\eta}_{\cdot k} &\stackrel{i.i.d}{\sim} &Dirichlet(\boldsymbol{\beta}_k) \\
    \mathbf{D}_{i\cdot} \vert Z_{i} = k  &\stackrel{i.i.d}{\sim}& Multinomial(n_i, \boldsymbol{\eta}_{\cdot k}) \\
\end{eqnarray*}
\end{center}
}  
\end{tcolorbox}
\vspace{-10mm}
\begin{center}
\Huge{+}
\end{center}
\vspace{-3mm}
\begin{tcolorbox}[colback=blue!0,colframe=cobalt, title= {\bf $\lambda$ $\cdot$ Unlabeled Data}]
{\small 
\vspace{-3mm}
\begin{center}
\begin{eqnarray*}
    \pi &\sim& Beta(\alpha_0, \alpha_1) \\
    Z_i = k&\stackrel{i.i.d}{\sim} &Bernoulli(\pi), \quad k \in \{0,1\}  \\
    \mathbf{\eta}_{\cdot k} &\stackrel{i.i.d}{\sim}& Dirichlet(\boldsymbol{\beta}_k) \\
    \mathbf{D}_{i\cdot} \vert Z_{i} = k  &\stackrel{i.i.d}{\sim}& Multinomial(n_i, \boldsymbol{\eta}_{\cdot k}) \\
\end{eqnarray*} 
\end{center}
}
\end{tcolorbox}

If document $i$ is unlabeled, we first draw $\pi = p(Z_i = 1)$, the overall probability that any given document belongs to the first class
(e.g., political documents), from a Beta distribution with hyperparameters $\alpha_0$ and $\alpha_1$. 
Similarly, for the other class (e.g., non-political documents), we have that $1 - \pi = p(Z_i = 0)$.
Given $\pi$, for each document indexed by $i$, we draw the latent cluster assignment indicator $Z_i$ from a 
Bernoulli distribution. % to the positive class 
Then, we draw features for document $i$ from a multinomial distribution governed by
the vector $\boldsymbol{\eta}_{\cdot k}$, where $\eta_{v k} = p(D_{i v} | Z_i = k)$, whose prior is the Dirichlet distribution.
If document $i$ is labeled, the main difference with the unlabeled data case is that
$Z_i$ has been hand-coded, and as a result, we do not draw it from a Bernoulli distribution but 
the rest of the model's structure remains the same. 

It is worth emphasizing that one of the most notorious problems with the implementation of supervised and semi-supervised approaches is the scarcity of labeled data, especially if compared to the abundance of unlabeled data. 
Due to this imbalance problem, for any classifier to be able to extract signal from the labeled data and not be informed by unlabeled data alone, it is key to devise ways to increase the relative importance of the labeled data. Otherwise, the unlabeled data will mute the signal coming from the labeled data. 
Following \cite{nigam2000text}, we down-weight information from unlabeled documents by $\lambda \in [0, 1]$.
%%The objective function to be maximized by the EM algorithm consists of the log prior, the log-likelihood of labeled data, and the log-likelihood of unlabeled data. 
Note that when the \(\lambda\) is equal to 1, the model treats each document equally, regardless of whether the document is labeled deterministically by a human, or probabilistically by the algorithm.
As \(\lambda\) moves from 1 towards 0, the model increasingly down-weights the information that the probabilistically labeled documents contribute to the estimation of \(\boldsymbol{\eta}\) and \(\pi\), such that when \(\lambda\) is 0, the model \textit{ignores} all information from the probabilistically labeled documents and therefore becomes a supervised algorithm (see SI~\ref{si-subsec:em}).
Finally, because the observed data log-likelihood of our model is difficult to maximize, we use the EM algorithm to 
estimate the parameters.\footnote{For a full derivation of the EM algorithm, see SI~\ref{si-subsec:em}.} %(See Appendix~\ref{subsec:em} for more details).

\subsection{Active Learning}
\label{subsec:active}
Our active learning algorithm (see Algorithm~\ref{alg:active_learning}) can be split into the following steps: \textit{estimation} of the probability that each unlabeled document belongs to the positive class, \textit{selection} of the unlabeled documents whose predicted class is most uncertain, and \textit{labeling} of the selected documents by human coders.
The algorithm iterates until a stopping criterion is met (Section~\nameref{subsec:active-learning}).
%The algorithm then iterates until one of three stopping conditions are met: (1) the model runs out of unlabeled documents to label; (2) the remaining unlabeled documents do not meet a particular uncertainty threshold; or (3) the maximum number of allowed active steps is reached.
We also describe an optional keyword upweighting feature, where a set of user-provided keywords provide prior information about the likelihood that a word is generated by a given class to the model.
These keywords can either be provided at the outset of the model or identified during the active learning process.

\subsubsection{Estimation}
\label{subsubsec:estimation}
In the first iteration, the model is initialized with a small number of labeled documents.\footnote{While we assume that these documents are selected randomly, the researcher may choose any subset of labeled documents with which to initialize the model.}
The information from these documents is used to estimate the parameters of the model: the probability of a document being of class 1 ($\pi$), and the probability of generating each word given a class, the $V \times 2$ matrix $\boldsymbol{\eta}$.
From the second iteration on, we use information from both labeled and unlabeled documents to estimate the parameters using the EM algorithm, with the log-likelihood of unlabeled documents being down-weighted by $\lambda$, and
%%\footnote{When the \(\lambda\) parameter is equal to 1, the model treats word count information from each document equally, regardless of whether the document labeled by a human or by the model.
%%As the \(\lambda\) parameter moves from 1 towards 0, the model increasingly down-weights the information that the pseudo-labeled documents contribute to the estimation of \(\boldsymbol{\eta}\) and \(\pi\), such that when \(\lambda\) is 0, the model \textit{ignores} all information from the pseudo-labeled documents.}
with the \(\boldsymbol{\eta}\) and \(\pi\) values from the previous iteration as the initial values.
Using the estimated parameters, we compute the posterior probability that each unlabeled document belongs to class 1.

\subsubsection{Selection}
Using the predicted probability that each unlabeled document belongs to class 1, we use Shannon Entropy to determine which of the probabilistically labeled documents that it was least certain about.
In the binary classification case, this is the equivalent of calculating the absolute value of the distance of the class 1 probability and 0.50 for each document.
%If, for a given document, the model predicts that it is \textit{equally likely} to belong to either class (i.e., the probability of belonging to each class is 0.5), we would say that the model is \textit{maximally uncertain} about the class that the document belongs to.
%On the other hand, if the model predicts that a document has a probability of belonging to particular class of 1.0, we would say that the model is \textit{maximally certain} about the class that the document belongs to.
Using this criterion, the model ranks all probabilistically labeled documents in descending order of uncertainty.
The \(n\) most uncertain documents are then selected for human labeling, where \(n\) is the number of documents to be labeled by humans at each iteration. 

\subsubsection{Labeling}
A human coder reads each document selected by the algorithm and imputes the ``correct'' label.
For example, the researcher may be asked to label as political or non-political each of the following sentences:
\begin{quotation}
 The 2020 Presidential Election had the highest turnout in US history.

 Qatar is ready to host the FIFA World Cup this coming November.
\end{quotation}
These newly-labeled documents are then added to the set of human-labeled documents, and the process is repeated from the estimation stage.

\subsubsection{Stopping Rule}
Our method is highly modular and supports a variety of stopping rules.
This includes an internal stability criterion, where stoppage is based on small amounts of change of the internal model parameters, as well as the use of a small held-out validation set to assess the marginal benefit of labeling additional documents on measures of model evaluation such as accuracy or F1.
With either rule, the researcher specifies some bound such that if the change in model parameters or out-of-sample performance is less than the pre-specified bound, then the labeling process ends.
We use the out-of-sample validation stopping rule with a bound of 0.01 for the F1 score in our reanalyses in Section~\nameref{sec:reanalysis}.

\begin{algorithm}[t!]
  \SetAlgoLined \KwResult{Obtain predicted classes of all documents.}

  Randomly select a small subset of documents, and ask humans to label them\;
  %Initialize $\mathbf{D}^l_{old}$ by sampling some documents randomly, and have humans
  %label them \; Initialize $\mathbf{D}^u \gets \mathbf{D} \setminus \mathbf{D}^l_{old}$\;

  \textcolor{gray}{[\textbf{Active Keyword}]: Ask humans to provide initial keywords}\;
  %Initialize keyword matrix
  %\(\boldsymbol{\kappa}\), where each element \(\kappa_{v,c}\) takes the value of \(\gamma\) if the word \(v\) is a keyword for class \(c\), otherwise \(0\).}

  \While{Stopping conditions are not met yet}{

    \textcolor{gray}{(1) [\textbf{Active Keyword}]: Up-weight the important of keywords associated with a class;}\
    %Up-weight elements of the \(\bbeta\) prior using \(\boldsymbol{\kappa}\)\; \quad
    %  \(\bbeta \gets \bbeta + \boldsymbol{\kappa}\)%\\ \quad
      % \(\beta_{vc} \leftarrow \beta_{vc} + \gamma, \forall v \in \kappa_{v,c} = 1, \forall c \in C\)
      % \(\beta \gets \{\beta \mid \beta_{c,v^{*}} + \gamma \,\, \forall \,\, v^{*} \in \kappa_{c, v^{*}}\}\)

    (2) Predict labels for unlabeled documents using EM algorithm\;

    (3) Select documents with the highest uncertainty among unlabeled documents, and ask humans to label them\;
    %Sample $n$ most uncertain documents in $\mathbf{D}^u$ and have humans labels
    %them\; \quad $\mathbf{D}^l_{new} \gets$ $n$ most uncertain documents in $\mathbf{D}^u$\;

    \textcolor{gray}{(4) [\textbf{Active Keyword}]:
    Select words most strongly associated with each class, and ask humans to label them;}\
      %Sample \(m\) non-keywords most associated with each class
      %and have humans label to create \(\boldsymbol{\kappa}_{new}\)\; \quad
      %\(\boldsymbol{\kappa} \gets \boldsymbol{\kappa}_{new}\)\; }

    (5) Update sets of labeled and unlabeled documents for the next iteration\;
    %and unlabeled documents\; \quad $\mathbf{D}^l \gets \mathbf{D}^l_{old} \cup \mathbf{D}^l_{new}$\; \quad
    %$\mathbf{D}^u \gets \mathbf{D} \setminus \mathbf{D}^l_{old}$
  }
  \caption{Active learning with EM algorithm to classify text}
  \label{alg:active_learning}
\end{algorithm}

\subsubsection{Active Keyword Upweighting}
\label{subsubsec:keywords}
The researcher also has the option to use an active keyword upweighting scheme, where a set of keywords is used to provide additional information.
This is done by incrementing elements of the \(\boldsymbol{\beta}\) (the prior of $\boldsymbol{\eta}$) by \(\gamma\), a scalar value chosen by the researcher.
In other words, we impose a tight prior on the probability that a given keyword is associated with each class.\footnote{See \citet{eshima2020keyword} for a similar approach for topic models.} 
% In the passive scheme, the user provides a set of keywords associated with each class at the outset of the model.
To build the set of keywords for each class, 1) \aT\ proposes a set of candidate words, 2) the researcher decides whether they are indeed keywords or not,\footnote{The researcher may also provide an initial set of keywords, and then iteratively adds new keywords.} and 3) \aT\ updates the parameters based on the set of keywords.

To select a set of candidate keywords, \aT\ calculates the ratio that each word was generated by a particular class using the \(\boldsymbol{\eta}\) parameter. 
Specifically, it computes $\eta_{vk}/\eta_{vk'}$ for $k = \{0, 1\}$ with $k'$ the opposite class of $k$, and chooses top \(m\) words whose $\eta_{vk}/\eta_{vk'}$ are the highest as candidate keywords to be queried for class $k$.\footnote{Words are excluded from candidate keywords if they are already in the set of keywords, or if they are already decided as non-keywords. Thus, no words are proposed twice as candidate keywords.)}
Intuitively, words closely associated with the classification classes are proposed as candidate keywords. 
For example, words such as ``vote,'' ``election,'' and ``president,'' are likely to be proposed as the keywords for the political class of documents in the classification between political vs. non-political documents. 

After \aT\ proposes candidate keywords, the researcher decides whether they are indeed keywords or not.
This is where the researcher can use her expertise to provide additional information. 
For example, she can decide names of legislators and acronyms of bills as keywords for the political class.\footnote{See SI~\ref{si-subsec:mislabel-keywords} for more discussion about what if the researcher mislabels keywords.}

Using the set of keywords for each class, \aT\ creates a \(V \times 2\) keyword matrix \(\boldsymbol{\kappa}\) where each element \(\kappa_{v,k}\) takes the value of \(\gamma\) if word \(v\) is a keyword for class \(k\), otherwise \(0\).
Before we estimate parameters in each active iteration, we perform a matrix sum \(\boldsymbol{\beta} \gets \boldsymbol{\kappa} + \bbeta\) to incorporate information from keywords.
The keyword approach therefore effectively upweights our model with prior information about words that the researcher thinks are likely to be associated with one class rather than another.

%%% Local Variables:
%%% mode: yatex
%%% TeX-master: "../active.tex"
%%% End:

%% file: inputs/performance.tex
\section{Validation Performance}
\label{sec:performance}

This section shows the performance comparisons between \aT and other classification methods. 
First, we show comparisons between active vs. passive learning as well as semi-supervised learning vs. supervised learning. 
For semi-supervised learning, we use \aT with $\lambda = 0.001$.
For supervised learning, we use active Support Vector Machines (SVM) from \citet{millerActiveLearningApproaches2020} with margin sampling.
Then, we compare classification and time performance between \aT\ and an off-the-shelf version of BERT, a state-of-the-art text classification model.
Furthermore, we show how keyword upweighting can improve classification accuracy.

We compare the classification performance on the following documents: internal forum conversations of Wikipedia editors (class of interest: toxic comment), BBC News articles (political topic), the United States Supreme Court decisions (criminal procedure), and Human Rights allegations (physical integrity rights allegation).\footnote{More information about preprocessing and descriptions about the dataset are in SI~\ref{si-sec:validation_specification}}
We use 80\% of each dataset for the training data and hold out the remaining 20\% for evaluation.
Documents to be labeled are sampled only from the training set, and documents in the test set are not included to train the classifier, even in our semi-supervised approach. 
The out-of-sample F1 score is calculated using the held-out testing data.
%All documents are labeled with ground-truth labels.
%%Then, 
% It also shows that different specifications of our methods can further improve the performance depending on various data structures.

\subsection{Comparison between \aT\ and Active SVM}
\begin{figure}[p!]
  \includegraphics[width=\linewidth]{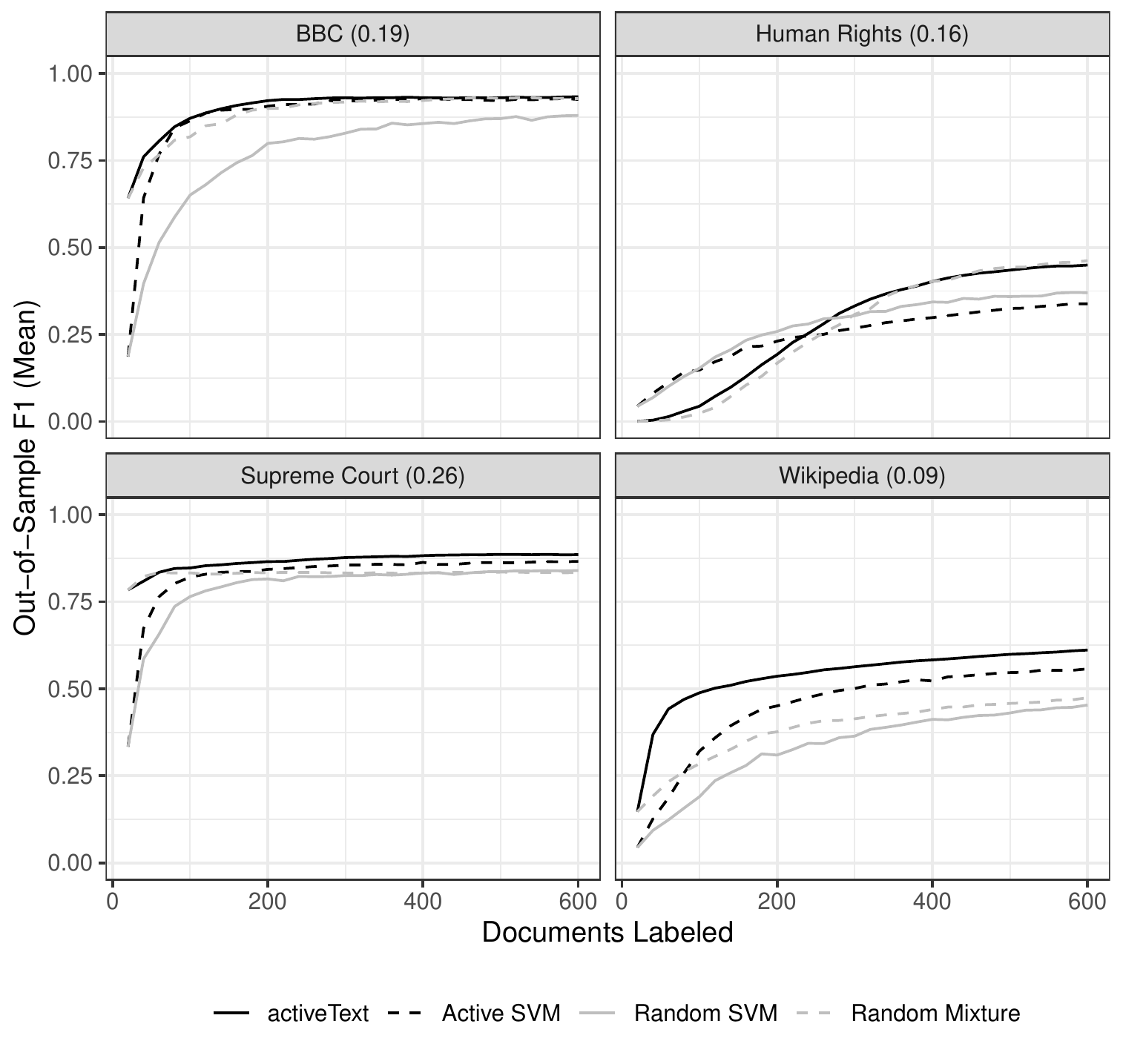}
  \caption{\textbf{Comparison of Classification Results across Random and Active Versions of \aT\ and SVM}
%  The rows correspond to different datasets and the columns correspond to various proportion with positive label documents in the corpus.
%  The y-axis indicates the out-of-sample F1 score and the x-axis shows the number of sampling steps. 
%  20 documents are labeled at each sampling step.
%  The colors correspond to different classifiers: the darker lines show the results of our mixture model and the lighter lines the results of SVM.
%  The line type shows different sampling schemes: the solid lines are for the active sampling and the dashed line are for the random sampling.
%  In the active sampling, we select the next set of documents to be labeled based on the entropy of the predicted probabilities of the classes.
%  \textit{Active Mixture} performs the best with most of the specifications. 
%  Active learning often performs better than passive learning when the proportion of positive labels are small.
  }
  \label{fig:main}
\end{figure}

Figure \ref{fig:main} shows the results from four model specifications, each representing one of the combinations of active or passive learning, and semi-supervised or supervised learning.
The first choice is between active learning (solid lines) vs passive learning (dashed lines).
In the active sampling, we select the next set of documents to be labeled based on the entropy of the predicted probabilities of the classes when we use our mixture model, and they are selected based on the margin sampling when we use SVM as the underlying classification method.
The second choice is between our semi-supervised learning (darker lines) vs. off-the-shelf supervised learning (lighter lines).
For the supervised learning, we replicate the results from \citet{millerActiveLearningApproaches2020} which uses SVM as the classifier.
% The rows correspond to different datasets and the columns correspond to various proportion with positive label documents in the corpus.
Each panel represents model performance in one of four datasets, with the number in parentheses indicating the proportion of documents associated with the class of interest using ground-truth labels in each dataset.
The y-axis indicates the average out-of-sample F1 score across 50 Monte Carlo iterations, and the x-axis shows the total number of documents labeled, with 20 documents labeled at each sampling step.\footnote{While we simulate human coders who label all documents correctly at the labeling stage, this may not be the case because humans can make mistakes in practice. SI~\ref{si-subsec:mislabel_documents} shows that honest (random) mistakes in the labeling of documents can hurt the classification performance.}

Among the four models, the combination of active learning with the mixture model (\aT\ in Figure \ref{fig:main}) performs the best with most of the specifications. 
% The results on the Wikipedia corpus with 5\% and 9\% (population) positive labels and on the Supreme Court corpus with 5\% positive labels highlight this most clearly.
% With other specifications, \aT performs slightly better or as well as other models.
The gain from active learning tends to be higher when the proportion of documents in the class of interest is small.
On the Wikipedia corpus with the proportion of the positive labels being 9\%, active learning outperforms passive learning, particularly when the number of documents labeled is smaller.
In SI~\ref{si-sec:main_results_appdx}, we further examine how the class-imbalance influences the benefit of active learning, by varying the proportion of the positive class between 5\% and 50\%.\footnote{See SI~\ref{si-sec:validation_specification} for how we generate data with class-imbalance.}
It shows that active learning performs better than passive learning consistently when the proportion of one class is 5\%.% while the difference is marginal when the proportion of positive class is 50\%. 
%An exception is on the human rights corpus where active learning did not improve the performance even when the proportion of positive labels is 5\%, and this is the case for both SVM and our mixture model.
%Furthremore, our semi-supervised approach to perform better than supervised approach (SVM) on BBC, Wikipedia, and Supreme Court data. 
%The \textit{Random Mixture} (dashed dark line) is above the \textit{Random SVM} (dashed light line) at the beginning but they converge later. 
%This makes sense because the relative contribution from unlabeled documents in the mixture model decreases as the size of labeled documents increases.
 One limitation is that \aT\ did not perform better than SVM on the human rights corpus when the number of documents labeled is small (less than 200 in Figure~\ref{fig:main}).
We examine how the optional keyword labeling can assist such a situation in \nameref{subsec:keyword_results}.

\subsection{Comparison between \textit{activeText} and BERT}
\begin{figure}[p!]
  \includegraphics[width=\linewidth]{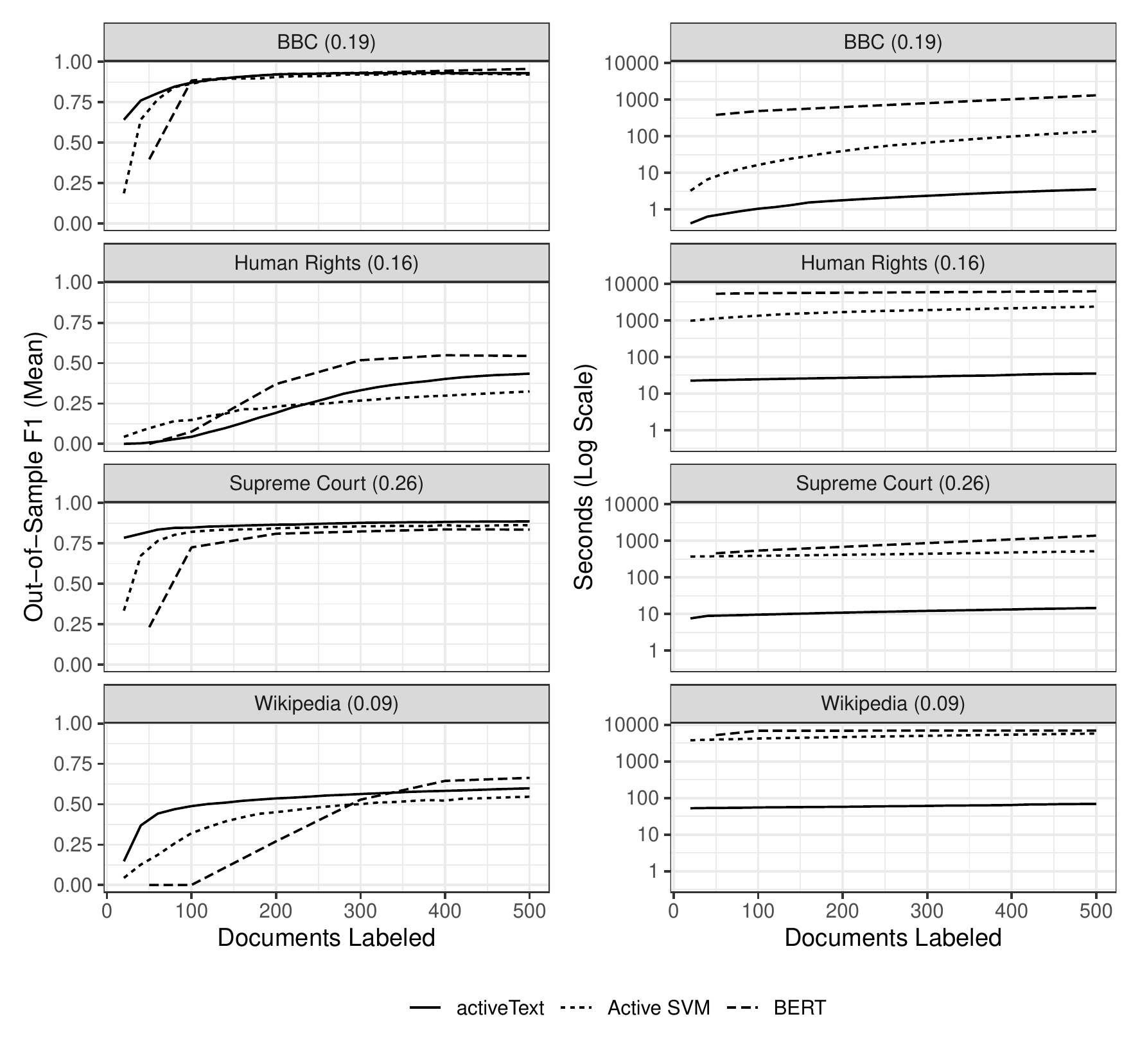}
  \caption{\textbf{Comparison of Classification and Time Results across \aT, Active SVM, and BERT}}
  \label{fig:main_bert}
\end{figure}

In Figure \ref{fig:main_bert}, we compare both classification performance and computational time for \aT, Active SVM, and BERT, a state-of-the-art text classification model.\footnote{For a technical overview of BERT, and the Transformers technology underpinning it, see \citet{devlin2018bert} and \citet{vaswani2017attention}, respectively.}
We trained two sets of models for the F1 and time comparisons, respectively.
The left-hand column of panels shows F1 (the y-axis) as a function of the number of documents labeled (the x-axis), as with the results shown in Figure~\ref{fig:main}.
We trained models using 50 random initializations for the \aT\ and Active SVM models.
We trained the BERT models using 10 random initializations using V100 GPUs on a cluster computing platform.

The F1 comparison in the left-hand column of Figure~\ref{fig:main_bert} shows that for all four of our corpora, \aT\ performs favorably in comparison to our off-the-shelf implementation of the BERT language model.
We show that with each of the BBC, Supreme Court, and Wikipedia corpora (the first, third, and fourth rows of panels), we significantly outperform BERT when there are very few documents labeled.
As the number of labeled documents increases, BERT as expected performs well and even exceeds the F1 score of \aT\ in the case of Wikipedia.
And as shown in the results for the Human Rights corpus (the second row of panels), BERT does outperform \aT\ at all levels of documents labeled.

The right-hand column of panels in Figure~\ref{fig:main_bert} shows computational time, rather than F1, as a function of documents labeled.
For this analysis, our goal was to compare how long it would take a researcher without access to a cluster computing platform or a high-powered GPU to train these models.
To this end, we re-trained the \aT, Active SVM, and BERT models on a base model M1 Macbook Air with 8 GB of RAM and 7 GPU cores.
While the Active SVM and \aT\ models were trained using a single CPU, we used the recent implementation of support for the GPU in M1 Macs in PyTorch\footnote{See \href{}{https://pytorch.org/blog/introducing-accelerated-pytorch-training-on-mac/}.} to parallelize the training of the BERT model using the M1 Mac's GPU cores.\footnote{Specifically, we trained a \textit{DistilBERT} model (see \citet{sanh2019distilbert}) for three epochs (the number of passes of the entire training dataset BERT has completed) using the default configuration from the Transformers and PyTorch libraries for the Python programming language and used the trained model to predict the labels for the remaining documents for each corpus.}
We also computed the time values \textit{cumulatively} for the \aT\ and Active SVM models, since it is expected that model will be fit over and over again as part of the active learning process, whereas for a model like BERT we expect that the model would only be run once, and as such do not calculate its run-time cumulatively.
For the Human Rights and Wikipedia corpora, which each have several hundred thousand entries, we used a random subsample of 50,000 documents. For the Supreme Court and BBC corpora, we used the full samples. Finally, we present the time results in logarithmic scale to improve visual interpretation.

The right-hand panel of Figure~\ref{fig:main_bert} shows that the slight advantages of the BERT models come at a cost of several orders of magnitude of computation time. Using the Wikipedia corpus as an example, at 500 documents labeled the baseline \aT\ would have run to convergence 25 times, and the sum total of that computation time would have amounted to just under 100 seconds. With BERT, however, training a model with 500 documents and labeling the remaining 45,500 on an average personal computer would take approximately 10,000 seconds (2.78 hours).

\subsection{Benefits of Keyword Upweighting}
\label{subsec:keyword_results}
%We also test the effect of implementing the keyword upweighting scheme described above.
%%in \nameref{subsubsec:keywords}.
In Figure \ref{fig:main}, active learning did not improve the performance on the human rights corpus, and the F1 score was lower than other corpora in general. 
One reason for the early poor performance of \aT\ may be length of documents.
Because each document of the human rights corpus consists of one sentence only, the average length is shorter than other corpora.\footnote{With the population data, the average length of each document is 121 (BBC), 17 (Wikipedia), 1620 (Supreme Court), and 9 (Human Rights)} 
%Recall that the predicted probability of the classes is determined by the difference in the distribution of words in documents with positive and negative labels.
This means that the information the models can learn from labeled documents is less compared to the other corpora with longer documents. 
In situations like this, providing keywords in addition to document labels can improve classification performance because it directly shifts the values of the word-class probability matrix, $\boldsymbol{\eta}$, even when the provided keywords is not in the already labeled documents.

\begin{figure}[t!]
  \includegraphics[width=\linewidth]{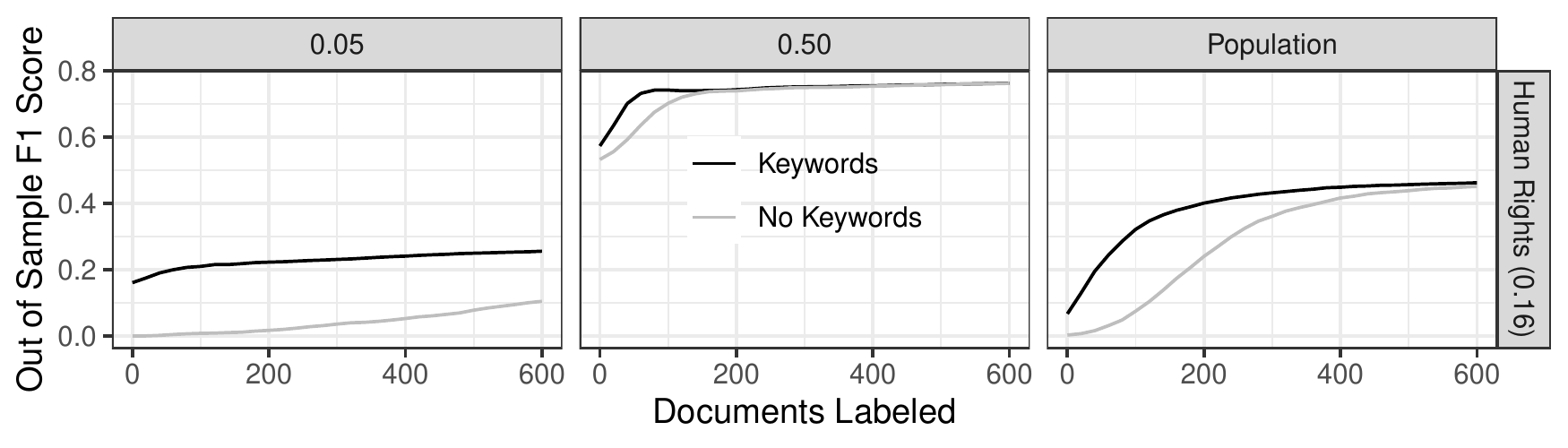}
  \caption{\textbf{Classification Results of \aT\ with and without Keywords}
  %The darker lines show the results with keywords and the lighter lines without.
  %The classification methods is \textit{Active Mixture} with 2 latent clusters for both lines. 
  %The columns correspond to various proportion of positive labels in the corpus.
  %The y-axis indicates the out-of-sample F1 score and the x-axis show the number of sampling steps. 
  %Providing keywords improves classification performance with the human rights dataset.
  }
  \label{fig:keywords}
\end{figure}

Figure \ref{fig:keywords} compares the performance with and without providing keywords. 
The darker lines show the results with keywords and the lighter lines without.
The columns specify the proportion of documents associated with the class of interests: 5\%, 50\% and the population proportion (16\%). 
As in the previous exercises, 20 documents are labeled at each sampling step, and 100 Monte Carlo simulations are performed to stabilize the randomness due to the initial set of documents to be labeled. 
We simulated the process of a user starting with no keywords for either class, and then being queried with extreme words indexed by $v$ whose $\eta_{vk}/\eta_{vk'}$ is the highest for each class $k$, with up to 10 keywords for each class being chosen based on the estimated $\boldsymbol{\eta}$ at a given iteration of the active process.
To determine whether a candidate keyword should be added to the list of keywords or not, our simulated user checked whether the word under consideration was among the set of most extreme words in the distribution of the `true' $\boldsymbol{\eta}$ parameter, which we previously estimated by fitting our mixture model with the complete set of labeled documents.\footnote{Specifically, the simulated user checked whether the word in question was in the top 10\% of most extreme words for each class using the `true' $\boldsymbol{\eta}$ parameter. If the candidate word was in the set of `true' extreme words, it was added to the list of keywords and upweighted accordingly in the next active iteration.}
% Keywords are selected by first obtaining the estimated $\boldsymbol{\eta}$ using the fully labeled dataset, and then selecting \textbf{How many} words, indexed by $v$, whose $\eta_{vk}$ is the highest for each $k$.

The results suggest that providing keywords improves the performance when the proportion of documents is markedly imbalanced across classes.
The keywords scheme improved the performance when the number of labeled documents is smaller on the corpus with 5\% or 16\% (population) labels associated 
with the class of interest. By contrast, it did not on the corpus where both classes were evenly balanced.
These results highlight that our active keyword approach benefits the most when the dataset suffers from serious class-imbalance problems.\footnote{SI~\ref{si-sec:keywords_visual} demonstrates how active keyword works by visualizing the word-class matrix, $\boldsymbol{\eta}$, at each active iteration.}

% THE FOLLOWING PARAGRAPH WAS MOVED TO APPENDIX
%Figure \ref{fig:keywords_eta} illustrates how the word-class matrix $\boldsymbol{\eta}$ is updated with and without keywords across iterations.
%A subset of the keywords supplied are labeled and highlighted by black dots. 
%The x-axis shows the log of $\eta_{v1} / \eta_{v0}$, where $\eta_{v1}$ corresponds the probability of observing the word $v$ in a document with a positive label and $\eta_{v0}$ for a document with a negative label. 
%The high value in x-axis means that a word is more strongly associated with positive labels.
%The y-axis is the log of word frequency.
%A word with high word frequency has more influence in shifting the label probability. 
%In our mixture model, words that appear often and whose ratio of $\eta_{vk^*}$ vs $\eta_{vk}$ is high play a central role in the label prediction.
%By shifting the value of $\boldsymbol{\eta}$ of those keywords, we can accelerate the estimation of $\boldsymbol{\eta}$ and improve the classification performance. 

One caveat is that we provided `true' keywords, in the sense that we used the estimated $\boldsymbol{\eta}$ from a fully labeled dataset.
In practice, researchers have to decide if candidate keywords are indeed keywords using their substantive knowledge.
In this exercise, we believe that the keywords supplied to our simulation are what researchers with substantive knowledge 
about physical integrity rights can confidently adjudicate. 
For example, the keywords, such as ``torture,'' ``beat,'' and ``murder,'' match our substantive understanding of physical integrity right violation.
Nevertheless, humans can make mistakes, and some words may be difficult to judge. 
Thus, we examined the classification performance with varying degrees in the amount of error at the keyword labeling step.  
In SI~\ref{si-subsec:mislabel-keywords}, we show that the active keyword approach still improves the classification performance compared to the no-keyword approach -- even in the presence of small amounts (less than 20\%) of ``honest'' (random) measurement error in keyword labeling.

%% file: inputs/reanalysis.tex
\section{Reanalysis with Fewer Human Annotations}\label{sec:reanalysis}

To further illustrate our proposed approach for text classification, in this section, we reanalyze the results in \cite{gohdes2020repression} and \cite{park:etal:2020}.
We show that via \aT, we arrive at the same substantive conclusions advanced by these authors but using only a small fraction of the labeled data they originally used.

\subsection{Internet Accessibility and State Violence \citep{gohdes2020repression}}\label{subsec:gohdes}

In the article ``Repression Technology: Internet Accessibility and State Violence,'' \cite{gohdes2020repression} argues that higher levels of Internet accessibility are associated with increases in targeted repression by the state. The rationale behind this hypothesis is that through the rapid expansion of the Internet, governments have been able to improve their digital surveillance tools and target more accurately those in the opposition. Thus, even when digital censorship is commonly used to diminish the opposition's capabilities, \cite{gohdes2020repression} claims that digital surveillance remains a powerful tool, especially in areas where the regime is not fully in control.

To measure the extent to which killings result from government targeting operations, \cite{gohdes2020repression} collects 65,274 reports related to lethal violence in Syria. These reports contain detailed information about the person killed, date, location, and cause of death. The period under study goes from June 2013 to April 2015.  Among all the reports, 2,346 were hand-coded by Gohdes, and each hand-coded report can fall under one of three classes: 1) government-targeted killing, 2) government-untargeted killing, and 3) non-government killing. Using a document-feature matrix (based on the text of the reports) and the labels of the hand-coded reports, \cite{gohdes2020repression} trained and tested a state-of-the-art supervised decision tree algorithm (extreme gradient boosting, \texttt{xgboost}). Using the parameters learned at the training stage, \cite{gohdes2020repression} predicts the labels for the remaining reports for which the hand-coded labels are not available. For each one of the 14 Syrian governorates (the second largest administrative unit in Syria), \cite{gohdes2020repression} calculates the proportion of biweekly government targeted killings. In other words, she collapses the predictions from the classification stage at the governorate-biweekly level.

We replicate \cite{gohdes2020repression} classification tasks using \aT. In terms of data preparation, we adhere to the very same decisions made by \cite{gohdes2020repression}. To do so, we use the same 2,346 hand-labeled reports (1,028 referred to untargeted killing, 705 to a targeted killing, and 613 a non-government killing) of which 80\% were reserved for training and 20\% to assess classification performance. In addition, we use the same document-feature matrices.\footnote{\cite{gohdes2020repression} removed stopwords, punctuation, and words that appear in at most two reports, resulting in 1,342 features and a document-feature matrix that is 99\% sparse. The median number of words across documents is 13.}  As noted in \nameref{subsec:active}, because \aT\ selects (at random) a small number of documents to be hand-labeled to initialize the process, we conduct 100 Monte Carlo simulations and present the average performance across initializations. As in \nameref{sec:performance}, we set $\lambda = 0.001$. The performance of \aT\ and \texttt{xgboost} is evaluated in terms of out-of-sample F1 score. Following the discussion in \nameref{subsec:active-learning}, we stopped the active labeling process at the 30th iteration when the out-of-sample F1 score stopped increasing by more than 0.01 units (our pre-specified threshold). Table \ref{tbl:syria} presents the results\footnote{The values in the bottom row are based on \cite{gohdes2020repression}, Table A9.}. Overall, we find that as the number of active learning steps increases, the classification performance of \aT\ is similar to the one in \cite{gohdes2020repression}. However, the number of hand-labeled documents that are required by \aT\ is significantly smaller (around one-third) if compared to the ones used by \cite{gohdes2020repression}.

\begin{table}[h!]
	\centering
	\caption{Classification Performance: Comparison with \cite{gohdes2020repression} results}
	\label{tbl:syria}
	\footnotesize
	\begin{tabular}{l l l *{9}{c}}\\
 	                       &             &       & \multicolumn{3}{c}{Ouf-of-sample F1 Score per class}  \\
    \cmidrule(lr){4-6}
    Model                & Step        & Labels& {Untargeted} & {Targeted} & {Non-Government} \\
    \midrule
    \aT                  & 0           &  20   &   0.715 &  0.521 &  0.800 \\
			 & 10          &  220  &   0.846 &  0.794 &  0.938 \\
                         & 20          &  420  &   0.867 &  0.828 &  0.963 \\
                         & \textbf{30} &  \textbf{620}  &   \textbf{0.876} &  \textbf{0.842} &  \textbf{0.963} \\
                         & 40          &  820  &   0.879 &  0.845 &  0.961 \\
		\midrule
    \cite{gohdes2020repression}         &             & 1876  &   0.910 &  0.890 & 0.940 \\
	\end{tabular}
\end{table}

In social science research, oftentimes, text classification is not the end goal but a means to quantify 
a concept that is difficult to measure and make inferences about the relationship between
this concept and other constructs of interest. In that sense,
to empirically test her claims, \cite{gohdes2020repression}  conducts regression analyses 
where the proportion of biweekly government targeted killings is the dependent variable and Internet accessibility is the main 
independent variable -- both covariates are measured at the governorate-biweekly level. \cite{gohdes2020repression} finds that 
there is a positive and statistically significant relationship between Internet access and the proportion of targeted killings by the
Syrian government. Using the predictions from \aT, we construct the main dependent variable and 
replicate the main regression analyses in \cite{gohdes2020repression}.

\begin{figure}[t!]
  \centering
  \includegraphics[width=0.8\textwidth]{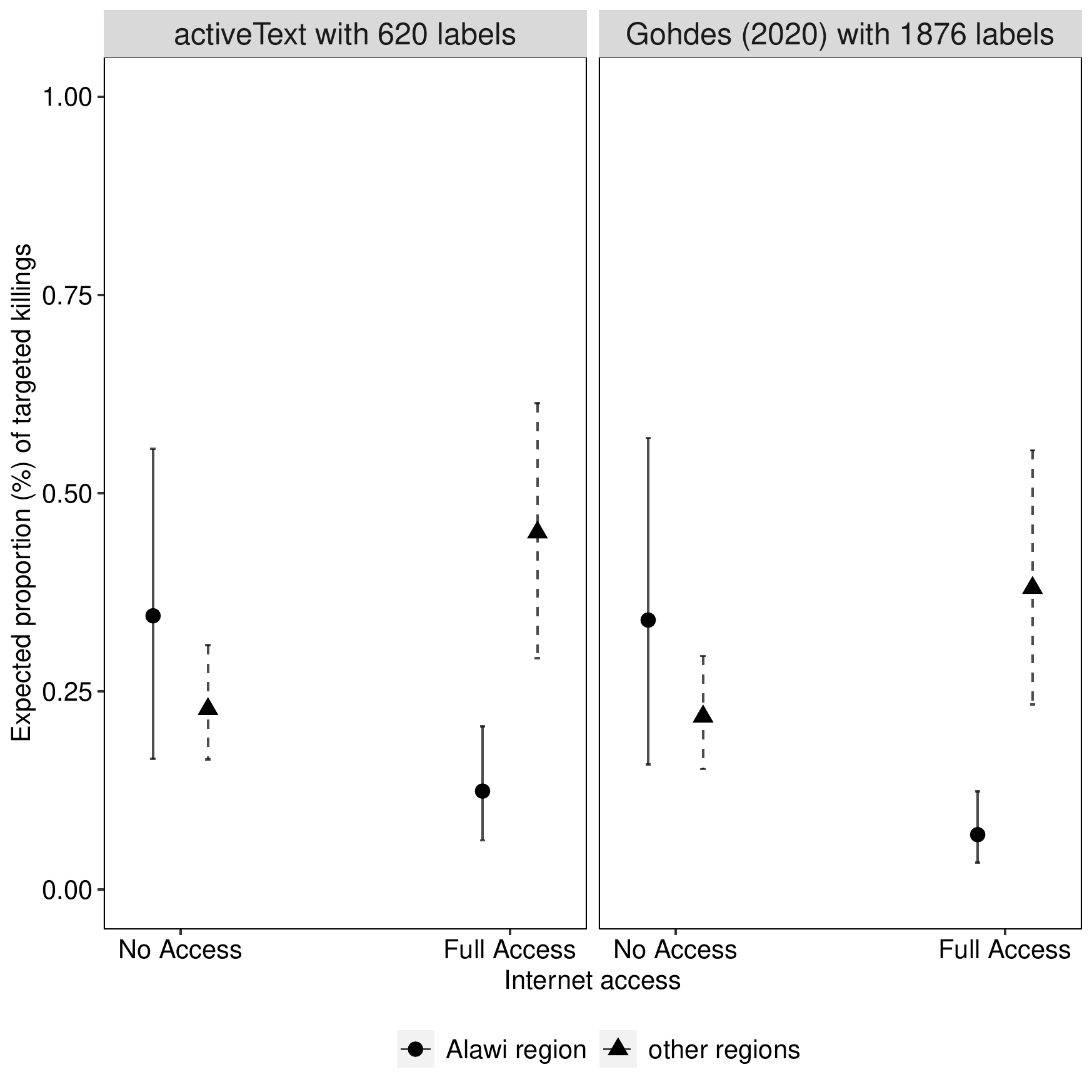}
  \caption{\textbf{Replication of Figure 3 in \cite{gohdes2020repression}: Expected Proportion of Target Killings, Given
  Internet Accessibility and Whether a Region is Inhabitated by the Alawi Minority.} The results from \aT\ are presented
  in the left panel and those of  \cite{gohdes2020repression} are on the right. 
  }
  \label{fig:gohdes_fig3}
\end{figure}

Tables in SI~\ref{si-sec:syria_regression} reports the estimated coefficients, across the
same model specifications in \cite{gohdes2020repression}. The point estimates and the standard errors 
are almost identical whether we use \texttt{xgboost} or \aT. 
Moreover, Figure~\ref{fig:gohdes_fig3} presents the expected proportion of targeted killings by region and Internet accessibility. 
\cite{gohdes2020repression} finds that in the Alawi region (known to be loyal to the regime) when Internet access is
at its highest, the expected proportion of targeted killings is significantly smaller compared to other regions of Syria. 
In the absence of the Internet, however, there is no discernible difference across regions (see Figure~\ref{fig:gohdes_fig3}, right panel).
Our reanalysis does not change the substantive conclusions by \cite{gohdes2020repression} (Figure~\ref{fig:gohdes_fig3}, left panel),
however, it comes just at a fraction of the labeling efforts (labeling 620 instead of 1876 reports). 
As noted above, these gains come from our active sampling scheme as it can select the most informative
 documents to be labeled. 

%% Checked until here.
%%What is the takeaway from these results? 
%%The main implications is that \aT does not change the substantive conclusion by \cite{gohdes2020repression}, but the authors could have reached the same conclusion with much fewer hand labels with our methods.  
%%In the main regression, the estimated coefficients are almost equivalent to the original results.  
%%However, the number of hand-label documents provided is significantly smaller with \aT compared to the original classification with xgboost. 
%%Therefore,
%%While our method did not change the substantive implications of \cite{gohdes2020repression}, our methods allow researchers to reach the same conclusion with less labeling cost. 

%%SK: removed the regression table and put the full tables in Appendix

\subsection{Human Rights are Increasingly Plural \citep{park:etal:2020}}

The question that drives the work of \cite{park:etal:2020} is as follows: how the rapid growth (in the last four decades) of information communication technologies (ICTs) has changed the composition of texts referring to human rights? \cite{park:etal:2020}  make the observation that the average sentiment with which human rights reports are written has not drastically changed over time. Therefore, \cite{park:etal:2020} advance the argument that if one wants to really understand the effect of changes in the access to information on the composition of human rights reports, it is necessary to internalize the fact that human rights are plural (bundles of related concepts). In other words, the authors argue that having access to new information has indeed changed the taxonomy of human rights over time, even when the tone has not. 

To empirically test such a proposition, \cite{park:etal:2020} conduct a two-step approach. First, via an SVM for text classification with three classes (negative, neutral, and positive sentiment), the authors show that the average sentiment of human rights reports has indeed remained stable even in periods where the amount of information available has become larger.\footnote{As explained in Appendix A1 of \cite{park:etal:2020}, negative sentiment refers to text about a clear ineffectiveness in protecting or to violations of human rights;  positive sentiment refers to text about clear support (or no restrictions) of human rights; and neutral sentiment, refers to stating a simple fact about human rights.} Second, they use a network modeling approach to show that while the average sentiment of these reports has remained constant over time, the taxonomy has drastically changed. In this section, using \aT, we focus on replicating the text classification task of \cite{park:etal:2020} (which is key to motivating their puzzle). 

As in the replication of \cite{gohdes2020repression}, we adhere to the same pre-processing decisions made by \cite{park:etal:2020} when working with their corpus of Country Reports on Human Rights Practices from 1977 to 2016 by the US Department of State. In particular, we use the same 4000 hand-labeled human rights reports (1182 are positive, 1743 are negative, and 1075 are neutral) and use the same document-feature matrices (which contain 30,000 features, a combination of unigrams and bigrams). Again, we conduct 100 Monte Carlo simulations and present the average performance across initializations. We stopped the active labeling process at the 25th iteration of our algorithm as the out-of-sample F1 score (from an 80/20 training/test split) does not increase by more than 0.01 units (see Figure~\ref{si-fig:colaresi_fig2} in SI~\ref{si-sec:colaresi}).\footnote{The only point where we depart from \cite{park:etal:2020} is that we use an 80/20 split for training/testing, while they use $k$-fold cross-validation. Conducting $k$-fold cross-validation for an active learning algorithm would require over-labeling and it would be computationally more expensive (the process should be repeated $k$ times). Because of this difference we refrain from comparing our model performance metrics to theirs.}  Using the results from the classification task via \aT, the sentiment scores of 2,473,874 documents are predicted. With those predictions, we explore the evolution of the average sentiment of human rights reports per average information density score.\footnote{Information density is a proxy for ICTs based on a variety of indicators related to the expansion of communications and access to information, see Appendix B in \citet{park:etal:2020}.} 

Figure~\ref{fig:colaresi_fig1} shows that by labeling only 500 documents with \aT, instead of 4000 labeled documents used by \citet{park:etal:2020} to fit their SVM classifier, we arrive at the same substantive conclusion: the average sentiment of human rights reports has remained stable and almost neutral over time. In Figure~\ref{si-fig:colaresi_fig3} of SI~\ref{si-sec:colaresi}, we also show that this result is not an artifact of our stopping rule and it is robust to the inclusion of additional label documents (e.g, labeling 1000, 1500, and 2000 documents instead of just 500).

\begin{figure}[t!]
  \centering
  \includegraphics[width=0.8\textwidth]{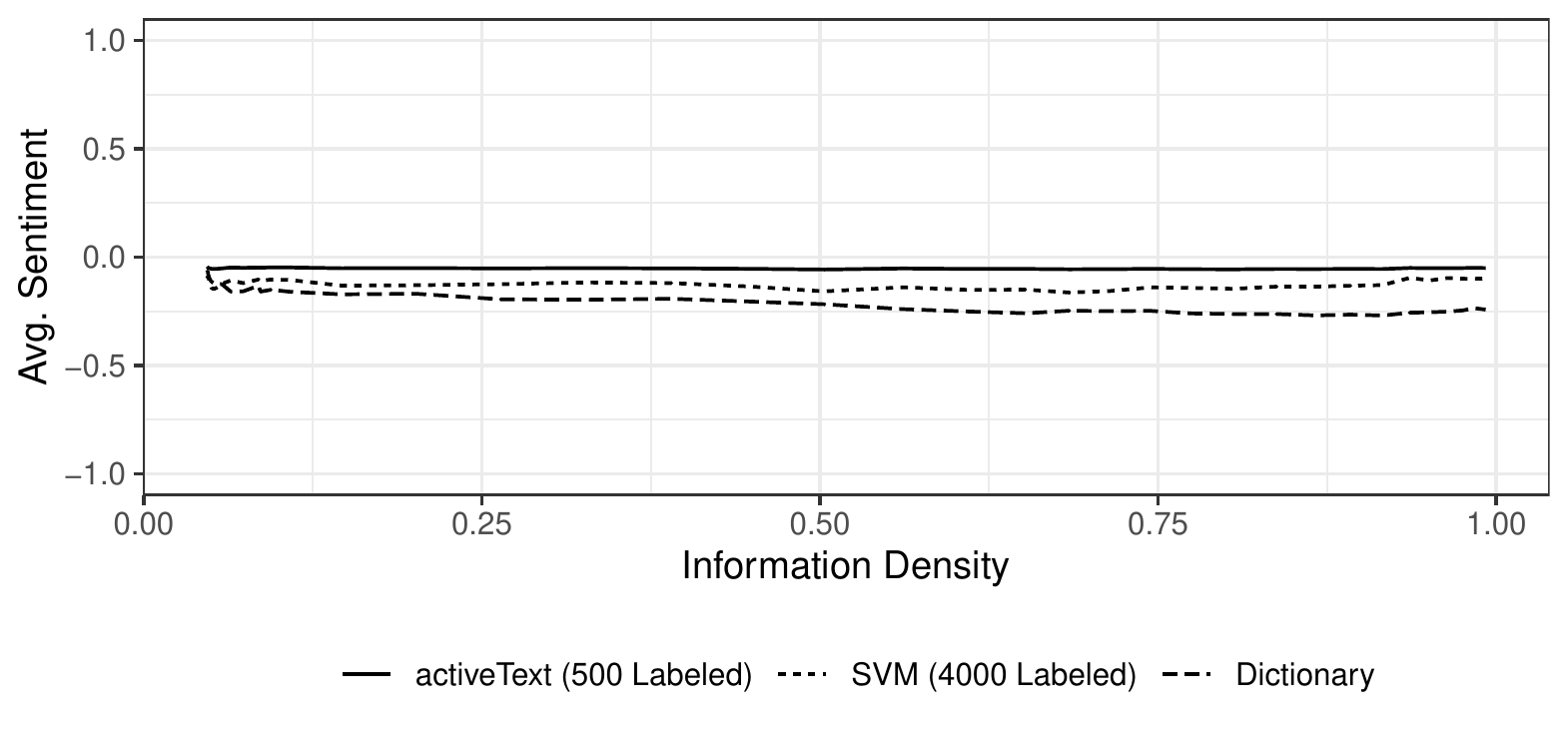}
  \caption{\textbf{Replication of Figure 1 in \cite{park:etal:2020}: The Relationship Between Information Density and Average Sentiment Score.}}
  \label{fig:colaresi_fig1}
\end{figure}

%%\begin{figure}[t!]
%%  \centering
%%  \includegraphics[width=0.8\textwidth]{figs/colaresi_fig2.pdf}
%%  \caption{\textbf{Using the Difference in Out of Sample F1 Score to Decide a Stopping Point.}}
%%  \label{fig:colaresi_fig2}
%%\end{figure}

% \begin{figure}[t!]
%   \centering
%   \includegraphics[width=0.8\textwidth]{inputs/figs/colaresi_fig1.pdf}
%   \caption{\textbf{Replication of Figure X in \cite{park:etal:2020}: The Relationship Between Information Density and Average Sentiment Score.}}
%   \label{fig:colaresi_fig1}
% \end{figure}

%% file: inputs/discussion.tex
\section{Discussion}
\label{sec:discussion}

% In this section, we will discuss several practical concerns and possible improvement of the model. 

\subsection{Tuning the value of $\lambda$}
%%The first practical question is how to choose the value of $\lambda$,  which is the weight of the unlabeled documents relative to the labeled documents.
As noted above, we downweight the information from unlabeled documents as we typically have more unlabeled than labeled documents. Moreover, since the labeled documents have been classified by an expert, we want to rely more on the information they bring for prediction.

An important practical consideration is: how to select the value of $\lambda$ that maximizes the performance.
One possible approach would be to adopt popular model selection methods (e.g. cross-validation) to choose the appropriate $\lambda$ value during the model initialization process.\footnote{Indeed, it may be beneficial to tune the lambda value \textit{across} active learning iterations.}
However, cross-validation may not be practical when the labeled data is scarce (or absent at the beginning of the process). Using our active learning approach is particularly,
%%While we do not yet have a formal way of choosing the appropriate value of $\lambda$, 
we have observed across a variety of applications that very small values (e.g., 0.001 or 0.01) work the best on the corpora we used (see SI~\ref{si-sec_si:add_results}). However, more work is needed to clearly understand the optimality criteria needed to select $\lambda$. We leave this question for future research.

% \subsection{Choosing the number of clusters}
% Our mixture model assumes that there are fixed number of latent clusters, and we showed that choosing the correct number of clusters can improve the prediction performance (see Figure \ref{fig:cluster}).
% A practical question is how to choose the number of latent clusters.
% One solution is to rely on the substantive knowledge about the dataset.
% In the case of BBC news article, we assume that users know that there are 5 news categories and the goal is to predict if a document falls into one of the categories.
% An alternative data-driven solution is to fit unsupervised clustering methods (e.g. $k$-means or a simple mixture model without supervision) and select the number of clusters via some measures such as AIC or BIC before providing any labels.
% Future research could adopt a non-parametric methods (e.g. Dirichlet Process) to identify the number of clusters from the data.
% However, the choice of the number of clusters does not matter too much, unless a corpus has a strong grouping structure and a user knows the structure.
% At worst, our simulations show that increasing the number of cluster does not worsen the classification performance compared to the baseline 2-cluster model (see Appendix Figure \ref{fig:fig_cluster_keywords}).

\subsection{Labeling Error}
While our empirical applications assume that labelers are correct, human labelers do make mistakes.
In SI~\ref{si-sec:mislabel}, we examine how mislabeling keywords and documents affect classification performance.
Our results show that, if compared to the no-keyword approach, a small amount of random noise (classical measurement error) on keyword labeling does not hurt the classification performance. In contrast, random perturbations from true document labels do hurt the classification performance.
%Future research can address this point by comparing the robustness to labelling error between supervised vs semi-supervised classifiers.
A promising avenue for future research should center on developing new active learning algorithms that assign labelers based on their labeling ability and/or are robust to more pervasive forms of labeling error (differential and non-differential measurement error).
For instance, assigning the most competent labelers with the most uncertain or difficult documents and assigning the least competent labelers with easier documents can optimize the workload of the labelers.
At the same time, we note that users may be able to improve the quality of human labeling by other means, such as polishing cateogry concepts and better training of coders, in practical settings.

%\subsection{Difference with Topic model}
%\begin{itemize}
%\item Topic model assigns a latent variable to a word while our model assigns it to a document 
%\item Better when we want to classify documents because classification tasks assumes one label/topic with a document. 
%\item Topic model assumes that a document is a mixture of topic/label. Unnecessary compliaction.
%\end{itemize}
%\subsection{Multiple clusters for the positive class}
%Our model assume an asymmetric mapping between classification classes and latent clusters.
%For the positive class, there is only one corresponding latent cluster; for the negative class, multiple latent clusters can be linked.
%This assumption is made because, in practice, users might be interested in one of the latent categories in the documents (e.g. In the BBC news article application, we chose politics category to be the positive class while the negative class includes all other categories like business, entertainment, technology, etc.).
%There is no reason to relax this assumption and allow the positive class to be mapped to multiple latent clusters as well.

% Potential Barriers/Challenges: It is difficult to evaluate the quality of a probabilistic model of text classification without ground-truth labels, and such labels do not exist or need to be manually curated (or actively learned). Thus, we will need to manually label documents in order to evaluate the quality of the approach.

%%% Local Variables:
%%% mode: yatex
%%% TeX-master: "../active.tex"
%%% End:

%% file: inputs/concluding.tex
\section{Conclusion}
\label{sec:conclusion}

Human labeling of documents is the most labor-intensive part of social science research that uses text data.
For automated text classification to work, a machine classifier needs to be trained on the relationship between text features and class labels, and the labels in training data are given manually.
In this paper we have described a new active learning algorithm that combines a mixture model and active learning to incorporate information from labeled and unlabeled documents and better select which documents to be labeled by a human coder.
Our validation study showed that the proposed algorithm performed at least as well as state-of-the-art methods such as BERT while reducing computational costs dramatically.
We replicated two published political science studies to show that our algorithm lead to the same conclusions as the original papers but needed much fewer labeled documents.
In sum, our algorithm enables researchers to save their manual labeling efforts without sacrificing quality.

Machine learning techniques are becoming increasingly popular in political science, but the barrier to entry remains too high for researchers without a technical background to make use of advances in the field.
As a result, there is an opportunity to democratize access to these methods.
Towards this, we continue to work towards publishing the R package \textit{activeText} on CRAN.
We believe that our model will provide applied researchers a tool that they can use to efficiently categorize documents in corpuses of varying sizes and topics.

% Potential Barriers/Challenges: It is difficult to evaluate the quality of a probabilistic model of text classification without ground-truth labels, and such labels do not exist or need to be manually curated (or actively learned). Thus, we will need to manually label documents in order to evaluate the quality of the approach.

%%% Local Variables:
%%% mode: yatex
%%% TeX-master: "../active.tex"
%%% End:

%% file: inputs/appendix.tex
%\section{Table of Political Science papers that use text classification}
%\label{subsec:lit_review}
%
%\input{inputs/table_litreview.tex}
%\clearpage

\noindent Note that to facilitate exposition, in the main text, we use the words political and non-political labels to describe 
the problem of binary classification. Without loss of generality, in this supplemental information material, 
we use the positive vs. negative class dichotomy instead. 

\section{Detailed explanations about the EM algorithm to estimate parameters}
\label{subsec:em}

Let $\mathbf{D}^{lp}$, $\mathbf{D}^{ln}$ and $\mathbf{D}^u$ be the document feature matrices for documents 
with positive labels, documents with negative labels, and unlabeled documents, respectively. 
Also let $N^{lp}$, $N^{ln}$, and $N^u$ be the number of documents with positive labels, negative labels, and documents without labels.
Likewise, $\mathbf{C}^{lp}$ and $\mathbf{C}^{ln}$ be the vectors of positive and negative labels. Then, the observed-data likelihood is: 

\begin{equation}
\begin{split}
  &p(\pi, \boldsymbol{\eta} \vert \mathbf{D}, \mathbf{C}^{lp}, \mathbf{C}^{ln}) \\
    &\propto p(\pi) p(\boldsymbol{\eta}) p(\mathbf{D}^{lp}, \mathbf{C}^{lp} \vert \pi, \boldsymbol{\eta}) p(\mathbf{D}^{ln}, \mathbf{C}^{ln} \vert \pi, \boldsymbol{\eta}) \Big[p(\mathbf{D}^u \vert \pi, \boldsymbol{\eta})\Big]^{\lambda} \\
    &= p(\pi) p(\boldsymbol{\eta}) 
    \times \prod_{i=1}^{N^{lp}} p(\mathbf{D}_i^{lp} \vert Z_{i} = 1, \eta) p(Z_{i} = 1 \vert \pi) \times \prod_{i=1}^{N^{ln}} \Big\{ p(\mathbf{D}_i^{ln} \vert Z_{i} = 0, \eta) p(Z_{i} = 0 \vert \pi) \Big\} \\
    &\quad \times \Bigg[\prod_{i=1}^{N^{u}} \Big\{ p(\mathbf{D}_i^{u} \vert Z_{i} = 1, \boldsymbol{\eta}) p(Z_{i} = 1 \vert \pi) + p(\mathbf{D}_i^{u} \vert Z_{i} = 0, \boldsymbol{\eta}) p(Z_{i} =0\vert \pi) \Big\} \Bigg]^{\lambda} \\
    &\propto \underbrace{\big\{(1-\pi)^{\alpha_0 - 1} \prod_{v=1}^V  \eta_{v0}^{\beta_{0v} - 1}\big\} \times \big\{\pi^{\alpha_1 - 1} \prod_{v=1}^V \eta_{v1}^{\beta_{1v} - 1}\big\}}_\text{prior}
    \times 
    \underbrace{\prod_{i=1}^{N^{lp}} \Big\{ \prod_{v=1}^V \eta_{v1}^{D_{iv}}\times \pi \Big\}}_\text{positive labeled doc. likelihood} \\
    &\quad \times \underbrace{\prod_{i=1}^{N^{ln}} \Big\{ \prod_{v=1}^V \eta_{v0}^{D_{iv}}\times (1-\pi) \Big\}}_\text{negative labeled doc. likelihood} 
    \times \underbrace{\Bigg[\prod_{i=1}^{N^{u}} \Big\{ \prod_{v=1}^V \eta_{v0}^{D_{iv}}\times (1-\pi)  \Big\} + \Big\{ \prod_{v=1}^V \eta_{v1}^{D_{iv}}\times \pi  \Big\}\Bigg]^{\lambda}
}_\text{unlabeled doc. likelihood}
\end{split}
\end{equation}
We weigh the part of the observed likelihood that refers to the unlabeled document with $\lambda \in [0, 1]$. 
This is done because we typically have many more unlabeled documents than labeled documents.
By downweighting the information from the unlabeled document (i.e., setting $\lambda$ to be small), 
we can use more reliable information from labeled documents than from unlabeled documents.

We estimate the parameters $\pi$ and $\eta$ using EM algorithm~\cite{dempster1977maximum} and our implementation is presented as pseudocode in Algorithm~\ref{alg:em}.
Note that by taking the expectation of the log complete likelihood function (Q function),
\begin{equation}
\begin{split}
  Q &\equiv \E_{\mathbf{Z} \vert \pi^{(t)}, \boldsymbol{\eta}^{(t)}, D, C}[p(\pi, \boldsymbol{\eta}, \mathbf{Z} \vert \mathbf{D}, \mathbf{C})] \\
    &= (\alpha_0 -1) \log (1 - \pi^{(t)}) + (\alpha_1 - 1) \log \pi^{(t)} + \sum_{v=1}^V \left\{ (\beta_{0v} - 1) \log \eta_{v0}^{(t)} + (\beta_{1v} - 1) \log \eta_{v1}^{(t)} \right\} \\
    &\quad + \sum_{i=1}^{N^{lp}} \Big\{ \sum_{v=1}^V D_{iv} \log \eta_{v1}^{(t)} + \log \pi^{(t)} \Big\} + \sum_{i=1}^{N^{ln}} \Big\{ \sum_{v=1}^V D_{iv} \log \eta_{v0}^{(t)} + \log (1-\pi^{(t)}) \Big\} \\
    &\quad + \lambda \left[ \sum_{i=1}^{N^{u}}  p_{i0} \Big\{ \sum_{v=1}^V D_{iv} \log \eta_{v0}^{(t)} + \log (1-\pi^{(t)}) \Big\} + p_{i1} \Big\{ \sum_{v=1}^V D_{iv} \log \eta_{v1}^{(t)} + \log \pi^{(t)} \Big\}\right] 
\end{split}
\end{equation}
where $p_{ik}$ is the posterior probability of a document $i$ being assigned to the $k$ th cluster, $k = \{0, 1\}$, given data and the parameters at $t$ th iteration.
If a document has a positive label, $p_{i0} = 0$ and $p_{i1} = 1$.
%%If a document has a negative label, $p_{i0} = 1$ and $p_{i1} = 0$. 

\begin{algorithm}[t]
  \SetAlgoLined \KwResult{Maximize
    $p(\pi^{(t)}, \boldsymbol{\eta}^{(t)} \mid \mathbf{D}^l, \mathbf{Z}^l, \mathbf{D}^u, \boldsymbol{\alpha}, \boldsymbol{\beta})$}
  % \textcolor{gray}{[\textbf{Keyword Scheme Only}]: For each class \(c \in C\),
  %   increment the element of \(\beta\) associated with the keyword
  %   \(\kappa^{c}_{v^*}\) by the scalar \(\gamma\).}

  \eIf{In the first iteration of Active learning}{ Initialize $\pi$ and $\boldsymbol{\eta}$
    by Naive Bayes\; \quad $\pi^{(0)} \gets$ NB($\mathbf{D}^l$, $Z^l, \balpha$)\; \quad
    $\boldsymbol{\eta}^{(0)} \gets$ NB($\mathbf{D}^l$, $\mathbf{Z}^l, \bbeta$)\; }{ Inherit $\pi^{(0)}$ and
    $\boldsymbol{\eta}^{(0)}$ from the previous iteration of Active learning\; }

  \While{$p(\pi^{(t)}, \boldsymbol{\eta}^{(t)} \mid \mathbf{D}^l, \mathbf{Z}^l, \mathbf{D}^u, \balpha, \bbeta)$ does not
    converge}{

    (1) E step: obtain the probability of the class for unlabeled documents\;
    \quad $p(\mathbf{Z}^u \mid \pi^{(t)}, \boldsymbol{\eta}^{(t)} \mathbf{D}^l, \mathbf{Z}^l, \mathbf{D}^u) \gets$ E step($\mathbf{D}^u$,
    $\pi^{(t)}$, $\boldsymbol{\eta}^{(t)}$)\;

    (2) Combine the estimated classes for the unlabeled docs and the known
    classes for the labeled docs\; \quad
    $p(\mathbf{Z} \mid \pi^{(t)}, \boldsymbol{\eta}^{(t)}, \mathbf{D}^l, \mathbf{Z}^l, \mathbf{D}^u)\gets$ combine($\mathbf{D}^l$, $\mathbf{D}^u$,
    $Z^l$, $p(Z^u \mid \pi^{(t)}, \boldsymbol{\eta}^{(t)}, \mathbf{D}^l, \mathbf{Z}^l, \mathbf{D}^u)$)\;

    (3) M step: Maximize
    $Q \equiv \mathbb{E}[p(\pi, \boldsymbol{\eta}, \mathbf{Z}^u \mid \mathbf{D}^l, \mathbf{Z}^l, \mathbf{D}^u, \balpha, \bbeta)]$
    w.r.t $\pi$ and $\boldsymbol{\eta}$\; \quad $\pi^{(t+1)} \gets \text{argmax}\ Q$\; \quad
    $\boldsymbol{\eta}^{(t+1)} \gets \text{argmax}\ Q$\;

    (4) Check convergence: Obtain the value of
    $p(\pi^{(t+1)}, \boldsymbol{\eta}^{(t+1)} \mid \mathbf{D}^l, \mathbf{Z}^l, \mathbf{D}^u, \balpha, \bbeta)$\; }
  \caption{EM algorithm to classify text}
  \label{alg:em}
\end{algorithm}

If a document has no label, 
\begin{eqnarray}
%%\begin{split}
\label{eq:pred}
%%    p_{i0} = \frac{\prod_{v=1}^V \eta_{v0}^{D_{iv}} \times (1-\pi)}{\prod_{v=1}^V \left\{\eta_{v0}^{D_{iv}} \times (1-\pi)\right\} + \prod_{v=1}^V \left\{ \eta_{v1}^{D_{iv}} \times \pi \right\} } \\
p_{i0} &=& 1 - p_{i1} \nonumber \\
    p_{i1} &=& \frac{\prod_{v=1}^V \eta_{v1}^{D_{iv}} \times \pi}{ \prod_{v=1}^V \left\{\eta_{v0}^{D_{iv}} \times (1-\pi)\right\} + \prod_{v=1}^V \left\{ \eta_{v1}^{D_{iv}} \times \pi \right\}}
%%\end{split}
\end{eqnarray}

Equation \ref{eq:pred} also works as the prediction equation.
The predicted class of a document $i$ is $k$ that maximizes this posterior probability.

In the M-step, we maximize the Q function,  and obtain the updating equations for $\pi$ and $\eta$.
The updating equation for $\pi$ is the following.
\begin{equation}
\begin{split}
  \pi^{(t+1)} = \frac{\alpha_1 - 1 + N^{lp} + \lambda \sum_{i=1}^{N^u} p_{i1} }{\left(\alpha_1 - 1 + N^{lp} + \lambda \sum_{i=1}^{N^u} p_{i1}\right) +\left(\alpha_0 - 1 + N^{ln} + \lambda \sum_{i=1}^{N^u} p_{i0}\right)}
\end{split}
\end{equation}

The updating equation for $\eta$ is the following.
\begin{equation}
\begin{split}
  \hat{\eta}_{v0}^{(t+1)} &\propto (\beta_{v0} -1) + \sum_{i=1}^{N^{ln}}  D_{iv} + \lambda \sum_{i=1}^{N^{u}} p_{i0}  D_{iv}, \quad v = 1, \ldots, V \\
  \hat{\eta}_{v1}^{(t+1)} &\propto (\beta_{v1} -1) + \sum_{i=1}^{N^{lp}}  D_{iv} + \lambda \sum_{i=1}^{N^{u}} p_{i1}  D_{iv}, \quad v = 1, \ldots, V 
\end{split}
\end{equation}
%%Note that we downweight the information from unlabeled document by $\lambda$, to utilize more reliable information from labeled documents.

\newpage
\section{EM algorithm for binary classification with multiple clusters}
\label{sec:multiple_cluster_model}

\subsection{Summary}

The model outlined above assumes that there are two latent clusters, 
each linked to the positive and the negative class.
However, this assumption can be relaxed to link multiple clusters to the negative class.
In the world of mixture models, the simplest setup is to let $K=2$ since the classification
goal is binary, and we can link each latent cluster to the final classification categories.  
A more general setup is to use $K>2$ even when a goal is a binary classification. 
If $K>2$, but our focus is to uncover the identity of one cluster,  
we can choose one of the latent clusters to be linked to the ``positive'' class 
and let all other latent clusters be linked to the ``negative'' class (see e.g., \citealt{lars:rubi:01} for 
a similar idea in the realm of record linkage).
In other words, we collapse the $K-1$ latent clusters into one class for the classification purpose.     
Using $K>2$ makes sense if the ``negative'' class consists of multiple sub-categories. 
For instance, suppose researchers are interested in classifying news articles into political news or not.  
Then, it is reasonable to assume that the non-political news category consists of multiple sub-categories, such as technology, entertainment, and sports news. 
%%Using the number of clusters $K>2$ may help improve the classification performance in the situations like this.
%%For instance, BBC corpus consists of 5 categories, politics, business, sports, technology, and entertainment, and the classification goal here is to identify documents with the politics category.
%In Appendix~\ref{sec:multiple_cluster_model}, we present such an extension and present evidence that, in some settings, $K > 2$ can result in higher accuracy if compared to $K = 2$.
%\footnote{We have experimented with $K=5$ when we evaluate the model performance with BBC news article. The results showed that using $K=5$ increased the classification performance compared to $K=2$. \textbf{Refer to Appendix?}}

\subsection{Model}
This section presents a model and inference algorithm when we use more than 2 latent clusters
in estimation but the final classification task is binary. In other words, we impose a hierarchy 
where many latent clusters are collapsed into the negative class. In contrast, the positive class
is made out of just one class. The model presented is as follows:
%% in the main paper is a special case of the following model where the number of latent clusters is 2, i.e. $K =2$.

\begin{equation}
\begin{split}
  \pi &\sim Dirichlet(\balpha) \\
    Z_i &\stackrel{i.i.d}{\sim} Categorical(\bpi) \\
    \mathbf{\eta}_{\cdot k} &\stackrel{i.i.d}{\sim} Dirichlet(\boldsymbol{\beta}_k), \quad k = \{1,\ldots, K\}  \\
   \mathbf{D}_{i\cdot} \vert Z_{i} = k  &\stackrel{i.i.d}{\sim} Multinomial(n_i, \boldsymbol{\eta}_{\cdot k}) \\
\end{split}
\end{equation}

Note that $\bpi$ is now a probability vector of length $K$, and it is drawn from a Dirichlet distribution.  

Let $k^*$ be the index of the cluster linked to the positive class.
The observed likelihood is the following.
\begin{equation}
\begin{split}
  &p(\bpi, \boldsymbol{\eta} \vert \mathbf{D}, \mathbf{C}^{lp}, \mathbf{C}^{ln}) \\
    &\propto p(\bpi) p(\boldsymbol{\eta}) p(\mathbf{D}^{lp}, \mathbf{C}^{lp} \vert \bpi, \boldsymbol{\eta}) p(\mathbf{D}^{ln}, \mathbf{C}^{ln} \vert \bpi, \boldsymbol{\eta}) \Big[p(\mathbf{D}^u \vert \bpi, \boldsymbol{\eta})\Big]^{\lambda} \\
    &= p(\bpi) p(\boldsymbol{\eta}) 
    \times \prod_{i=1}^{N^{lp}} p(\mathbf{D}_i^{lp} \vert Z_{i} = k^*, \eta) p(Z_{i} = k^* \vert \bpi) \\
    &\quad \times \prod_{i=1}^{N^{ln}}\sum_{k\neq k^*} \Big\{ p(\mathbf{D}_i^{ln} \vert Z_{i} = k, \eta) p(Z_{i} = k \vert \bpi) \Big\} \times \Bigg[\prod_{i=1}^{N^{u}}\sum_{k=1}^K \Big\{ p(\mathbf{D}_i^{u} \vert Z_{i} = k, \boldsymbol{\eta}) p(Z_{i} = k \vert \bpi)\Big\} \Bigg]^{\lambda} \\
    &\propto \underbrace{\prod_{k=1}^K\left\{\pi_k^{\alpha_k - 1} \prod_{v=1}^V \eta_{vk}^{\beta_{kv} - 1}\right\}}_\text{prior}
    \times 
    \underbrace{\prod_{i=1}^{N^{lp}} \Big\{ \prod_{v=1}^V \eta_{vk^*}^{D_{iv}}\times \pi_k \Big\}}_\text{positive labeled doc. likelihood} \\
    &\quad \times \underbrace{\prod_{i=1}^{N^{ln}}\sum_{k\neq k^*} \Big\{ \prod_{v=1}^V \eta_{vk}^{D_{iv}}\times \pi_k \Big\}}_\text{negative labeled doc. likelihood} 
    \times \underbrace{\Bigg[\prod_{i=1}^{N^{u}} \sum_{k=1}^K \Big\{\prod_{v=1}^V \eta_{vk}^{D_{iv}}\times \pi_k  \Big\}\Bigg]^{\lambda}
}_\text{unlabeled doc. likelihood}
\end{split}
\end{equation}

The Q function (the expectation of the complete log likelihood) is 
\begin{equation}
\begin{split}
  Q &\equiv \E_{\mathbf{Z} \vert \bpi^{(t)}, \boldsymbol{\eta}^{(t)}, D, C}[p(\bpi, \boldsymbol{\eta}, \mathbf{Z} \vert \mathbf{D}, \mathbf{C})] \\
    &= \sum_{k=1}^K \left[(\alpha_k - 1) \log \pi_k^{(t)} + \sum_{v=1}^V \left\{ (\beta_{kv} - 1) \log \eta_{vk}^{(t)} \right\}\right]  \\
    &\quad + \sum_{i=1}^{N^{lp}} \Big\{ \sum_{v=1}^V D_{iv} \log \eta_{vk^*}^{(t)} + \log \pi_{k^*}^{(t)} \Big\} 
    + \sum_{i=1}^{N^{ln}} \sum_{k\neq k^*} p_{ik}\Big\{ \sum_{v=1}^V D_{iv} \log \eta_{vk}^{(t)} + \log \pi_k^{(t)} \Big\} \\
    &\quad + \lambda \left[ \sum_{i=1}^{N^{u}}\sum_{k=1}^K  p_{ik} \Big\{ \sum_{v=1}^V D_{iv} \log \eta_{vk}^{(t)} + \log \pi_k^{(t)} \Big\}\right] 
\end{split}
\end{equation}

The posterior probability of $Z_i = k$, $p_{ik}$, is 
\begin{equation}
\begin{split}
  p_{ik} &= \frac{\prod_{v=1}^V \eta_{vk}^{D_{iv}} \times \pi_k}{\sum_{k=1}^K \left[\prod_{v=1}^V \eta_{vk}^{D_{iv}} \times \pi_k \right]}
\end{split}
\end{equation}

M step estimators are
The updating equation for $\pi$ is the following.
\begin{equation}
\begin{split}
  \hat{\pi_k} \propto
  \begin{cases}
    \alpha_k - 1 +  \sum_{i=1}^{N^{ln}} p_{ik} + \lambda \sum_{i=1}^{N^u} p_{ik} &\text{if}\ k \neq k^*\\
    \alpha_k - 1 + N^{lp} + \lambda \sum_{i=1}^{N^u} p_{ik^*} &\text{if}\ k = k^*
  \end{cases}  
\end{split}
\end{equation}

The updating equation for $\eta$ is the following.
\begin{equation}
\begin{split}
\hat{\eta}_{vk} \propto 
    \begin{cases}
      (\beta_k -1) + \sum_{i=1}^{N^{ln}} p_{ik}  D_{iv} +
      \lambda \sum_{i=1}^{N^{u}} p_{ik}  D_{iv} & \text{if}\ k \neq k^* \\
      (\beta_k -1) + \sum_{i=1}^{N^{lp}}  D_{iv} +
      \lambda \sum_{i=1}^{N^{u}} p_{ik^*} D_{iv} & \text{if}\ k = k^* 
    \end{cases}
\end{split}
\end{equation}
Note that we downweight the information from the unlabeled documents by $\lambda$, to utilize more reliable information from labeled documents.

\subsection{Results}
\label{subsec:add_res}

\begin{figure}[t!]
  \includegraphics[width=\linewidth]{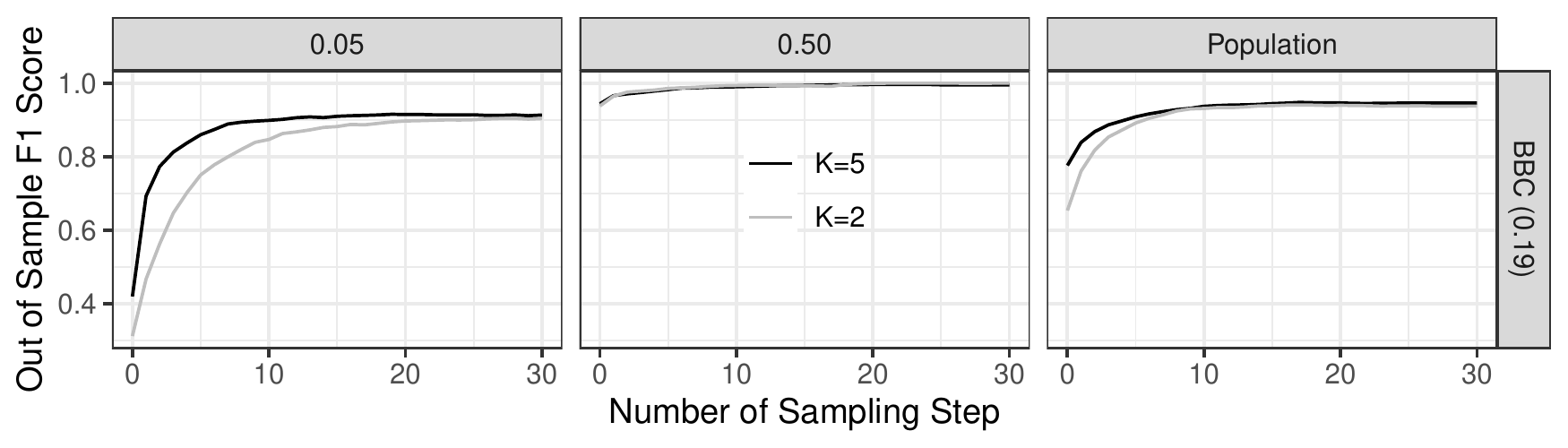}
  \caption{\textbf{Classification Results with 2 and 5 Clusters.}
  The darker lines show the results with 5 latent clusters and the lighter lines show 2 latent clusters.
  The columns correspond to various proportions of positive labels in the corpus.
  The y-axis indicates the out-of-sample F1 score and the x-axis show the number of sampling steps.
  Using multiple clusters improves the classification performance when the number of latent clusters matches the data generating process.
  }
  \label{fig:cluster}
\end{figure}

Figure \ref{fig:cluster} shows the results of a model with just two latent clusters vs. a model with 5 latent clusters but 
only two final classes (positive vs. negative).
The darker lines show the results with 5 latent clusters and the lighter lines show the results with 2 latent clusters.
Overall, the model with 5 clusters performs better or as well as the model with 2 clusters.
The gain from using 5 clusters is the highest when the proportion of positive labels is small and when the size of labeled data is small.

\begin{figure}[p!] 
  \centering
  \includegraphics[scale=0.85]{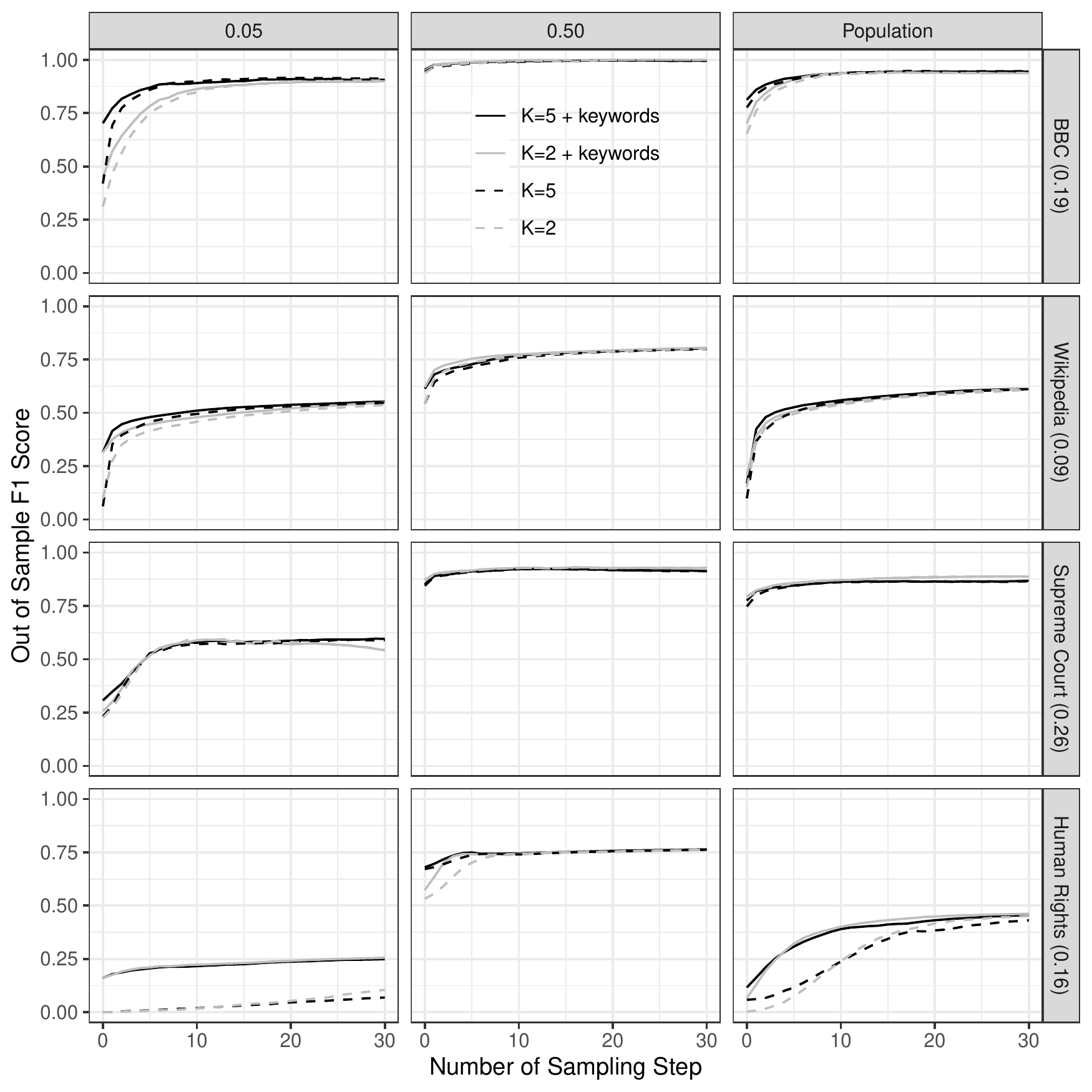}
  \caption{\textbf{Classification Results with Multiple Clusters and Keywords.}
  The rows correspond to different datasets and the columns correspond to various proportions of positively labeled documents in the corpus.
  The y-axis indicates the out-of-sample F1 score and the x-axis show the number of sampling steps. 
  The linetype show whether keywords are supplied: the solid lines show the results with keywords and the dashed lines without keywords.
  The colors show the number of latent clusters in the mixture model: the darker lines show the results with 5 latent clusters and the lighter lines with 2 latent clusters. 
  Using 5 clusters leads to as good or slightly better performance than using 2 clusters.
  The performance improvement is the largest with the BBC corpus, which consists of 5 news topic categories. 
  Likewise, our mixture models with keywords leads to as good or better performance than the models without keywords.
  The improvement is the largest with the human rights corpus, where the number of words per document is the smallest.
  }
  \label{fig:fig_cluster_keywords}
\end{figure}

Figure~\ref{fig:fig_cluster_keywords} shows the results when the multiple cluster approach and keyword upweighting approaches are combined.

\clearpage
\section{Multiclass Classification}
\label{sec:multiple_class}

\subsection{Model}
This section presents a model and inference algorithm for multiclass classification. 
Let $K$ be the number of the clusters and is equal to the number of classes to be classified,
with $K \geq 2$. Differently than in SI~\ref{sec:multiple_cluster_model}, we do not impose any hierarchies and 
the model is a true multi-class mixture model, where the end goal is to classify documents
in $K \geq 2$ classes. In other words, the model presented below is a generalization
of the model presented in the main text. 
%%The generative model is identical to the model for binary classification with multiple latent clusters. 

\begin{equation}
\begin{split}
  \pi &\sim Dirichlet(\balpha) \\
    Z_i &\stackrel{i.i.d}{\sim} Categorical(\bpi) \\
    \mathbf{\eta}_{\cdot k} &\stackrel{i.i.d}{\sim} Dirichlet(\boldsymbol{\beta}_k), \quad k = \{1,\ldots, K\}  \\
   \mathbf{D}_{i\cdot} \vert Z_{i} = k  &\stackrel{i.i.d}{\sim} Multinomial(n_i, \boldsymbol{\eta}_{\cdot k}) \\
\end{split}
\end{equation}

Note that $\bpi$ is now a probability vector of length $K$, and it is drawn from a Dirichlet distribution.  

The observed likelihood is the following.
\begin{equation}
\begin{split}
  p(\bpi, \boldsymbol{\eta} \vert \mathbf{D}, \mathbf{C}^l) 
    &\propto p(\bpi) p(\boldsymbol{\eta}) p(\mathbf{D}, \mathbf{C} \vert \bpi, \boldsymbol{\eta}) \Big[p(\mathbf{D}^u \vert \bpi, \boldsymbol{\eta})\Big]^{\lambda} \\
    &= p(\bpi) p(\boldsymbol{\eta}) 
    \times \prod_{k=1}^K \prod_{i=1}^{N^k} p(\mathbf{D}_i^{l} \vert Z_{i} = k, \eta) p(Z_{i} = k \vert \bpi) \\
    &\quad \times \Bigg[\prod_{i=1}^{N^{u}}\sum_{k=1}^K \Big\{ p(\mathbf{D}_i^{u} \vert Z_{i} = k, \boldsymbol{\eta}) p(Z_{i} = k \vert \bpi)\Big\} \Bigg]^{\lambda} \\
    &\propto \underbrace{\prod_{k=1}^K\left\{\pi_k^{\alpha_k - 1} \prod_{v=1}^V \eta_{vk}^{\beta_{kv} - 1}\right\}}_\text{prior}
    \times 
    \underbrace{\prod_{k=1}^K \prod_{i=1}^{N^k} \Big\{ \prod_{v=1}^V \eta_{vk}^{D_{iv}}\times \pi_k \Big\}}_\text{labeled doc. likelihood} 
    \times \underbrace{\Bigg[\prod_{i=1}^{N^{u}} \sum_{k=1}^K \Big\{\prod_{v=1}^V \eta_{vk}^{D_{iv}}\times \pi_k  \Big\}\Bigg]^{\lambda}
}_\text{unlabeled doc. likelihood}
\end{split}
\end{equation}

The Q function (the expectation of the complete log-likelihood) is 
\begin{equation}
\begin{split}
  Q &\equiv \E_{\mathbf{Z} \vert \bpi^{(t)}, \boldsymbol{\eta}^{(t)}, D, C}[p(\bpi, \boldsymbol{\eta}, \mathbf{Z} \vert \mathbf{D}, \mathbf{C})] \\
    &= \sum_{k=1}^K \left[(\alpha_k - 1) \log \pi_k^{(t)} + \sum_{v=1}^V \left\{ (\beta_{kv} - 1) \log \eta_{vk}^{(t)} \right\}\right]  \\
    &\quad + \sum_{k=1}^K \sum_{i=1}^{N^k} \Big\{ \sum_{v=1}^V D_{iv} \log \eta_{vk}^{(t)} + \log \pi_{k}^{(t)} \Big\} \\
    &\quad + \lambda \left[ \sum_{i=1}^{N^{u}}\sum_{k=1}^K  p_{ik} \Big\{ \sum_{v=1}^V D_{iv} \log \eta_{vk}^{(t)} + \log \pi_k^{(t)} \Big\}\right] 
\end{split}
\end{equation}

The posterior probability of $Z_i = k$, $p_{ik}$, is 
\begin{equation}
\begin{split}
  p_{ik} &= \frac{\prod_{v=1}^V \eta_{vk}^{D_{iv}} \times \pi_k}{\sum_{k=1}^K \left[\prod_{v=1}^V \eta_{vk}^{D_{iv}} \times \pi_k \right]}
\end{split}
\end{equation}

M step estimators are
The updating equation for $\pi$ is the following.
\begin{equation}
\begin{split}
  \hat{\pi_k} \propto \alpha_k - 1 +  N^k + \lambda \sum_{i=1}^{N^u} p_{ik} 
\end{split}
\end{equation}

The updating equation for $\eta$ is the following.
\begin{equation}
\begin{split}
\hat{\eta}_{vk} \propto (\beta_k -1) + \sum_{i=1}^{N^k}  D_{iv} + \lambda \sum_{i=1}^{N^{u}} p_{ik} D_{iv} 
\end{split}
\end{equation}
Note that we downweight the information from the unlabeled documents by $\lambda$, to utilize more reliable information from labeled documents.

\subsection{Results}
\begin{figure}[h!]
\centering
\includegraphics[scale=0.85]{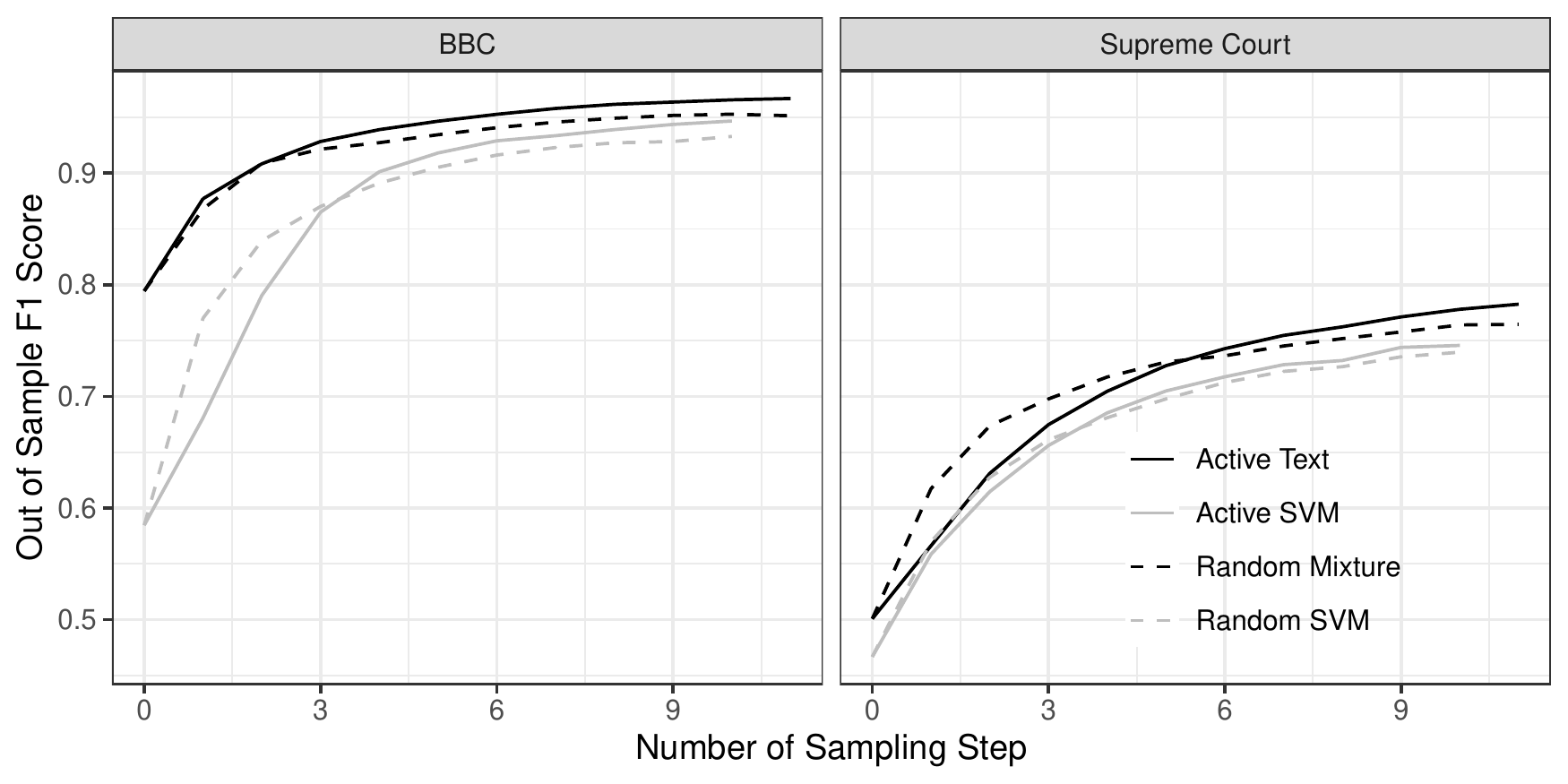}
  \caption{\textbf{Multiclass Classification Results.}\\
  The darker lines show the results with \aT and the lighter lines show the results with SVM.
  The solid lines use active sampling to decide the next set of documents to be labeled, and the dashed lines use random (passive) sampling.
  The y-axis indicates the out-of-sample F1 score and the x-axis show the number of sampling steps.
  The left column shows the results on BBC corpus, where the target classes are ``Politics,'' ``Entertainment,'' ``Business,'' ``Sports,'' and ``Technology.''
  ``Politics'' class has 5\% of the total dataset, and the rest 95\% is evenly split across the rest of classes.
  The right column shows the results on the Supreme Court corpus, where the target classes are ``Criminal Procedure'' (32.4\% of the corpus), ``Civil Rights'' (21.4\%), ``Economic Activity'' (22.2\%), ``Judicial Power'' (15.4\%), ``First Amendment (8.6\%).''
  In our model, we set the number of latent clusters to be the same as the classification categories and linked each latent cluster to one classification category.
  \aT performs the best across the four specifications on both corpora.
  }
  \label{fig:multiclass}
\end{figure}

\begin{figure}[h!]
\centering
\includegraphics[scale=0.85]{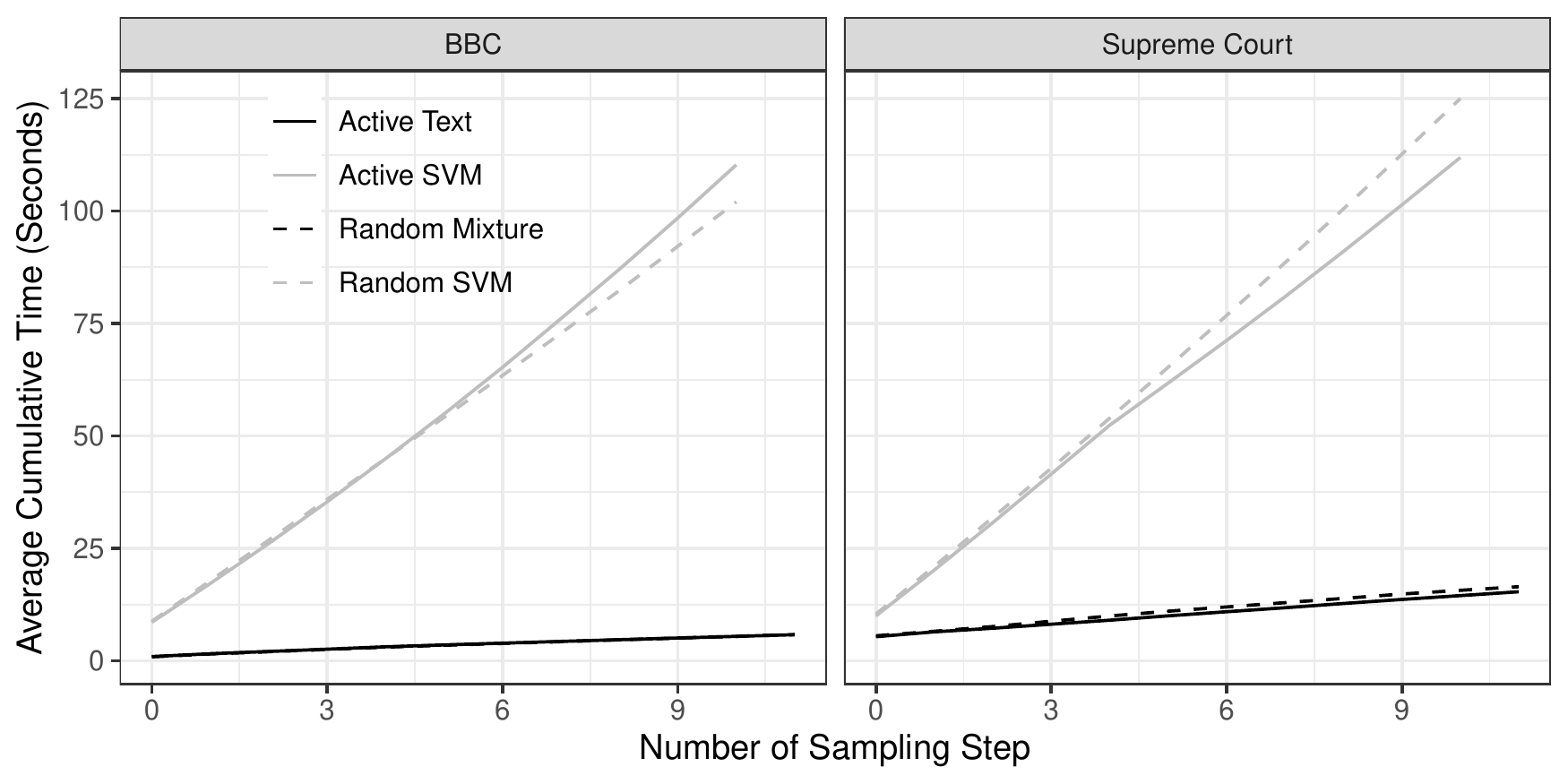}
  \caption{\textbf{Time comparison of Multiclass Classification Results.}\\
  The darker lines show the results with \aT and the lighter lines show the results with SVM.
  The solid lines use active sampling to decide the next set of documents to be labeled, and the dashed lines use random (passive) sampling.
  The y-axis indicates the average cumulative computational time and the x-axis shows the number of sampling steps.
  The left column shows the results on BBC corpus, and the right column shows the results on the Supreme Court corpus.
  \aT is much faster than SVM in multiclass classification. 
  This is because multiclass classification with SVM requires fitting the model repeatedly at least the same time as the number of target classes.
  By contrast, \aT requires to fit only once regardless of the number of target classes.
  }
  \label{fig:multiclass_time_linear}
\end{figure}

\clearpage
\section{Model Specifications and Description of the Datasets in the Validation Performance}
\label{sec:validation_specification}

We explain our decisions regarding pre-processing steps, model evaluation, and model specifications, followed by a detailed discussion of the results for each dataset.

%\subsection{Specifications}
\subsection{Pre-processing}
We employ the same pre-processing step for each of the four datasets using the \textit{R} package \textit{Quanteda}.\footnote{See \url{https://quanteda.io}}
For each dataset, we construct a \textit{document-feature matrix} (DFM), where each row is a document and each column is a feature.
Each feature is a stemmed unigram. We remove stopwords, features that occur extremely infrequently, as well as all features under 4 characters.%\footnote{For complete DFM specifications, see Appendix.}

To generate dataset with the proportion of positive class $p$ (e.g. 5\% or 50\%), we randomly sample documents from the original dataset so that it achieves the proportion of the positive class $p$. 
Suppose the number of documents in the original dataset is $N$ with $N_{pos}$ and $N_{neg}$ the number of positive and negative documents, respectively.
We compute $M_{pos}=\text{floor}(Np)$ and $M_{neg}=N - M_{pos}$ as the ideal numbers of positive and negative documents.  
While $M_{pos} > N_{pos}$ or $M_{neg} > N_{neg}$, we decrement $M_{pos}$ and $M_{neg}$ keeping the positive proportion to $p$.
With $M_{pos} < N_{pos}$ and $M_{neg} < N_{neg}$, we sample $M_{pos}$ positive documents and $M_{neg}$ negative documents from the original dataset. 
Finally, combine the sampled positive and negative documents to obtain the final dataset.
\subsection{Datasets}

\paragraph{BBC News}
The BBC News Dataset is a collection of 2,225 documents from 2004 to 2005 available at the BBC news website \citep{greene06icml}.
This dataset is divided equally into five topics: business, entertainment, politics, sport, and technology. 
The classification exercise is to correctly predict whether or not an article belongs to the `politics' topic.

\paragraph{Wikipedia Toxic Comments}
The Wikipedia Toxic Comments dataset is a dataset made up of conversations between Wikipedia editors in Wikipedia's internal forums.
The dataset was made openly available as part of a Kaggle competition,\footnote{See \url{https://www.kaggle.com/c/jigsaw-toxic-comment-classification-challenge}} and was used as a principle dataset of investigation by \cite{millerActiveLearningApproaches2020}.
The basic classification task is to label a given speech as toxic or not, where toxicity is defined as including harassment and/or abuse of other users.\footnote{While the dataset also contains finer gradation of `types' of toxicity, we like \cite{millerActiveLearningApproaches2020} stick to the binary toxic-or-not classification task.}
The complete dataset is comprised of roughly 560,000 documents, roughly 10 percent of which are labeled as toxic.

\paragraph{Supreme Court Cases}
The Supreme Court Rulings dataset is a collection of the text of 2000 US Supreme Court rulings between 1946 and 2012.
We use the majority opinion of each case and the text was obtained through Caselaw Access Project.\footnote{\url{https://case.law}}
For the classification label, we use the categories created by the Supreme Court Database.\footnote{For a full list of categories, see \url{http://www.supremecourtdatabase.org/documentation.php?var=issueArea}.}
The classification exercise here is to correctly identify rulings that are categorized as `criminal procedure', which is the largest category in the corpus (26\% of all rulings). 

\paragraph{Human Rights Allegation}
Human Rights Allegation dataset contains more than 2 million sentences of human rights reports in 196 countries between 1996 and 2016, produced by Amnesty International, Human Rights Watch and the US State Department \citep{farissphysical}.
The classification goal is to identify sentences with physical integrity rights allegation (16\% of all reports).
Example violations of physical integrity rights include torture, extrajudicial killing, and arbitrary arrest and imprisonment.

\clearpage
\section{Additional Results on Classification Performance}\label{sec_si:add_results}

To complement the results presented in Figure~1 in the main text, Table~\ref{tab:comparison.ran.ent}
presents the results (across datasets) of fitting our model at the initial (iteration 0) and last active step (iteration 30). 
It is clear from the table that the improvements \aT
brings in terms of the F1-score, precision, and recall. Furthermore, after labeling
600 documents (20 per iteration), uncertainty sampling 
outperforms random sampling across evaluation metrics, which empirically validates the promise of 
active learning in terms of text classification. 

\begin{table}[h!]
	\centering
  \caption{\textbf{Classification Performance: Uncertainty vs Random Sampling with $\lambda = 0.001$}}
  \label{tab:comparison.ran.ent}
	\footnotesize
	\begin{tabular}{l l *{6}{c}}
		\toprule
		Dataset    & Active Step & \multicolumn{3}{c}{Uncertainty Sampling} & \multicolumn{3}{c}{Random Sampling} \\
		\cmidrule(lr){3-5} \cmidrule(lr){6-8}
		& & Precision & Recall & F1-score & Precision & Recall & F1-score \\ \midrule
				\multirow{2}{*}{\texttt{Wikipedia}}
		& 0              &   0.71    &   0.13  &  0.22 & 0.71   &    0.13  &  0.22 \\
		& 30   & 0.71    &   0.54  &  0.61 &  0.45  &     0.56  & 0.50 \\
		\midrule
				\multirow{2}{*}{\texttt{BBC}}
		& 0              & 0.33   &    0.86  &  0.48 &  0.33    &   0.86  &  0.48\\
		& 30   &  0.92    &   0.96  &  0.94 & 0.92     &  0.94  &  0.93				\\
		\midrule
		\multirow{2}{*}{\texttt{Supreme Court}}
		& 0              &   0.46     &   0.98  &  0.63  &  0.46   &    0.98  &  0.63\\
		& 30   &   0.85     &  0.91  &  0.88 & 0.75    &   0.96  &  0.84		 \\
		\midrule
				\multirow{2}{*}{\texttt{Human Rights}}
		& 0              &   0.61   &    0.01  &  0.02 &0.61   &    0.01  &  0.02 \\
		& 30   & 0.53   &    0.42  &  0.47  &          0.46   &    0.44  & 0.45 \\
		\bottomrule
	\end{tabular}
\end{table}

\begin{figure}[t!]
  \includegraphics[width=\linewidth]{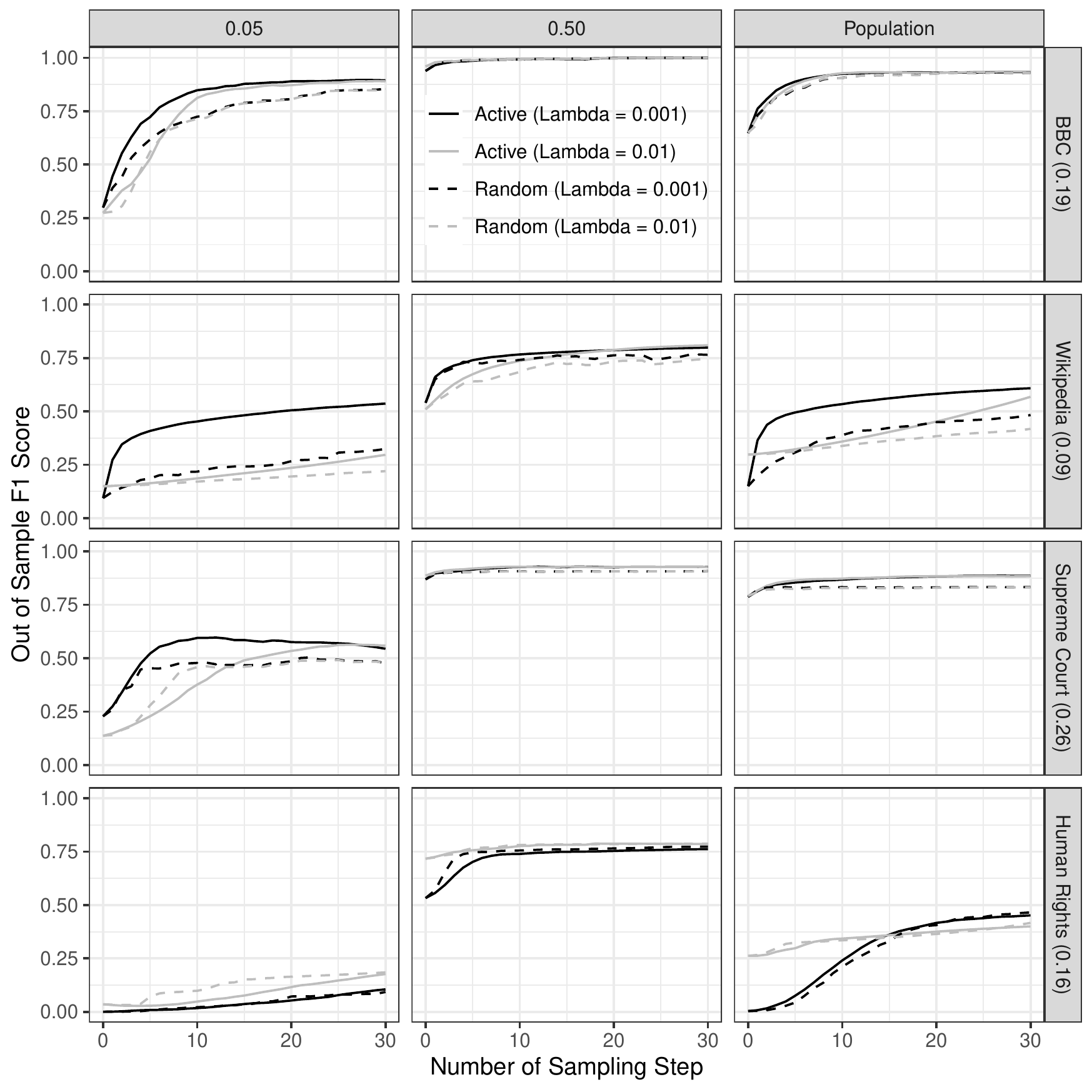}
  \caption{\textbf{Classification Results with 2 Clusters and $\lambda = 0.01$ vs $\lambda = 0.001$.}
  The darker lines show the results with $\lambda = 0.001$ and the lighter lines show $\lambda = 0.01$.
  The columns correspond to various proportion of positive labels in the corpus.
  The y-axis indicates the out-of-sample F1 score and the x-axis show the number of sampling steps.
  The smaller the value of $\lambda$ the better the performance of our model.
  }
  \label{fig:lambda_1}
\end{figure}

Similarly, and as noted in the main text, our results appear to be not
too sensitive to the selection of the weighting parameter $\lambda$, provided that its
value remains small. Figures~\ref{fig:lambda_1} confirms this finding. 
After 30 active steps, the performance of \aT is better in terms
of F1-score when $\lambda = 0.001$ if compared to $\lambda = 0.01$  

\clearpage
\section{Main Results when Varying Positive Class Rate}
\label{sec:main_results_appdx}
\begin{figure}[h!]
\centering
\includegraphics[scale=1]{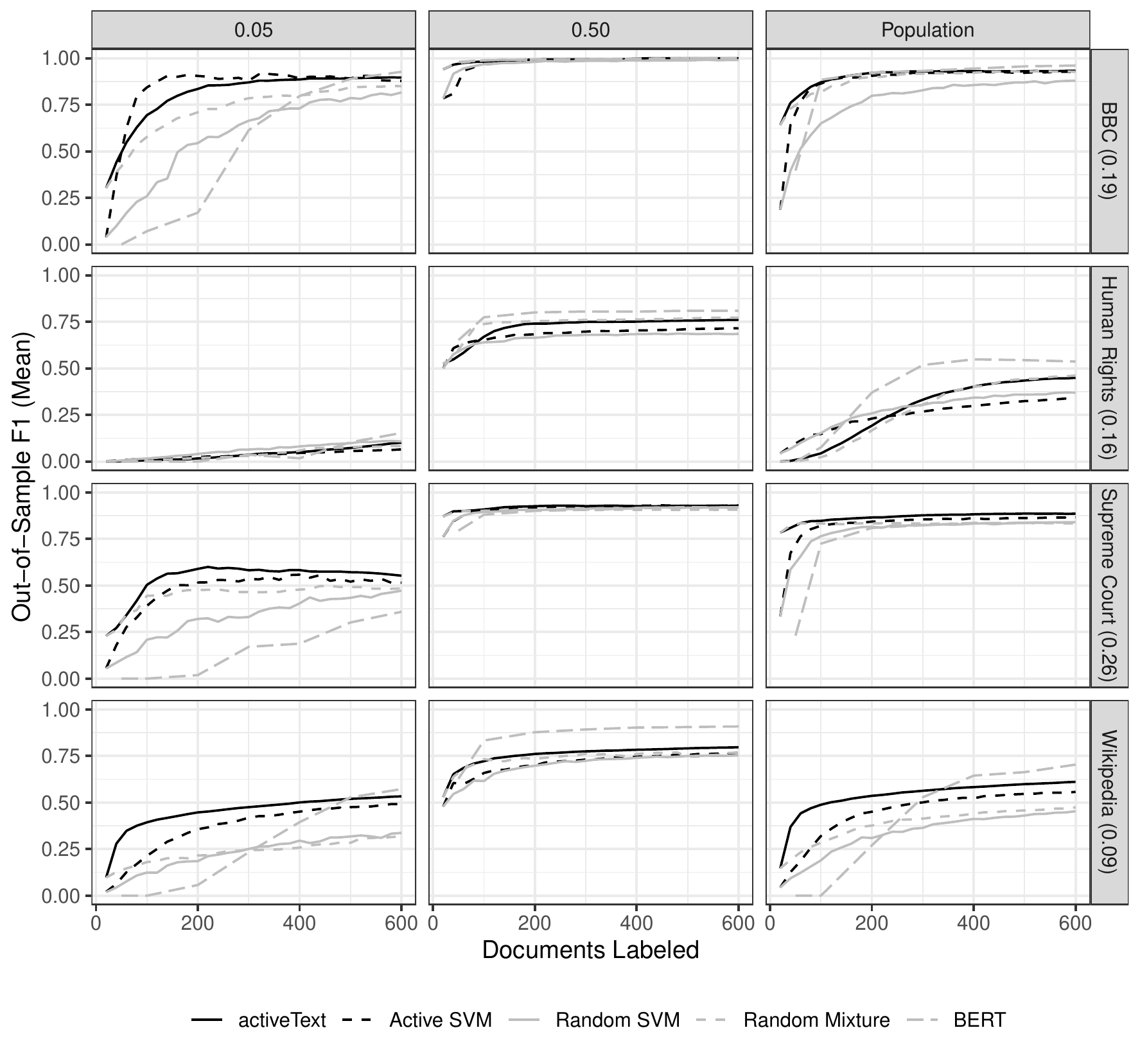}
  \caption{\textbf{Replication of F1 performance from Figures 2 and 3 with 0.05, 0.5, and population positive class rate}}
  \label{sifig:F1comparison}
\end{figure}

\clearpage
\section{Visual Demonstration of Active Keyword}
\label{sec:keywords_visual}

\begin{figure}[p!]
  \includegraphics[width=\linewidth]{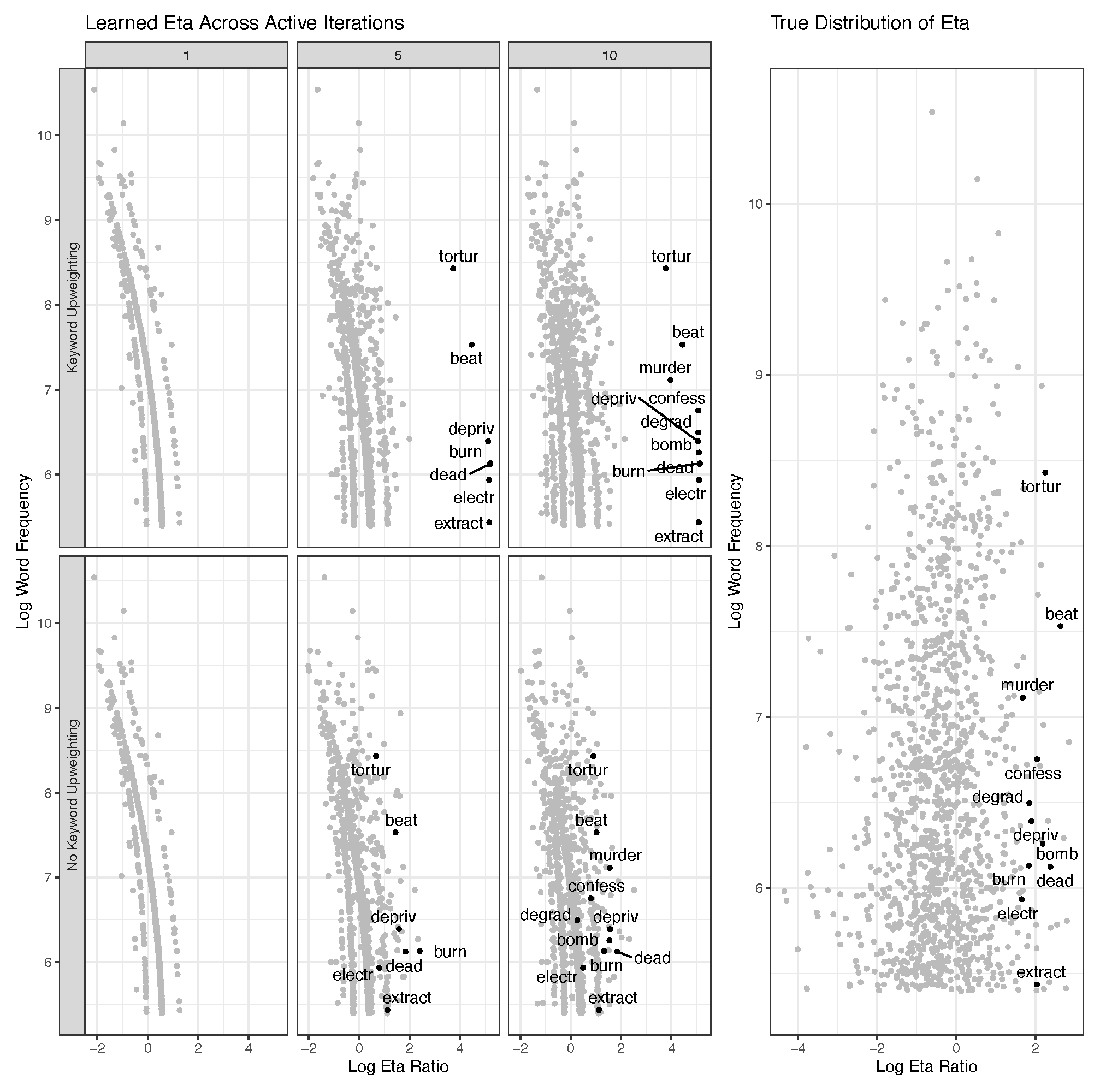}
  \caption{\textbf{Update of the Word-class Matrix ($\boldsymbol{\eta}$) with and without Keywords}
  %The right figure shows the distribution of the estimated word-class matrix $\boldsymbol{\eta}$ with fully labeled corpus.
  %The left figure shows how $\boldsymbol{\eta}$ is updated across active iterations.
  %The top row shows the updating process of $\boldsymbol{\eta}$ with keywords and the bottom row without.
  %The first column shows the initial values of $\boldsymbol{\eta}$, the middle column shows $\boldsymbol{\eta}$ after 5 iterations (100 labels),
  %and the left columns after 10 iterations (200 labels).
  %The x-axis shows the log ratio of $\eta_{v1}$ vs $\eta_{v0}$, where $K=2$ is linked to the positive class.
  %If this value is high, a word $v$ is more strongly associated with positive labels.
  %The y-axis is the log of word frequency. A word with high word frequency has more influence in shifting the label probability. 
  %A subset of keywords are labeled and highlighted by black dots.
  %Keywords scheme accelerates the learning process of $\boldsymbol{\eta}$ by upweighting the value of corresponding $\eta_{vk}$ in the positive direction. 
  }
  \label{fig:keywords_eta}
\end{figure}

Figure \ref{fig:keywords_eta} illustrates how the word-class matrix $\boldsymbol{\eta}$ is updated with and without keywords across iterations.
A subset of the keywords supplied is labeled and highlighted by black dots. 
The x-axis shows the log of $\eta_{v1} / \eta_{v0}$, where $\eta_{v1}$ corresponds the probability of observing the word $v$ in a document with a positive label and $\eta_{v0}$ for a document with a negative label. 
The high value in the x-axis means that a word is more strongly associated with positive labels.
The y-axis is the log of word frequency.
A word with high word frequency has more influence in shifting the label probability. 
In the generative model for \aT, words that appear often and whose ratio of $\eta_{vk^*}$ vs $\eta_{vk}$ is high play a central role in the label prediction.
By shifting the value of $\boldsymbol{\eta}$ of those keywords, we can accelerate the estimation of $\boldsymbol{\eta}$ and improve the classification performance. 

\clearpage

\section{Classification Performance with Mislabels}
\label{sec:mislabel}

\subsection{Mislabeled Keywords}\label{subsec:mislabel-keywords}
The rows correspond to different datasets and the columns correspond to various values of $\gamma$, which controls the degree of keyword upweighting.
The y-axis indicates the out-of-sample F1 score and the x-axis shows the number of sampling steps. 
At each sampling step, 20 documents are labeled.
We use $\lambda = 0.001$ to downweight information from unlabeled documents.
The lines correspond to different levels of mislabels at the keyword labeling. 
At each iteration, 10 candidate keywords are proposed, and a hypothetical oracle decides if they are indeed keywords or not. 
`True' keywords are defined in the same way as in Section~\nameref{subsec:keyword_results}. 
In other words, a candidate keyword $v$ for the positive class is a `true' keyword, if the value of $\eta_{v,k}/\eta_{v,k'}$ is above 90\% quantile, where $k$ is the positive class and $k'$ is the negative class, and this $\boldsymbol{\eta}$ is what we obtain by training the model with the full labels.
The same goes for the negative class.
When the probability of mislabeling keywords is $p$\%, an oracle makes a mistake in the labeling with probability $p$.
Specifically, if a candidate keyword $v$ is a `true' keyword, the oracle would not label $v$ as a keyword with probability $p$.
Likewise, if a candidate keyword $v$ is not a `true' keyword, they would label $v$ as a keyword.

\begin{figure}[t!]
  \centering
  \includegraphics[width=0.9\textwidth]{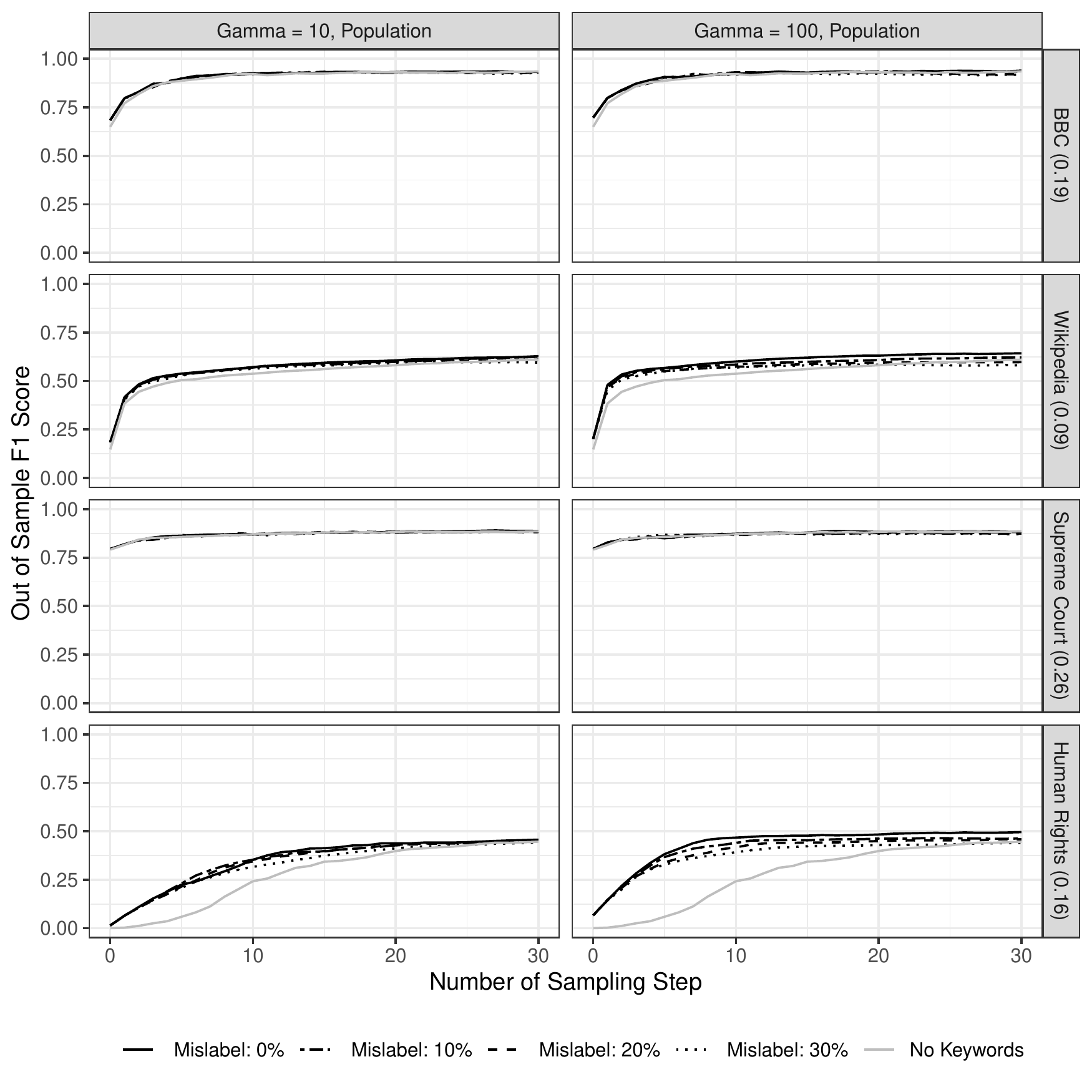}
  \caption{\textbf{Classification Results with Mislabels in Active Keywords}\\
  }
  \label{fig:keyword_mislabel}
\end{figure}

\clearpage
\subsection{Mislabeled Documents}
\label{subsec:mislabel_documents}

\begin{figure}[h!]
  \centering
  \includegraphics[width=\textwidth]{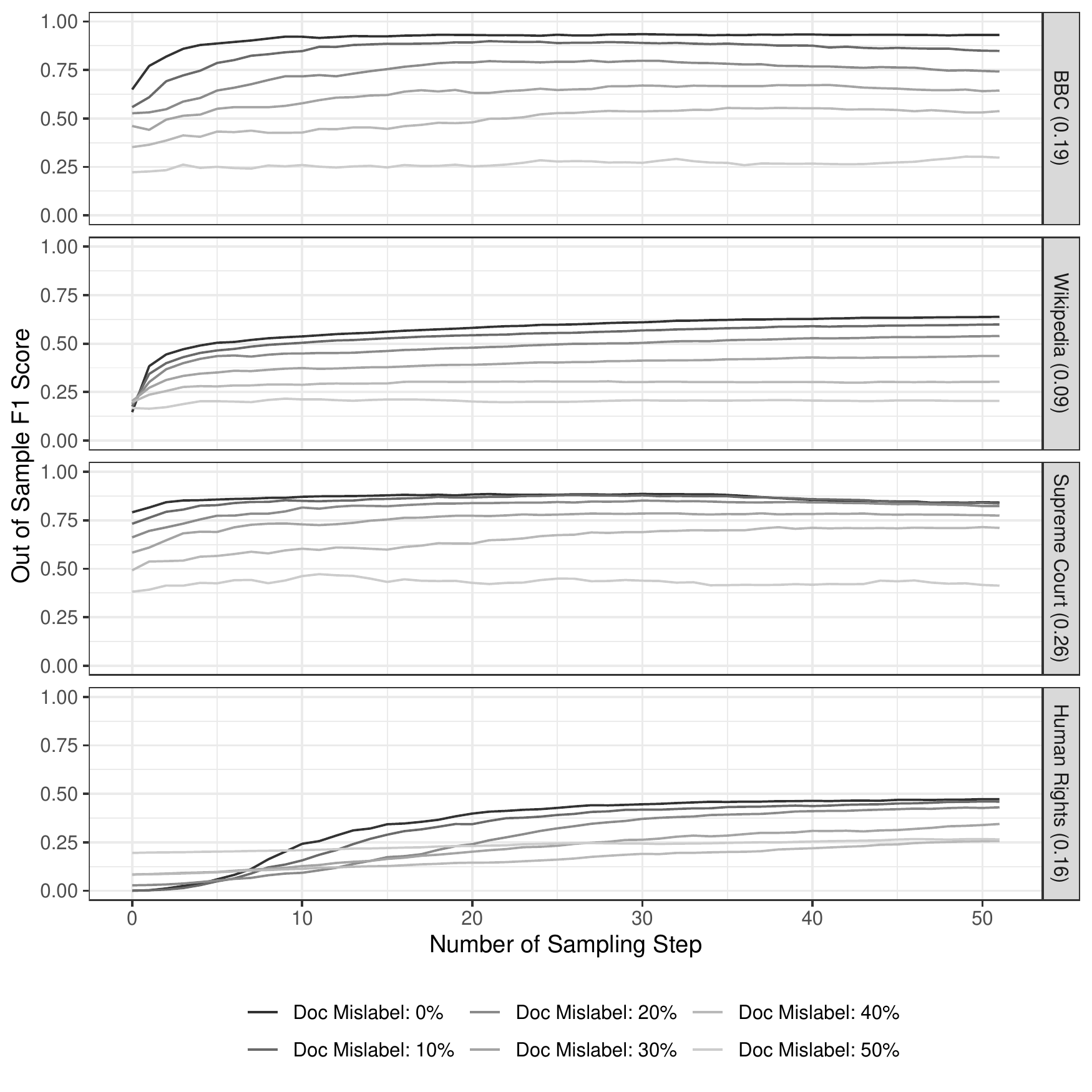}
  \caption{\textbf{Classification Results with Mislabels in Active Document Labeling}\\
    The rows correspond to different datasets.
  The y-axis indicates the out-of-sample F1 score and the x-axis shows the number of sampling steps. 
  20 documents are labeled at each sampling step.
  The colors correspond to different levels of mislabels in the labeling of documents. 
  We find that as the proportion of mislabels increases, the classification performance of \aT decreases. 
  }
 \label{fig:doc_mislabel}
\end{figure}

In this section, we present results about the effect of  `honest' (random) mislabeling of
documents on the mapping of documents to classes. As Figure~\ref{fig:doc_mislabel} shows,
as the proportion of mislabels increases, the classification performance of \aT decreases. 

% \clearpage
% \section{Time Comparison With BERT}
% \begin{figure}[h!]
%   \centering
%   \includegraphics[width=0.8\linewidth]{bert_time_comparison.pdf}
%   \label{fig:bert_time_compare}
%   \caption{
%     \textbf{Comparing Time Performance Between BERT and Active SVM and Mixture}\\
%     The x-axis indicates the number of documents labeled, and the y-axis indicates the logged amount of time in seconds that it takes each model to run. The values for the Active SVM and Active Mixture models are calculated cumulatively, since active models are re-run in 20 documents increments, whereas the values for the BERT model are not calculated cumulatively, since there is no need to re-run a BERT model in an active loop. Because the time to label documents is constant across all models, only the time to fit the models is included. All models were trained using a 50,000 document subsample of the Wikipedia toxic comments corpus on a base 2021 M1 Macbook Air. To train the BERT models, we used a PyTorch model with MPS graphical acceleration enabled (for more on this, see \href{https://pytorch.org/blog/introducing-accelerated-pytorch-training-on-mac/}{here}). The results show that the Active Mixture model is extremely quick relative to the Active SVM and BERT models.
%   }
% \end{figure}

\clearpage
\section{Comparison of the predictions between \aT and xgboost predictions for the \citet{gohdes2020repression} data}

Table \ref{tbl:gohdes_confusion_matrix} shows the confusion matrix between the prediction based on \aT and the prediction by xgboost used in the original paper. 
Most observations fall in the diagonal cells of the matrix, and the correlation between the two predictions is quite high (0.93).
One difference is that \aT classifies more documents to target killings compared to the original predictions.  
Note that either prediction claims the ground truth. Both are the results of different classifiers. 
\begin{table}[h!]
  \begin{tabular}{cc|ccc}
                         &                & \multicolumn{3}{c}{Original}\\
                         &                & untargeted & targeted & non-government\\
    \hline
    \multirow{3}{*}{\aT} & untargeted     & 50327      & 411      & 135\\
                         & targeted       & 1630       & 10044    & 31\\
                         & non-government & 382        & 34       & 2280\\
  \end{tabular}
  \caption{Confusion matrix between \aT and xgboost predictions}
\label{tbl:gohdes_confusion_matrix}
\end{table}

\begin{figure}[t!]
  \includegraphics[width=\linewidth]{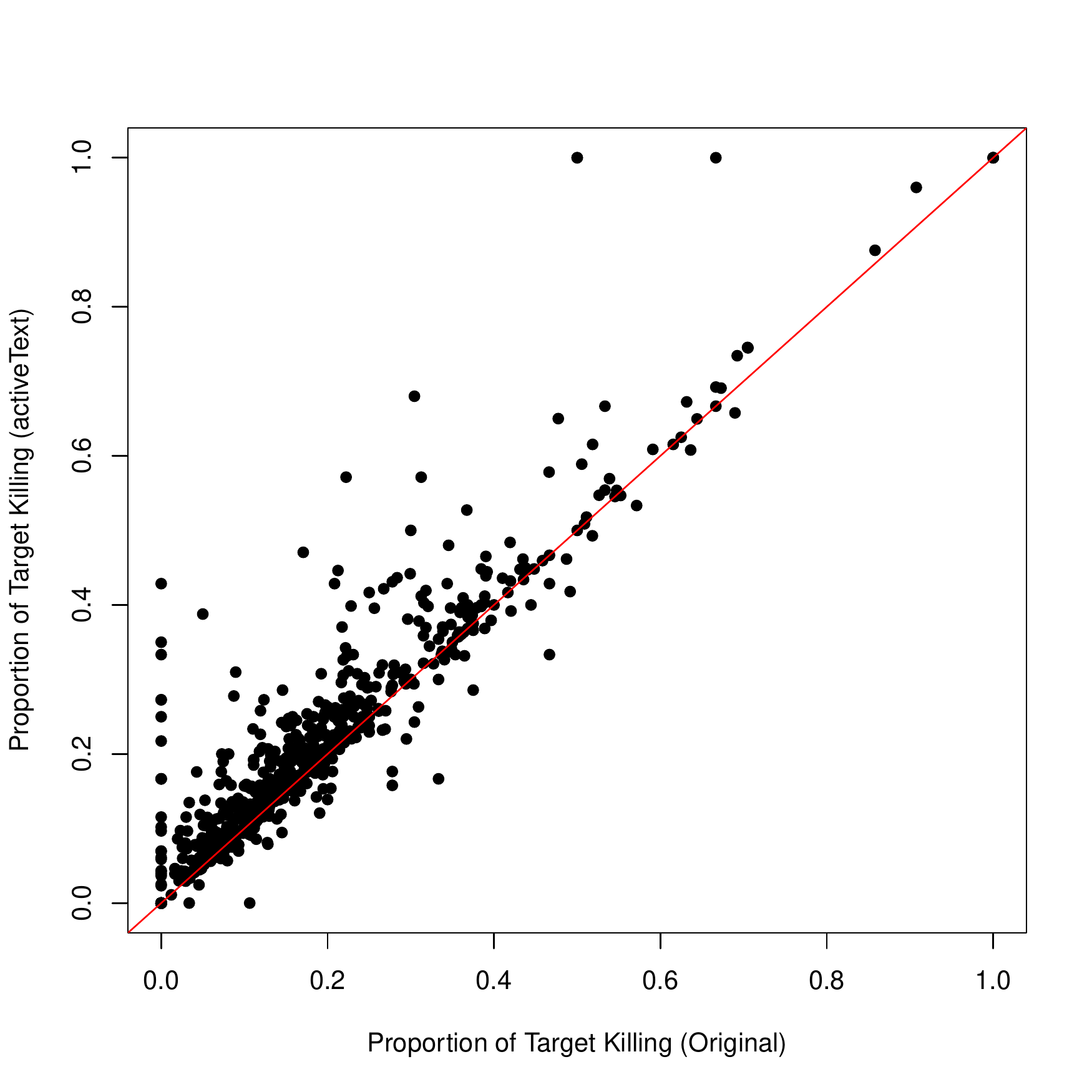}
  \caption{\textbf{Scatter plot of the dependent variable between the one constructed by \aT vs. the original}\
  The author performs a binomial logit regression where the dependent variable is the ratio of the number of targeted killings to the total number of government killings.
  We compare the dependent variable used in the original paper vs. the one we constructed using \aT. 
  The 45-degree line (in red) corresponds to equality between measures. 
  We can see that most observations lie around the 45-degree line while there are some values in the upper triangle. 
  This suggests that \aT yields a similar dependent variable to the original one, while there may be some overestimations of the proportion of target killing with \aT. 
  }
  \label{fig:scatter_predict_compare}
\end{figure}

\clearpage
\section{Regression Table in \citet{gohdes2020repression}}
\label{sec:syria_regression}

\begin{figure}[t!]
  \includegraphics[width=\linewidth]{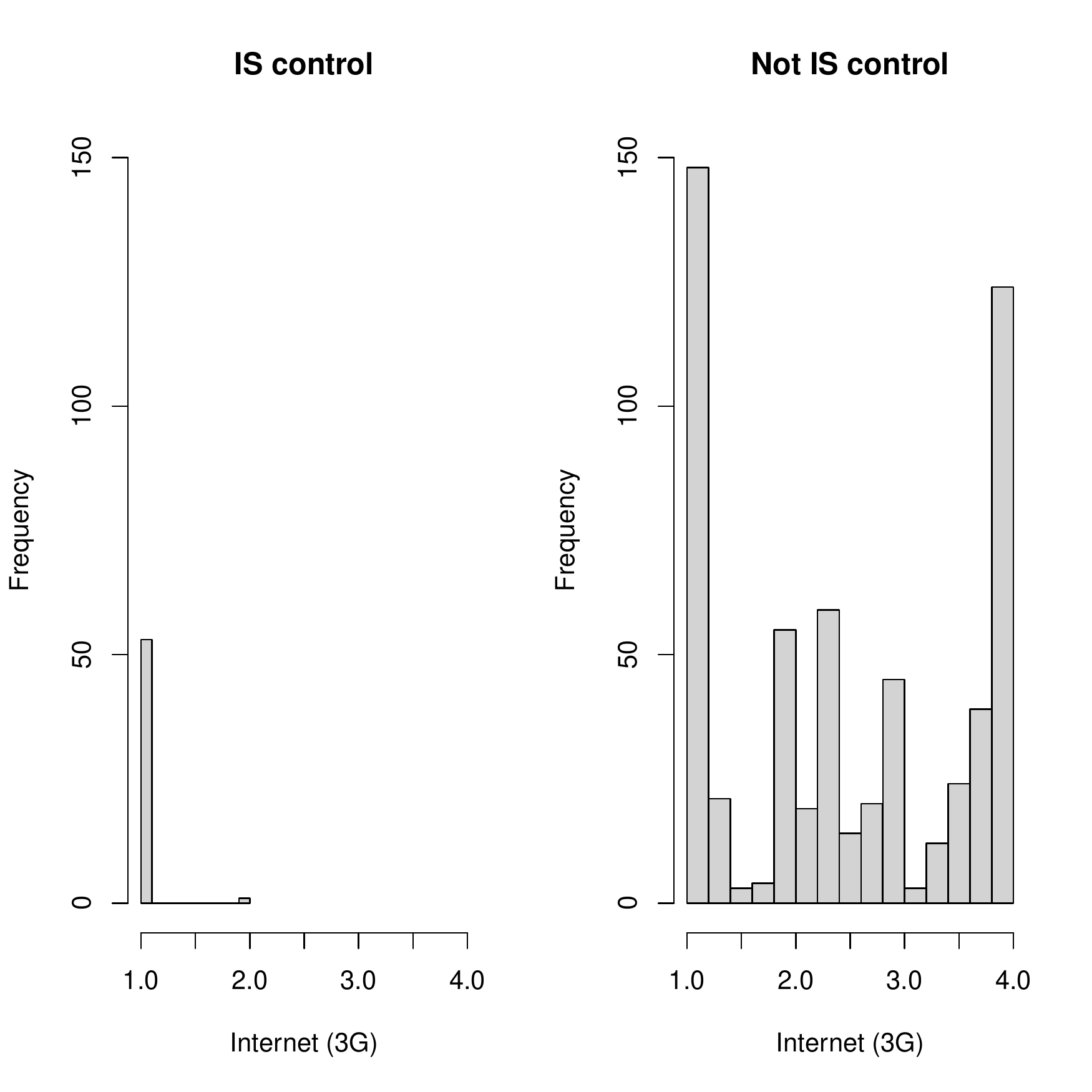}
  \caption{\textbf{Histogram of the Internet (3G) variable by the IS control in the original data}\
  The left histogram is the distribution of the Internet (3G) variable for the observation under IS control, and the right one is not under IS control.
  The number of observations with IS control is only 51 out of the total observation of 640. 
  In addition, among those with IS control, all observations except one takes the same value for the Internet access variable.
  This suggests that the regression coefficient on the interaction of IS control and Internet access can be highly unstable. 
  }
  \label{fig:internet_is}
\end{figure}

Table~\ref{tbl:gohdes_table_original} is the original regression table reported in \cite{gohdes2020repression} while Table~\ref{tbl:gohdes_table_active} is a replication of the original table using \aT.
In both tables, the coefficients on the Internet access variable are positive and statistically significant, which match the author's substantive conclusion.
One may wonder why the absolute values of the coefficients on the IS and Internet is larger in Table~\ref{tbl:gohdes_table_active}. 
However, we believe that this is because the number of observations in the IS control is small (51) and there is almost no variation of the Internet access variable within the observations with IS control, as shown in Figure~\ref{fig:internet_is}.

\begin{table}[h!]
\adjustbox{max width=\textwidth}{
\begin{tabular}{l c c c c c c c}
\hline
 & I & II & III & IV & V & VI & VII \\
\hline
Intercept                       & $-2.340^{***}$ & $-2.500^{***}$ & $-0.899^{*}$   & $-0.410$       & $-0.019$       & $-1.308$       & $-3.013^{**}$  \\
                                & $(0.205)$      & $(0.267)$      & $(0.403)$      & $(0.521)$      & $(0.357)$      & $(1.057)$      & $(1.103)$      \\
Internet access (3G)            & $0.224^{*}$    & $0.231^{*}$    & $0.200^{*}$    & $0.205^{*}$    & $0.265^{*}$    & $0.313^{**}$   & $0.909^{***}$  \\
                                & $(0.095)$      & $(0.094)$      & $(0.085)$      & $(0.087)$      & $(0.113)$      & $(0.116)$      & $(0.124)$      \\
\% Govt control                 &                &                &                &                &                &                & $0.016^{***}$  \\
                                &                &                &                &                &                &                & $(0.004)$      \\
Internet (3G) * \% Govt control &                &                &                &                &                &                & $-0.014^{***}$ \\
                                &                &                &                &                &                &                & $(0.001)$      \\
Govt control                    & $0.774^{*}$    & $0.803^{**}$   & $1.167^{***}$  & $1.180^{***}$  & $0.080$        & $0.856^{**}$   & $0.811^{***}$  \\
                                & $(0.332)$      & $(0.272)$      & $(0.284)$      & $(0.288)$      & $(0.344)$      & $(0.313)$      & $(0.237)$      \\
IS control                      & $2.027^{***}$  & $1.644^{***}$  & $1.045^{*}$    & $-0.324$       & $0.432$        & $0.787$        & $-0.663^{**}$  \\
                                & $(0.435)$      & $(0.462)$      & $(0.421)$      & $(0.209)$      & $(0.414)$      & $(0.418)$      & $(0.221)$      \\
Kurd control                    & $0.386$        & $-0.243$       & $-0.506$       & $-1.331$       & $-0.402$       & $0.033$        & $-0.616$       \\
                                & $(0.594)$      & $(0.843)$      & $(0.760)$      & $(1.134)$      & $(0.745)$      & $(0.802)$      & $(0.432)$      \\
Opp control                     & $1.160^{***}$  & $1.252^{***}$  & $0.727^{*}$    & $0.759^{*}$    & $-0.700^{*}$   & $-0.281$       & $-0.176$       \\
                                & $(0.298)$      & $(0.317)$      & $(0.293)$      & $(0.296)$      & $(0.283)$      & $(0.342)$      & $(0.164)$      \\
Internet (3G) * Govt control    & $-0.163$       & $-0.182$       & $-0.327^{**}$  & $-0.324^{**}$  & $-0.104$       & $-0.358^{**}$  &                \\
                                & $(0.132)$      & $(0.117)$      & $(0.119)$      & $(0.122)$      & $(0.133)$      & $(0.120)$      &                \\
Internet (3G) * IS control      & $-1.798^{***}$ & $-1.525^{***}$ & $-1.377^{***}$ &                & $-1.391^{***}$ & $-1.336^{***}$ &                \\
                                & $(0.220)$      & $(0.281)$      & $(0.251)$      &                & $(0.264)$      & $(0.261)$      &                \\
Internet (3G) * Kurd control    & $-0.133$       & $0.336$        & $0.093$        & $0.895$        & $-0.052$       & $-0.202$       &                \\
                                & $(0.444)$      & $(0.649)$      & $(0.569)$      & $(0.936)$      & $(0.553)$      & $(0.527)$      &                \\
Internet (3G) * Opp. control    & $-0.605^{***}$ & $-0.722^{***}$ & $-0.511^{**}$  & $-0.533^{***}$ & $0.316^{*}$    & $0.286$        &                \\
                                & $(0.159)$      & $(0.173)$      & $(0.157)$      & $(0.158)$      & $(0.151)$      & $(0.186)$      &                \\
\# Killings (log)               &                &                & $-0.273^{***}$ & $-0.271^{***}$ & $-0.354^{***}$ & $-0.412^{***}$ & $-0.584^{***}$ \\
                                &                &                & $(0.054)$      & $(0.055)$      & $(0.051)$      & $(0.072)$      & $(0.074)$      \\
Govt gains                      &                &                &                & $0.643$        &                &                &                \\
                                &                &                &                & $(0.385)$      &                &                &                \\
Govt losses                     &                &                &                & $0.632$        &                &                &                \\
                                &                &                &                & $(0.413)$      &                &                &                \\
Christian                       &                &                &                &                & $0.068$        & $0.345^{**}$   & $0.398^{***}$  \\
                                &                &                &                &                & $(0.111)$      & $(0.116)$      & $(0.110)$      \\
Alawi                           &                &                &                &                & $1.479^{**}$   & $-1.167^{***}$ & $-0.812^{***}$ \\
                                &                &                &                &                & $(0.522)$      & $(0.177)$      & $(0.176)$      \\
Druze                           &                &                &                &                & $-0.634^{***}$ & $-0.302$       & $0.135$        \\
                                &                &                &                &                & $(0.191)$      & $(0.191)$      & $(0.190)$      \\
Kurd                            &                &                &                &                & $-0.659^{***}$ & $-0.542^{*}$   & $-0.580^{**}$  \\
                                &                &                &                &                & $(0.194)$      & $(0.237)$      & $(0.212)$      \\
Internet (3G) * Alawi           &                &                &                &                & $-0.909^{***}$ &                &                \\
                                &                &                &                &                & $(0.163)$      &                &                \\
Pop (log)                       &                &                &                &                &                & $0.196$        & $0.408^{**}$   \\
                                &                &                &                &                &                & $(0.149)$      & $(0.150)$      \\
Unempl. (\%)                    &                &                &                &                &                & $-0.016$       & $-0.002$       \\
                                &                &                &                &                &                & $(0.012)$      & $(0.012)$      \\
\hline
AIC                             & $11956.847$    & $9993.704$     & $9665.749$     & $9495.591$     & $7671.979$     & $7873.915$     & $7327.796$     \\
BIC                             & $12001.524$    & $10239.427$    & $9915.941$     & $9744.552$     & $7944.509$     & $8150.913$     & $7595.858$     \\
Log Likelihood                  & $-5968.424$    & $-4941.852$    & $-4776.875$    & $-4691.796$    & $-3774.990$    & $-3874.958$    & $-3603.898$    \\
Deviance                        & $9519.651$     & $7466.508$     & $7136.554$     & $7026.891$     & $5132.784$     & $5332.720$     & $4790.601$     \\
Num. obs.                       & $640$          & $640$          & $640$          & $626$          & $640$          & $640$          & $640$          \\
\hline
\multicolumn{8}{l}{\scriptsize{$^{***}p<0.001$; $^{**}p<0.01$; $^{*}p<0.05$. Reference category: Contested control. Governorate-clustered SEs.}}
\end{tabular}}
\caption{Table 1 in Gohdes 2020: Original table}
\label{tbl:gohdes_table_original}
\end{table}

\begin{table}[h!]
\adjustbox{max width=\textwidth}{
\begin{tabular}{l c c c c c c c}
\hline
 & I & II & III & IV & V & VI & VII \\
\hline
Intercept                       & $-2.196^{***}$  & $-2.428^{***}$  & $-0.795^{*}$    & $-0.351$       & $-0.037$        & $-1.141$        & $-2.695^{*}$   \\
                                & $(0.197)$       & $(0.242)$       & $(0.390)$       & $(0.490)$      & $(0.348)$       & $(1.229)$       & $(1.227)$      \\
Internet access (3G)            & $0.277^{**}$    & $0.282^{***}$   & $0.242^{**}$    & $0.250^{**}$   & $0.342^{***}$   & $0.369^{***}$   & $0.853^{***}$  \\
                                & $(0.091)$       & $(0.081)$       & $(0.075)$       & $(0.077)$      & $(0.103)$       & $(0.107)$       & $(0.118)$      \\
\% Govt control                 &                 &                 &                 &                &                 &                 & $0.015^{***}$  \\
                                &                 &                 &                 &                &                 &                 & $(0.004)$      \\
Internet (3G) * \% Govt control &                 &                 &                 &                &                 &                 & $-0.013^{***}$ \\
                                &                 &                 &                 &                &                 &                 & $(0.001)$      \\
Govt control                    & $0.625^{*}$     & $0.672^{**}$    & $1.048^{***}$   & $1.058^{***}$  & $0.151$         & $0.843^{**}$    & $0.559^{*}$    \\
                                & $(0.319)$       & $(0.255)$       & $(0.269)$       & $(0.273)$      & $(0.358)$       & $(0.300)$       & $(0.249)$      \\
IS control                      & $15.157^{***}$  & $15.688^{***}$  & $15.072^{***}$  & $-0.275$       & $14.551^{***}$  & $14.877^{***}$  & $-0.600^{**}$  \\
                                & $(1.123)$       & $(1.148)$       & $(1.136)$       & $(0.200)$      & $(1.132)$       & $(1.134)$       & $(0.209)$      \\
Kurd control                    & $0.795$         & $0.099$         & $-0.227$        & $-0.440$       & $-0.157$        & $0.334$         & $-0.369$       \\
                                & $(0.516)$       & $(0.729)$       & $(0.671)$       & $(1.119)$      & $(0.677)$       & $(0.744)$       & $(0.405)$      \\
Opp control                     & $0.978^{***}$   & $1.134^{***}$   & $0.594^{*}$     & $0.634^{*}$    & $-0.606^{*}$    & $-0.197$        & $-0.278$       \\
                                & $(0.294)$       & $(0.304)$       & $(0.284)$       & $(0.289)$      & $(0.270)$       & $(0.322)$       & $(0.155)$      \\
Internet (3G) * Govt control    & $-0.169$        & $-0.190$        & $-0.334^{**}$   & $-0.335^{**}$  & $-0.183$        & $-0.408^{***}$  &                \\
                                & $(0.126)$       & $(0.103)$       & $(0.108)$       & $(0.111)$      & $(0.131)$       & $(0.111)$       &                \\
Internet (3G) * IS control      & $-14.829^{***}$ & $-15.506^{***}$ & $-15.351^{***}$ &                & $-15.392^{***}$ & $-15.330^{***}$ &                \\
                                & $(1.080)$       & $(1.096)$       & $(1.090)$       &                & $(1.091)$       & $(1.091)$       &                \\
Internet (3G) * Kurd control    & $-0.400$        & $0.138$         & $-0.080$        & $0.134$        & $-0.240$        & $-0.366$        &                \\
                                & $(0.324)$       & $(0.514)$       & $(0.463)$       & $(0.940)$      & $(0.473)$       & $(0.460)$       &                \\
Internet (3G) * Opp. control    & $-0.542^{***}$  & $-0.688^{***}$  & $-0.468^{**}$   & $-0.497^{**}$  & $0.181$         & $0.149$         &                \\
                                & $(0.159)$       & $(0.164)$       & $(0.150)$       & $(0.152)$      & $(0.145)$       & $(0.176)$       &                \\
\# Killings (log)               &                 &                 & $-0.278^{***}$  & $-0.274^{***}$ & $-0.356^{***}$  & $-0.415^{***}$  & $-0.567^{***}$ \\
                                &                 &                 & $(0.053)$       & $(0.054)$      & $(0.051)$       & $(0.071)$       & $(0.073)$      \\
Govt gains                      &                 &                 &                 & $0.512$        &                 &                 &                \\
                                &                 &                 &                 & $(0.349)$      &                 &                 &                \\
Govt losses                     &                 &                 &                 & $0.730^{*}$    &                 &                 &                \\
                                &                 &                 &                 & $(0.334)$      &                 &                 &                \\
Christian                       &                 &                 &                 &                & $0.092$         & $0.352^{**}$    & $0.369^{***}$  \\
                                &                 &                 &                 &                & $(0.115)$       & $(0.113)$       & $(0.105)$      \\
Alawi                           &                 &                 &                 &                & $1.329^{*}$     & $-0.928^{***}$  & $-0.585^{***}$ \\
                                &                 &                 &                 &                & $(0.528)$       & $(0.167)$       & $(0.168)$      \\
Druze                           &                 &                 &                 &                & $-0.628^{**}$   & $-0.310$        & $0.063$        \\
                                &                 &                 &                 &                & $(0.196)$       & $(0.197)$       & $(0.209)$      \\
Kurd                            &                 &                 &                 &                & $-0.565^{**}$   & $-0.502^{*}$    & $-0.615^{**}$  \\
                                &                 &                 &                 &                & $(0.204)$       & $(0.227)$       & $(0.207)$      \\
Internet (3G) * Alawi           &                 &                 &                 &                & $-0.782^{***}$  &                 &                \\
                                &                 &                 &                 &                & $(0.164)$       &                 &                \\
Pop (log)                       &                 &                 &                 &                &                 & $0.185$         & $0.391^{*}$    \\
                                &                 &                 &                 &                &                 & $(0.167)$       & $(0.168)$      \\
Unempl. (\%)                    &                 &                 &                 &                &                 & $-0.019$        & $-0.007$       \\
                                &                 &                 &                 &                &                 & $(0.012)$       & $(0.012)$      \\
\hline
AIC                             & $12050.644$     & $10116.531$     & $9739.975$      & $9570.556$     & $8038.596$      & $8197.433$      & $7735.527$     \\
BIC                             & $12095.321$     & $10362.255$     & $9990.166$      & $9819.517$     & $8311.125$      & $8474.431$      & $8003.589$     \\
Log Likelihood                  & $-6015.322$     & $-5003.266$     & $-4813.988$     & $-4729.278$    & $-3958.298$     & $-4036.717$     & $-3807.763$    \\
Deviance                        & $9500.059$      & $7475.946$      & $7097.391$      & $6986.658$     & $5386.011$      & $5542.849$      & $5084.942$     \\
Num. obs.                       & $640$           & $640$           & $640$           & $626$          & $640$           & $640$           & $640$          \\
\hline
\multicolumn{8}{l}{\scriptsize{$^{***}p<0.001$; $^{**}p<0.01$; $^{*}p<0.05$. Reference category: Contested control. Governorate-clustered SEs.}}
\end{tabular}}
  \caption{Table 1 in Gohdes 2020: Reanalysis with \aT}
\label{tbl:gohdes_table_active}
\end{table}

\clearpage
\section{Effect of Labeling More Sentences for the \citet{park:etal:2020} Reanalysis}\label{sec:colaresi}
In this section, we present additional results mentioned in the main text about our reanalysis of \citet{park:etal:2020}.

\begin{figure}[h!]
 \centering
 \includegraphics[width=0.8\textwidth]{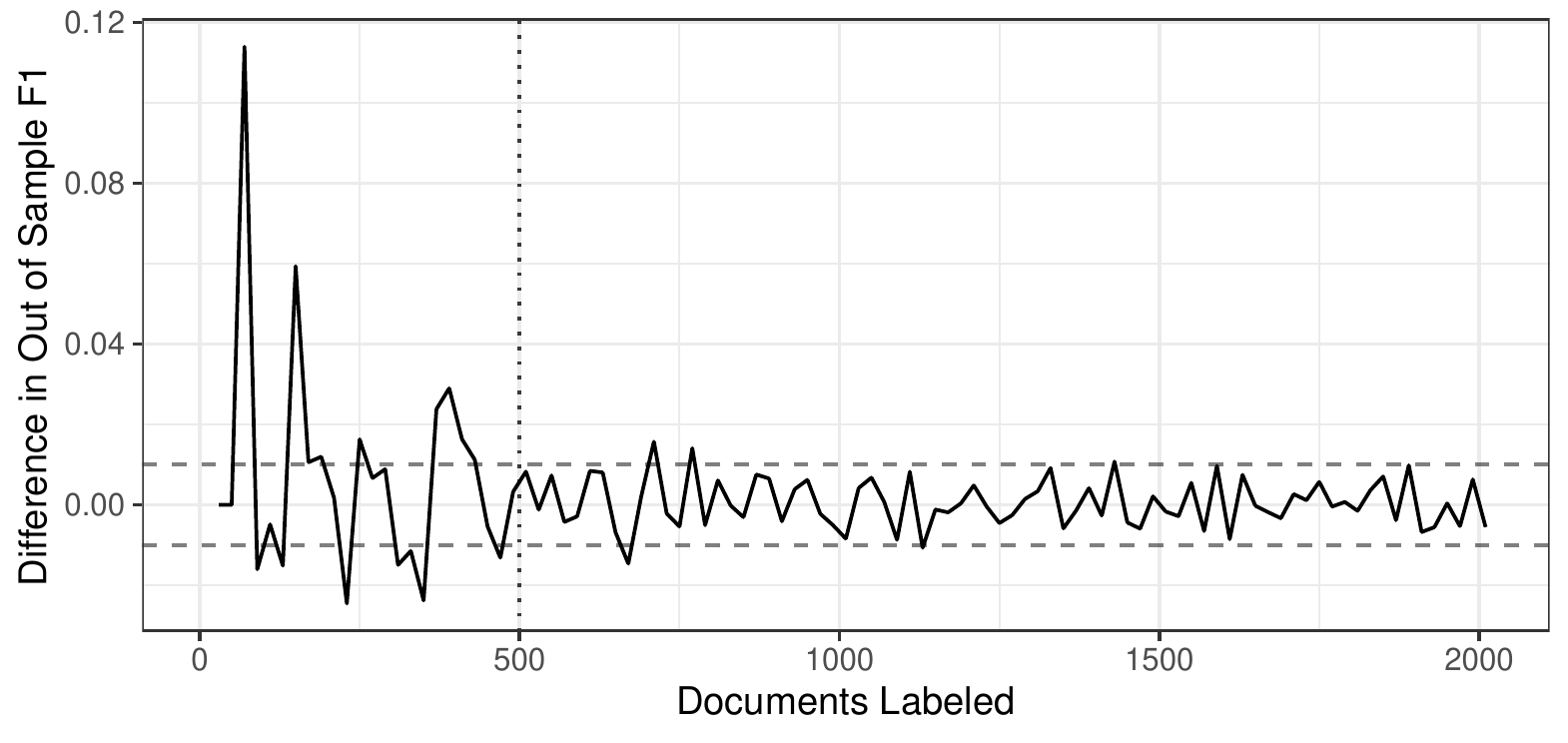}
 \caption{\textbf{Using the Difference in Out of Sample F1 Score to Decide a Stopping Point.}}
 \label{fig:colaresi_fig2}
\end{figure}

\begin{figure}[h!]
  \centering
  \includegraphics[width=0.8\textwidth]{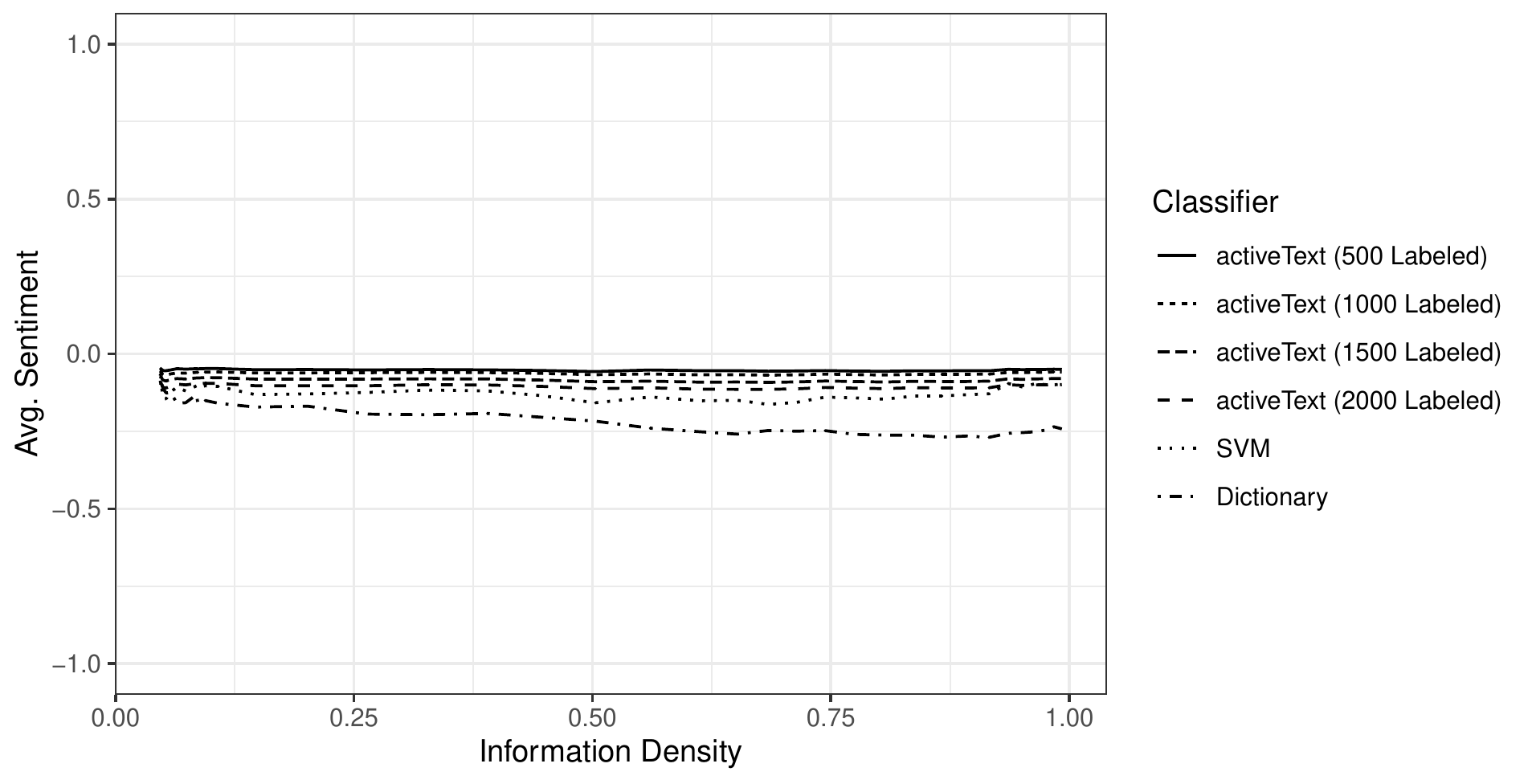}
  \caption{\textbf{Replication of Figure 1 in \cite{park:etal:2020}: The Relationship Between Information Density and Average Sentiment Score Across Different Settings for the Total Number of Labeled Documents.}}
  \label{fig:colaresi_fig3}
\end{figure}

% \begin{figure}[h!]
%   \centering
%   \includegraphics[width=0.8\textwidth]{figs/colaresi_fig4.pdf}
%   \caption{\textbf{Out of Sample F1 Score as a Function of Documents Labeled.} The vertical line represents the active step at which the improvements in the Out-of-sample F1 score grow less than 0.01 units from one iteration to the next.}
%   \label{fig:colaresi_fig4}
% \end{figure}

\clearpage
%%% Local Variables:
%%% mode: yatex
%%% TeX-master: "../active.tex"
%%% End: